\PassOptionsToPackage{dvipsnames,table}{xcolor}
\documentclass{article}
\usepackage{arxiv}

\usepackage[utf8]{inputenc}
\usepackage[T1]{fontenc}
\usepackage{amsmath,amsfonts}
\usepackage{bm}
\usepackage{algorithmic}
\usepackage{algorithm}
\usepackage{array}
\usepackage{orcidlink}
\usepackage[caption=false,font=normalsize,labelfont=sf,textfont=sf]{subfig}
\usepackage{textcomp}
\usepackage{url}
\usepackage{booktabs}
\usepackage{multirow}
\usepackage{hyperref}
\usepackage{verbatim}
\usepackage{graphicx}
\usepackage{cite}
\usepackage{microtype}

\newcommand{\deleted}[1]{}
\newcommand{\R}{\mathbb{R}}

\newcommand{\norm}[1]{\left\|#1\right\|}
\newcommand{\softadj}{\widetilde{\mathbf{A}}}   
\newcommand{\softaij}{\tilde{a}_{ij}}            
\newcommand{\softdeg}{\tilde{d}}                 
\newcommand{\logp}{\bm{\ell}}                    
\newcommand{\scos}[2]{\cos\!\bigl(#1,\,#2\bigr)} 
\newcommand{\emb}{\mathbf{z}}                    
\newcommand{\proto}{\bm{\pi}}                    
\newcommand{\featmat}{\widetilde{\mathbf{H}}}    
\newcommand{\backbone}{f_\phi}                   

\DeclareMathOperator*{\argmin}{arg\,min}

\def\BibTeX{{\rm B\kern-.05em{\sc i\kern-.025em b}\kern-.08em
    T\kern-.1667em\lower.7ex\hbox{E}\kern-.125emX}}

\begin{document}

\title{ProtoGuide: Prototype-Driven Guidance for\\Class-Conditional Graph Generation}

\renewcommand{\headeright}{Preprint}
\renewcommand{\undertitle}{Preprint}
\renewcommand{\shorttitle}{ProtoGuide: Prototype-Driven Guidance for Class-Conditional Graph Generation}

\author{
  \textbf{Salvatore Romano}\textsuperscript{1,2,3,\orcidlink{0009-0002-2550-4435}}\quad
  \textbf{Marco Grassia}\textsuperscript{1,\orcidlink{0000-0001-5841-6058}}\quad
  \textbf{Pietro Li\`{o}}\textsuperscript{2,\orcidlink{0000-0002-0540-5053}}\quad
  \textbf{Giuseppe Mangioni}\textsuperscript{1,*,\orcidlink{0000-0001-6910-0112}}\\[1.5ex]
  \textsuperscript{1}Department of Electrical, Electronic and Computer Engineering, University of Catania, Italy\\
  \textsuperscript{2}Department of Computer Science and Technology, University of Cambridge, UK\\
  \textsuperscript{3}University Campus Bio-Medico of Rome, Italy\\[1.2ex]
  \small *Corresponding author: \texttt{giuseppe.mangioni@unict.it}
}

\date{}
\maketitle

\begin{abstract}
     Discrete diffusion models are a prominent family for graph generation, but standard class-conditional mechanisms embed the class signal in the denoiser during training, tying the conditioning mechanism to the trained model.
     Classifier guidance avoids this coupling in continuous domains by steering a frozen model with a classifier's gradient, but discrete graph diffusion samples discrete edge states, so gradients cannot propagate through the sampled graph.
     We introduce ProtoGuide, a post-hoc, backbone-agnostic framework that recovers an analogous mechanism.
     At each reverse step the denoiser's per-edge output is relaxed into a differentiable soft adjacency, embedded by a frozen Siamese graph neural network, and scored against a target-class prototype and its nearest competitor; the resulting per-edge gradient, damped by a cosine schedule, is injected back into the denoiser output.
     All components stay frozen, so guidance is retargeted by supplying a different prototype.
     On five classes of real-world networks and two architecturally different backbones, EDGE and DiGress, ProtoGuide raises macro classification accuracy from 50.7\% to 73.5\% and from 73.6\% to 83.8\%, and outperforms DiGress's built-in conditional training under our configuration.
     Gains are largest where the unguided models are weakest, and are not uniform across classes.
     Per-graph coverage remains high in most settings, while distributional effects are class-dependent.
     A Best-of-$N$ selection baseline matches this accuracy given enough oversampling, but at a substantial cost in graph diversity.
     An independently initialized classifier, a directionality test, and a few-shot analysis support target-directed steering and robustness to very small support sets.
     \looseness=-1
\end{abstract}


\section{Introduction}
\label{sec:intro}

Complex systems across scientific and technological domains are commonly represented as networks of interacting entities, making network-based modeling a common language for studying systems ranging from technological infrastructures to the brain and biological organization.
Their topology is not incidental: different classes of real-world networks exhibit distinct combinations of degree heterogeneity, clustering, connectivity, and mesoscopic organization, which shape the processes taking place on them.
Generating realistic synthetic instances of such networks is therefore a broad problem across disciplines that study complex systems, with applications ranging from benchmarking algorithms~\cite{Lancichinetti_2008} and studying robustness and cascading failures~\cite{grassia2024robustness,adler2026latent} to modeling network structure in connectomics~\cite{grassia2026machineenhanced} and systems biology~\cite{Barabasi2004Network}.
Yet, this task is challenging because realism requires reproducing combinations of structural properties rather than isolated statistics. Throughout this work, we use \emph{network} when referring to the empirical system and \emph{graph} for its mathematical representation and generative model.

Recent progress in graph generation has increasingly relied on diffusion models.
Much of this progress, however, targets molecular datasets: relatively small graphs with constrained vocabularies of node and edge types.
Real-world complex networks pose a different generative problem: their structure is primarily encoded in topology and can exhibit broad heterogeneity in size, degree distributions, clustering, and mesoscopic organization.
Discrete denoising diffusion models, which operate directly on discrete edge variables, provide a flexible data-driven approach to modeling such graph distributions~\cite{DBLP:conf/iclr/VignacKSWCF23,3618408.3618589,haefeli2023diffusionmodelsgraphsbenefit}.
\looseness=-1

Alongside structural fidelity, many applications additionally require controllability: one may need realistic samples from a specific network class, or may wish to change the desired target or conditioning objective after the generative model has been trained.
A standard solution is conditional diffusion~\cite{DBLP:conf/iclr/VignacKSWCF23,10.1609/aaai.v37i4.25549}, where the class signal is incorporated into the denoiser during training.
This makes the conditioning mechanism part of the trained generative model, so introducing a target class or conditioning objective not represented during training generally requires adapting or retraining the model.

In image synthesis, classifier guidance~\cite{3540261.3540933} provides a different route: rather than incorporating the conditioning signal into the denoiser during training, it steers a frozen diffusion model at sampling time using gradients from an external classifier evaluated on the current noisy state. This decouples the conditioning objective from the generative model and allows the target to be changed without retraining the denoiser.
Transferring this mechanism to discrete graph diffusion, however, is not straightforward. Models such as EDGE~\cite{3618408.3618589} and DiGress~\cite{DBLP:conf/iclr/VignacKSWCF23} sample discrete edge states, so gradients cannot propagate through the sampled graph. Recent guidance mechanisms for discrete diffusion~\cite{ICLR2025_597254dc,journals/corr/abs-2412-10193,dou2026plugandplay} instead modify discrete transition probabilities, but are not designed around the permutation-invariant topology and graph-level class structure that characterize complex networks.

This paper introduces \textbf{ProtoGuide}, a post-hoc guidance framework that closes this gap.
ProtoGuide steers a frozen discrete graph diffusion model toward a target class at sampling time, without retraining any component.
The key idea is a differentiable \textit{soft adjacency matrix} that relaxes the denoiser's discrete per-edge output into a continuous representation, restoring the gradient that classifier guidance requires.
This relaxed graph is embedded by a frozen Siamese network~\cite{48e4505b7f124ee89368447f32b5cf0e}, a weight-sharing architecture trained so that graphs of the same class map to nearby points in an embedding space~\cite{romano2026mmdevaluatinggraphgenerative}, and scored against precomputed class prototypes via a contrastive objective that pulls the current graph toward the target class while pushing it away from competing ones.
The resulting per-edge gradient is modulated by a cosine annealing schedule and injected back into the denoiser output.
The framework is agnostic to the diffusion backbone; only the injection point is model-specific, and we instantiate it for two architecturally different models, EDGE and DiGress (Section~\ref{sec:guidance_framework}).
Because class prototypes are simple mean embeddings computed after training, the guidance target can be respecified at sampling time without retraining the denoiser or the Siamese encoder.
Our few-shot analysis (Appendix~\ref{ss:few_shot_prot}) shows that, for the classes considered here, useful prototype targets can be instantiated from very small support sets, while the directionality test (Appendix~\ref{ss:wrong_random_prot}) shows that the guidance effect depends on the identity of the target prototype rather than on a generic perturbation of the denoising process.

Our contributions are as follows:
\begin{itemize}
    \item \textbf{A differentiable soft adjacency relaxation} of the denoiser's per-edge output that makes gradient-based classifier guidance well defined for discrete graph diffusion, including differentiable relaxations of structural node features (degree, $\chi$-degree, local clustering coefficient) and a stop-gradient treatment of the non-differentiable $k$-core. \looseness=-1
    \item \textbf{A contrastive prototype objective} that replaces the noisy classifier of standard classifier guidance in image settings with a frozen Siamese GNN and class prototypes, enabling post-hoc target specification and allowing target prototypes to be instantiated from small support sets without retraining the generative backbone or Siamese encoder.
    \item \textbf{Backbone-agnostic guidance injection}, instantiated and analyzed on two structurally different discrete diffusion models: direct reverse-step guidance for EDGE and clean-data prediction guidance for DiGress, with model-specific injection points consistent with each backbone's native reverse-process parametrization.
    \item \textbf{An extensive evaluation protocol} on five classes of real-world networks, combining controllability (dual $k$-nearest-neighbor protocols; Section~\ref{sec:results}), structural fidelity (Maximum Mean Discrepancy (MMD)~\cite{JMLR:v13:gretton12a} (Appendix~\ref{app:mmd}), a bootstrap-calibrated Kolmogorov--Smirnov goodness-of-fit test, coverage of the real distribution, and pairwise diversity ratio; Section~\ref{sec:structural}), validation by an architecturally independent classifier, a synthetic-to-real transfer test, and ablations of guidance scale, annealing schedule, contrastive design, prototype directionality, few-shot prototypes, and a Best-of-$N$ selection baseline (Appendix~\ref{app:bestofn}).
\end{itemize}

Empirically, on five classes of real-world networks, ProtoGuide raises macro classification accuracy from $50.7\%$ to $73.5\%$ on EDGE and from $73.6\%$ to $83.8\%$ on DiGress (dynamic $k$-nearest-neighbor~\cite{DBLP:journals/corr/HassanatAAA14}; Section~\ref{sec:results}), with the largest per-class gain on Connectome under EDGE ($11.6\% \to 85.0\%$).
Per-graph coverage remains high in most settings, while distributional effects vary across classes and backbones.
An architecturally independent classifier corroborates the macro-level improvements, and a classifier trained only on guided samples achieves $79.2\%$ macro accuracy on real networks excluded from that classifier's own training, providing evidence that class-discriminative structure transfers from generated to real graphs.
\looseness=-1

\section{Related Work}
\subsection{Graph Generation}

A first family of graph generation methods consists of classical parametric and mechanistic random-graph models, which generate network structure according to predefined stochastic rules rather than learning a generative distribution directly from data.
The Erd\H{o}s--R\'{e}nyi~\cite{erdos59a} model generates graphs randomly by adding each possible edge independently with a fixed probability $p$, yielding a binomial degree distribution and no explicit mechanism for community structure.
The preferential attachment model of Barab\'{a}si--Albert~\cite{Baraba_si_1999} proposed a ``rich-get-richer'' mechanism of graph creation, producing heterogeneous, heavy-tailed degree distributions but without explicitly controlling higher-order properties such as community structure, clustering, or motifs.
The Stochastic Block Model (SBM)~\cite{HOLLAND1983109} captures mesoscopic organization by assigning nodes to groups and specifying different connection probabilities within and between them.

More advanced generative models, like the Lancichinetti-Fortunato-Radicchi (LFR) model~\cite{Lancichinetti_2008}, take into account heavy-tailed degree and community-size distributions through an adjustable mixing parameter, whereas the nonuniform Popularity-Similarity-Optimization (nPSO) model~\cite{Muscoloni_2018} places vertices in the hyperbolic plane with a non-uniform angular distribution to jointly model community structure and clustering.
These models can reproduce increasingly rich combinations of network properties, but their generative mechanisms and target structural characteristics are specified a priori rather than learned as a general empirical graph distribution.

The first deep-learning based methods adopted auto-regressive formulations, i.e., generating graphs sequentially by predicting one node at a time (or in small blocks) and their connections to previously generated nodes.
For instance, GraphRNN~\cite{conf/icml/YouYRHL18} generates graphs by sequentially predicting nodes and their edges using a Recurrent Neural Network (RNN), decomposing the adjacency matrix into a sequence of edge decisions.
GRAN~\cite{3454287.3454670} improves scalability by replacing this fully sequential construction with an attention-based Graph Neural Network (GNN)~\cite{4700287} that generates blocks of adjacency-matrix rows in parallel.
Despite these advances, autoregressive graph generation remains inherently sequential and generally requires choosing a node ordering, which can affect both computational efficiency and the representation being learned.
Variational Autoencoders (VAEs)~\cite{kipf2016variationalgraphautoencoders,simonovsky2018graphvaegenerationsmallgraphs} offered an alternative by learning a continuous latent space from which graphs can be decoded in a single pass.
GraphVAE encodes the input graph into a continuous latent distribution and decodes sampled latent representations into probabilistic graph structures, using graph matching to address node-permutation ambiguity.
\looseness=-1

More recently, diffusion models have become a prominent approach to graph generation, adapting iterative denoising to distributions over graph structure.
EDP-GNN~\cite{pmlr-v108-niu20a} applied score-based diffusion to graph adjacency matrices in continuous space.
GDSS~\cite{Jo2022ScorebasedGM} extended this approach by jointly diffusing node features and adjacency matrices through a system of stochastic differential equations.
DiGress~\cite{DBLP:conf/iclr/VignacKSWCF23}, on the other hand, introduced a fully discrete diffusion process that operates directly on categorical node and edge types, building on discrete denoising diffusion in discrete state space~\cite{3540261.3541637};
a Graph Transformer~\cite{3454287.3455360} predicts clean node and edge states from which the reverse transition is derived.
EDGE~\cite{3618408.3618589} proposed a degree-guided discrete diffusion model that leverages degree sequence information and restricts denoising to active nodes to improve the efficiency of graph generation.
Unlike EDP-GNN and GDSS, which diffuse a continuous relaxation of the adjacency matrix, DiGress and EDGE operate directly on discrete edge states, avoiding the final discretization step required by continuous graph-diffusion approaches; this direct treatment of discrete state spaces has been shown to benefit graph diffusion empirically~\cite{haefeli2023diffusionmodelsgraphsbenefit}.

\subsection{Guided and Conditional Diffusion}
Sohl-Dickstein et al.~\cite{pmlr-v37-sohl-dickstein15} first framed generative modeling as learning to reverse a fixed noising process; Ho et al.~\cite{3495724.3496298} and Song et al.~\cite{DBLP:conf/iclr/0011SKKEP21} later formalized this, respectively, as denoising diffusion probabilistic models and continuous score-based generative models.
In both formulations, a forward process progressively corrupts a clean sample $\mathbf{x}_0$ over $T$ steps, while a learned reverse process generates data by iteratively denoising from $\mathbf{x}_T$ back toward $\mathbf{x}_0$.
At each reverse step, the model estimates the information required to construct or approximate the distribution of $\mathbf{x}_{t-1}$ given $\mathbf{x}_t$.

In the image setting, classifier guidance~\cite{3540261.3540933} steers this reverse process toward a target class $y$, while leaving the diffusion model unchanged, by augmenting the score estimate with the gradient of a classifier $p_\psi(y \mid \mathbf{x}_t)$ trained on noisy images and kept frozen during sampling, where $\mathbf{x}_t$ denotes the current noisy sample:
\begin{equation}
  \hat{\boldsymbol{\epsilon}}_\theta(\mathbf{x}_t, t)
  \;\leftarrow\;
  \hat{\boldsymbol{\epsilon}}_\theta(\mathbf{x}_t, t)
  - \sqrt{1 - \bar\alpha_t}\;\lambda\,
    \nabla_{\mathbf{x}_t} \log p_\psi(y \mid \mathbf{x}_t),
  \label{eq:classifier_guidance}
\end{equation}
where $\lambda$ controls the guidance scale.
Increasing $\lambda$ induces a familiar controllability--fidelity trade-off: stronger guidance can improve alignment with the target condition, but excessive guidance may reduce sample diversity or move samples away from the unguided data distribution~\cite{3540261.3540933}.


Classifier-free guidance~\cite{ho2022classifierfreediffusionguidance} avoids the need for an additional classifier by training the diffusion model both with and without the conditioning signal and interpolating between the two at generation time; it is widely used in conditional image generation, where the conditioning signal is available during training.
Unlike classifier guidance, however, the conditioning mechanism is built into the generative model during training; changing that mechanism or introducing conditioning information not represented during training generally requires adapting or retraining the model.

Extending gradient-based guidance to graph diffusion raises problems that do not arise in the continuous image setting.
Continuous graph-diffusion approaches such as EDP-GNN and GDSS maintain real-valued intermediate states, providing a differentiable representation through which guidance gradients can in principle be propagated.
Discrete approaches such as DiGress and EDGE instead define reverse diffusion processes over finite, discrete edge-state spaces.
Sampling these edge states breaks the differentiable path through the generated graph on which standard classifier guidance relies.

Guidance for discrete diffusion has therefore been formulated directly in terms of the discrete reverse process.
Nisonoff et al.~\cite{ICLR2025_597254dc} derive guided reverse transitions by modifying the process's transition probabilities, while Schiff et al.~\cite{journals/corr/abs-2412-10193} develop a related mechanism that acts directly on discrete transition kernels.
Such methods provide general mechanisms for steering discrete diffusion, but they are not specifically designed around permutation-invariant graph topology or graph-level class structure, nor do they directly provide a differentiable graph representation on which a structural class-discriminative objective can be evaluated.

In the graph domain, conditional generation has been explored primarily via conditioning mechanisms incorporated into the generative model during training.
DiGress, for instance, supports class-conditional generation by concatenating one-hot class labels to the input, so the class-conditioning mechanism must be specified during training;
GDSS incorporates molecular property conditioning into its score network; and Huang et al.~\cite{10.1609/aaai.v37i4.25549} condition molecular-graph diffusion on graph-structure information.
These approaches are effective when the conditioning information is known during training, but offer less flexibility when the desired conditioning objective changes after the generative model has been trained.
Classifier guidance is attractive in this setting precisely because it decouples the guidance objective from the denoiser, provided that an appropriate gradient can be constructed for the discrete graph representation.
\looseness=-1

Several recent methods pursue more flexible inference-time control of graph diffusion.
FreeGress~\cite{10.1007/978-3-031-70359-1_19} extends DiGress with classifier-free guidance for molecular properties, requiring the corresponding conditioning mechanism to be incorporated during training through label dropout.
PRODIGY~\cite{sharma2024diffuse} steers molecular graph diffusion toward target chemical properties and structural hard constraints through projection-based guidance.
GGDiff~\cite{tenorio2025graphguideddiffusionunified} formulates guidance as stochastic optimal control and unifies gradient-based and gradient-free strategies, but operates on GDSS, a continuous SDE-based diffusion backbone, and targets constraint satisfaction such as motifs, fairness, and link prediction rather than class-conditional generation.
Recently, GILC~\cite{dou2026plugandplay} proposed a training-free logit-correction scheme for discrete diffusion that handles differentiable and non-differentiable rewards; it is applied to molecular graphs, DNA/protein sequences, and images.
Together, these works demonstrate growing interest in flexible guidance for graph and discrete diffusion, but class-level post-hoc control of purely topological real-world network generation remains comparatively unexplored.

Classifier guidance ordinarily requires training a classifier on noisy intermediates $\mathbf{x}_t$ at every diffusion timestep, since a classifier trained only on clean data degrades far from $t=0$.
ProtoGuide takes a different route: at each reverse step, it constructs a differentiable soft graph representation from the denoiser output and evaluates its class structure using a model trained on clean graphs. This allows the guidance signal to be defined in a learned graph-embedding space rather than by a timestep-specific noisy-state classifier. We obtain this class-discriminative representation through metric learning, which we review next.

\subsection{Metric Learning and Graph Classification}
\label{sec:metric_learning}
Metric Learning~\cite{NIPS2002_c3e4035a} focuses on building a feature space
such that semantically related inputs should be close to each other, while unrelated
inputs should be as far from each other as possible.
Contrastive objectives~\cite{1640964} learn such spaces by reducing distances between similar pairs while separating dissimilar ones.
Triplet loss~\cite{Schroff_2015} instead enforces a margin between an anchor--positive distance and the corresponding anchor--negative distance.
Siamese Networks~\cite{48e4505b7f124ee89368447f32b5cf0e} implement this principle through weight-sharing encoders that map different inputs into a common embedding space in which their similarity can be evaluated directly.
A closely related idea underlies Prototypical Networks~\cite{3294996.3295163}: each class is represented by the mean embedding of its labeled support samples, and new observations are classified according to their distance from these class prototypes.
This representation is particularly attractive in low-data settings because a prototype can be estimated from a small support set and updated without retraining the embedding model, provided that the learned embedding generalizes to the target class.
\looseness-1

In the graph domain, applying these metric-learning objectives requires an encoder that maps a graph to a fixed-size embedding, and Graph Neural Networks (GNNs)~\cite{4700287} are widely used for this purpose.
GNN architectures such as the Graph Convolutional Network (GCN)~\cite{kipf2017semisupervisedclassificationgraphconvolutional}, Graph Attention Network (GAT)~\cite{veličković2018graphattentionnetworks}, Graph Isomorphism Network (GIN)~\cite{DBLP:conf/iclr/XuHLJ19}, and GraphSAGE~\cite{3294771.3294869} compute node-level embeddings through iterative message passing, which are then aggregated into a graph-level representation via a pooling operator.
Feeding these representations into a Siamese framework~\cite{romano2026mmdevaluatinggraphgenerative} extends metric learning to graph classification and similarity tasks.

For the present setting, this combination provides both ingredients required for prototype-based guidance: a differentiable graph-level representation and an embedding space in which proximity to a target network class can be quantified.

\section{Background: Discrete Graph Diffusion Backbones}
\label{sec:background}
Related Work positions EDGE and DiGress among alternative graph-generative paradigms; this section instead fixes the notation and mechanics of their forward and reverse processes on which the guidance framework of Section~\ref{sec:guidance_framework} builds.

\subsection{EDGE: Efficient and Degree-Guided Graph Generation}
Let $G = (\mathcal{V}, \mathcal{E})$ be an undirected graph with $N = |\mathcal{V}|$ nodes.
We represent $G$ by the vector of its upper-triangular edge states $\mathbf{x}_0 \in \{0,1\}^{M}$, with $M = N(N{-}1)/2$.
EDGE~\cite{3618408.3618589} defines a \textit{forward} diffusion process that progressively removes edges, driving the graph toward the empty state as $t\!\to\!T$; the corruption factorizes over edges:
\begin{equation}
  q(\mathbf{x}_t \mid \mathbf{x}_0)
  \;=\;
  \prod_{(i,j)}
    q\!\left(x_t^{ij} \mid x_0^{ij}\right),
  \label{eq:forward}
\end{equation}
where $x_t^{ij} \in \{0,1\}$ is the state of edge $(i,j)$ at time~$t$ and the per-edge marginal interpolates between the clean state and the empty graph. Generation consequently starts from an empty graph endowed with a target degree sequence sampled from the empirical distribution of the training data.

The \textit{reverse} process is parameterized by a GNN denoiser $p_\theta$, which also factorizes over edges:
\begin{equation}
  p_\theta(\mathbf{x}_{t-1} \mid \mathbf{x}_t)
  \;=\;
  \prod_{(i,j)}
    p_\theta\!\left(x_{t-1}^{ij} \mid \mathbf{x}_t\right).
  \label{eq:reverse}
\end{equation}
For each candidate edge $(i,j)$ the network outputs a two-dimensional log-probability vector:
\begin{equation}
  \logp_{ij}(t) = \bigl[ \ell_{ij}^{(0)}(t),\; \ell_{ij}^{(1)}(t) \bigr] \in \R^2, 
  \qquad 
  \ell_{ij}^{(e)}(t) = \log p_\theta\!\left(x_{t-1}^{ij} = e \mid \mathbf{x}_t\right).
  \label{eq:logprobs}
\end{equation}

At each reverse step EDGE denoises only a subset of \textit{active} nodes, drawn from a degree-based posterior, and samples edge states only for pairs whose endpoints are both active; every remaining pair keeps its current state. For an active pair, the next state is obtained by converting the log-probabilities into a Bernoulli probability via the softmax:
\begin{equation}
  x_{t-1}^{ij} \sim \mathrm{Bernoulli}\!\bigl(\text{softmax}(\logp_{ij}(t))_1\bigr),
  \label{eq:sampling}
\end{equation}
where $\text{softmax}(\logp_{ij}(t))_1$ is the probability assigned to the edge being present ($e=1$).
The explicit degree conditioning and the active-node restriction are both properties that ProtoGuide must respect when injecting guidance; we return to them in Section~\ref{sec:injection}.

\subsection{DiGress: Discrete denoising diffusion for graph generation}
DiGress~\cite{DBLP:conf/iclr/VignacKSWCF23} represents the graph in dense form: a node-type matrix $\mathbf{X} \in \{0,1\}^{N \times d_X}$ and an edge-type tensor $\mathbf{E} \in \{0,1\}^{N \times N \times d_E}$, where $d_X$ and $d_E$ are the number of node and edge classes, respectively.
For plain graphs without node or edge attributes, $d_X =1$ and $d_E =2$ (edge absent or present).
All graphs in a batch are zero-padded to a common maximum number of nodes, with a boolean node mask $\mathbf{m} \in \{0,1\}^N$ indicating valid nodes.
The \textit{forward} process independently corrupts each node and edge type via a discrete Markov chain governed by transition matrices:
\begin{equation}
  q(\mathbf{X}_t, \mathbf{E}_t \mid \mathbf{X}_0, \mathbf{E}_0)
  \;=\;
  \prod_{i}\, q(X_t^i \mid X_0^i)
  \;\prod_{i<j}\, q(E_t^{ij} \mid E_0^{ij}),
  \label{eq:digress_forward}
\end{equation}
where each per-element marginal is defined by a transition matrix $\bar{\mathbf{Q}}_t$ that interpolates between the identity and a limiting noise distribution (at $t=0$ and $t=T$, respectively).
DiGress uses a \textit{cosine} noise schedule to control the rate of corruption.

The \textit{reverse} process is parameterized by a Graph Transformer denoiser~$p_\theta$.
DiGress predicts the clean data distribution:
\begin{equation}
  \hat{p}_\theta(\mathbf{X}_0, \mathbf{E}_0 \mid \mathbf{X}_t, \mathbf{E}_t)
  = \prod_{i} \hat{p}_\theta(X_0^i \mid \mathbf{X}_t, \mathbf{E}_t) 
    \prod_{i<j} \hat{p}_\theta(E_0^{ij} \mid \mathbf{X}_t, \mathbf{E}_t)
  \label{eq:digress_x0pred}
\end{equation}
For each edge $(i,j)$ the network outputs a two-dimensional probability vector over edge types:
\begin{equation}
  \hat{\mathbf{p}}_{ij}(t) = \bigl[\hat{p}_{ij}^{(0)}(t),\;\hat{p}_{ij}^{(1)}(t)\bigr] \in [0,1]^2, 
  \qquad 
  \hat{p}_{ij}^{(e)}(t) = \hat{p}_\theta\!\left(E_0^{ij} = e \mid \mathbf{X}_t, \mathbf{E}_t\right).
  \label{eq:digress_pred}
\end{equation}
The one-step reverse distribution $p_\theta(\mathbf{E}_{t-1} \mid \mathbf{E}_t)$ is then obtained via Bayes' rule using the forward transition matrices:
\begin{equation}
  p_\theta(E_{t-1}^{ij} = e \mid \mathbf{E}_t)
  \propto \sum_{e_0} \hat{p}_{ij}^{(e_0)}(t)\, q(E_{t-1}^{ij} = e \mid E_0^{ij} = e_0,\, E_t^{ij})
  \label{eq:digress_posterior}
\end{equation}
where the posterior $q(E_{t-1}^{ij} \mid E_0^{ij},\, E_t^{ij})$ is computed in closed form from the noise schedule transition matrices $\mathbf{Q}_t$ and $\bar{\mathbf{Q}}_s$ (with $s=t{-}1$).
The next edge state is then sampled from this distribution at each reverse step.

\section{Method}

\begin{figure}[ht!]
    \centering
    \includegraphics[width=\textwidth]{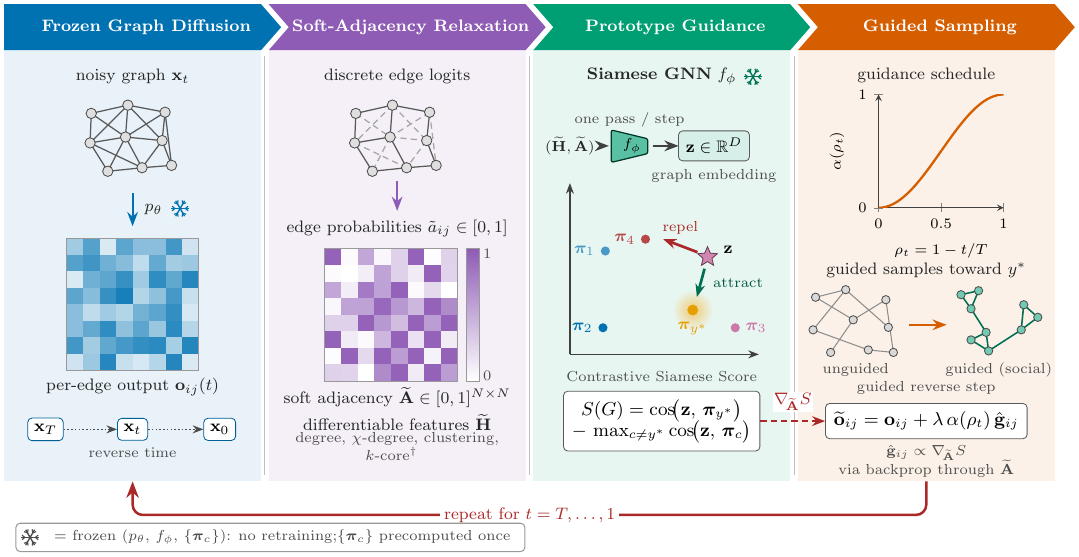}
    \caption{\textbf{ProtoGuide pipeline.}
    At each step~$t$ of the reverse diffusion process (running from $t=T$, pure noise, down to $t=1$), the frozen denoiser~$p_\theta$ produces per-edge outputs $\mathbf{o}_{ij}(t)$ (Panel~1).
    These are converted into a differentiable soft adjacency matrix $\widetilde{\mathbf{A}} \in [0,1]^{N \times N}$, together with structural node features (Panel~2).
    A frozen Siamese GNN~$f_\phi$ embeds this relaxed graph into $\mathbf{z} \in \mathbb{R}^D$ and computes a contrastive Siamese score $S(G)$, which pulls~$\mathbf{z}$ toward the target-class prototype~$\proto_{y^*}$ and pushes it away from the competing class prototype (Panel~3).
    Backpropagating the contrastive score through the soft adjacency construction yields per-edge gradients $\mathbf{g}_{ij}=\partial S/\partial\mathbf{o}_{ij}(t)$, which are $\ell_2$-normalized per-edge to obtain $\hat{\mathbf{g}}_{ij}$, scaled by a guidance scale $\lambda$ and a cosine annealing schedule~$\alpha(\rho_t)$, and injected into the denoiser output to steer generation toward the target class (Panel~4; the Social class is shown here as an illustrative example). The annealing schedule suppresses guidance early in generation (large $t$, high noise) and strengthens it later.
    Denoiser, Siamese GNN, and class prototypes all remain frozen throughout: no component is retrained.
    ${}^\dagger$\,The $k$-core feature is non-differentiable; it is computed on the hard-thresholded graph and treated as a constant (stop-gradient) during backpropagation.}
    \label{fig:pipeline}
\end{figure}

Our goal is to steer a frozen, already-trained discrete diffusion model toward a target class $y^*$ at every reverse step, without modifying or retraining the denoiser.
In the continuous (i.e., image) setting, classifier guidance achieves this by adding the gradient $\nabla_{\mathbf{x}_t} \log p_{\psi}(y \mid \mathbf{x}_t)$ to the score estimate (Eq.~\ref{eq:classifier_guidance}).
Adapting this mechanism to discrete graph diffusion faces a structural obstacle: the reverse process samples discrete edge states, which break the differentiable path through the sampled graph required to compute the guidance gradient.

ProtoGuide addresses this obstacle with three core components.
(i)~\textbf{Class prototypes}, computed once from a frozen Siamese GNN, define the guidance target. Once the Siamese encoder is trained, each prototype is estimated solely from embeddings of its own class, so prototype construction itself does not depend on the relative frequencies of the other classes. Prototypes are also cheap to recompute and allow the guidance target to be respecified at sampling time without retraining either the denoiser or the encoder.
(ii)~A differentiable \textbf{soft adjacency matrix}, built from the per-edge output of the denoiser, provides a continuous approximation of the discrete graph through which gradient can flow.
(iii)~A \textbf{cosine annealing} schedule that suppresses guidance during the early noisy timesteps, where the soft adjacency carries little structural signal, and gradually increases it as the graph structure consolidates.
The soft adjacency provides the continuous path through which guidance is differentiated: degree, $\chi$-degree, and clustering are constructed as differentiable functions of it, while the non-differentiable $k$-core feature is held fixed during the backward pass.
Every component described in this section is shared identically across both diffusion models, EDGE and DiGress; the only model-specific element is the injection point, i.e., which internal quantity of the denoiser receives the gradient update (Section~\ref{sec:injection}).
The ProtoGuide pipeline is summarized in Figure~\ref{fig:pipeline} and Algorithm~\ref{alg:protoguide} in Appendix~\ref{app:algorithm}.

\subsection{Siamese Encoder}
The class-discriminative signal used throughout this section comes from a frozen Siamese network $\backbone$: two identical, weight-sharing branches that embed a graph into a fixed-size vector, trained so that graphs of the same class land close together and graphs of different classes land far apart.
In our experiments this encoder is instantiated as a Graph Attention Network (GAT)~\cite{veličković2018graphattentionnetworks}, selected via a grid search over architecture and hyperparameters (Section~\ref{sec:exp_setup}, Supplementary Information~\ref{app:encoder_grid}).

\paragraph{Architecture}
we employ an $L$-layer GAT encoder with multi-head attention. Each intermediate layer uses $H$ attention heads and hidden dimension $d_h$; the final layer uses a single head with output dimension $d_o$.
A Jumping Knowledge~\cite{pmlr-v80-xu18c} concatenation scheme aggregates representations across all depths: the outputs of every GAT layer, together with the raw input, are concatenated per node:
\begin{equation}
  \mathbf{r}_i = \bigl[\mathbf{o}_i^{(0)} \;\|\; \mathbf{o}_i^{(1)} \;\|\; \cdots \;\|\; \mathbf{o}_i^{(L)}\bigr] \in \R^{D}, 
  \qquad 
  D = F + (L{-}1)\,d_h H + d_o,
  \label{eq:skipcat}
\end{equation}
where $\|$ denotes concatenation, $\mathbf{o}_i^{(0)} \in \R^F$ is the input node-feature vector (the $F=4$ topological descriptors defined in Section~\ref{sec:guidance_framework}), and $\mathbf{o}_i^{(\ell)}$ for $\ell = 1,\dotsc,L$ are the GAT layer outputs. An attentional aggregation module with a two-layer gate MLP pools the node representations into a single graph-level vector:
\begin{equation}
  \emb = \frac{\mathbf{u}}{\norm{\mathbf{u}}_2},
  \qquad
  \mathbf{u} = \sum_{i=1}^{N} \sigma\!\bigl(g(\mathbf{r}_i)\bigr)\,\mathbf{r}_i \;\in\; \R^{D},
  \label{eq:pool}
\end{equation}
where $g\!:\!\R^{D}\!\to\!\R$ is the gate network and $\sigma$ the sigmoid. The final $\ell_2$ normalization ensures $\norm{\backbone(G)}_2 = 1$ for every graph $G$, so that cosine similarity reduces to a dot product (Eq.~\ref{eq:score}). Because Eqs.~\ref{eq:skipcat}--\ref{eq:pool} are differentiable functions of the node features $\featmat$, applying $\backbone$ to the relaxed features computed from the soft adjacency during guidance requires no change to the architecture.

\paragraph{Training} $\backbone$ is trained once on real graphs with triplet margin loss~\cite{Schroff_2015} and then frozen; exact hyperparameters (layers, heads, dimensions, margin, learning rate) are reported in Section~\ref{sec:exp_setup}.

\subsection{ProtoGuide Generation Framework}
\label{sec:guidance_framework}
\paragraph{Class Prototypes}
before generation, we embed all training graphs with the frozen encoder and compute an $\ell_2$-normalized prototype for each class.
For class~$c$ with training graphs $\mathcal{D}_c = \{G_1^c,\dotsc,G_{N_c}^c\}$:
\begin{equation}
  \proto_c = \frac{\bar{\emb}_c}{\norm{\bar{\emb}_c}_2}, 
  \qquad 
  \bar{\emb}_c = \frac{1}{N_c}\sum_{i=1}^{N_c} \backbone(G_i^c).
  \label{eq:prototype}
\end{equation}
Since each $\backbone(G_i^c)$ has unit norm, the normalized mean prototype also has unit norm.
We denote the target prototype $\proto_{y^*}$ and the set of negative prototypes $\{\proto_c : c \neq y^*\}$.
All prototypes are computed once and held fixed during generation.

Retargeting guidance to another available class only requires selecting its stored prototype, while defining a target from a new support set requires computing its mean embedding.
We perform two sanity checks on this prototype-based formulation: wrong/random-prototype and few-shot-prototype analyses, reported in Appendix~\ref{ss:wrong_random_prot}--\ref{ss:few_shot_prot}.

\paragraph{Guidance Objective}
given a target class $y^*$, we define a contrastive Siamese score:
\begin{equation}
  S(G)
  = \underbrace{\scos{\backbone(G)}{\proto_{y^*}}}_{%
      \text{attract to target}}
  \;-\;
    \underbrace{\max_{c \,\neq\, y^*}\;
      \scos{\backbone(G)}{\proto_c}}_{%
      \text{repel from nearest competitor}},
  \label{eq:score}
\end{equation}
where $\cos(\mathbf{u}, \mathbf{v}) = \mathbf{u}^\top\mathbf{v} /(\norm{\mathbf{u}}_2\norm{\mathbf{v}}_2)$ is the cosine similarity.
The positive term drives the embedding toward the target prototype, while the negative term penalizes similarity to the nearest competing prototype and thereby encourages inter-class separation.
We ablate this design choice against an attraction-only variant in Appendix~\ref{ss:attraction_vs_contrastive}, where its contribution is class-dependent.
\looseness=-1

\paragraph{Soft Adjacency from Edge Probabilities}
the score $S(G)$ is a function of a discrete graph, but to obtain a gradient we need it to depend on a continuous quantity.
We therefore construct a differentiable relaxation of the denoiser output using its current edge-presence probabilities.
At each reverse step $t$, the denoiser produces a per-edge output from which we extract the probability that edge $(i,j)$ is present, $\softaij(t) \in [0,1]$.
The exact derivation of $\softaij$ from the denoiser output differs between models (Section~\ref{sec:injection}); in all cases it is a differentiable function of that output.
From these probabilities we build a symmetric matrix $\softadj \in[0,1]^{N \times N}$ with $\widetilde{A}_{ij}=\widetilde{A}_{ji}=\softaij$.
This relaxed adjacency provides the continuous representation through which gradient-based guidance can be applied to the contrastive Siamese score; its precise interpretation depends on the backbone and is detailed in Section~\ref{sec:injection}.

\paragraph{Node Features}
\label{sec:node_features}
to evaluate the contrastive Siamese score on the soft adjacency, we first need per-node features.
The Siamese encoder characterizes each node by four topological descriptors: degree, $\chi$-degree, local clustering coefficient, and $k$-core number. Degree, clustering, and $k$-core are normalized to $[0,1]$ on binary graphs; the $\chi$-degree is a dispersion statistic and is not bounded above.
Together, these metrics cover structural properties at complementary scales: degree measures local connectivity, $\chi$-degree indicates how much a vertex differs from the mean degree (for instance, highlighting hubs), clustering coefficient measures local cohesion, and $k$-core number represents meso-level cohesion.
On a discrete graph $G$ they are defined as:
\begin{align}
    h_i^{(0)} &= \frac{\deg(i)}{\max_{j} \deg(j)},
    \label{eq:chi}\\
    h_i^{(1)} &= \frac{\left(h_i^{(0)} - \bar{d}\right)^2}{\bar{d}}, \quad \bar{d} = \frac{1}{N} \sum_{j=1}^{N} h_j^{(0)}, \nonumber
\end{align}
\begin{align}
    h_i^{(2)} &= \frac{2 \, |\{\{j, k\} \subset \mathcal{N}(i) : \{j,k\} \in \mathcal{E}\}|}{\deg(i)\bigl(\deg(i) - 1\bigr)},
    \label{eq:clustering}\\
    h_i^{(3)} &= \frac{\mathrm{core}(i)}{\max_{j} \mathrm{core}(j)}, \nonumber
\end{align}
where $\mathrm{core}(i)$ is the degeneracy index: the largest $k$ such that node $i$ belongs to a subgraph in which every node has degree $\geq k$.

\paragraph{Differentiable Features}
the node features above are defined on discrete graphs; during guided generation, the guidance gradient $\nabla_{\!\softadj}S$ is obtained by back-propagating through the entire ProtoGuide pipeline $\softadj \to \mathbf{h} \to \text{GAT} \to S$,
so the feature construction must provide a differentiable path from $\softadj$ to the score.
Three of the four node features admit closed-form differentiable relaxations; the $k$-core feature is instead treated as a constant during backpropagation.

\begin{align}
    \softdeg_i &= \sum_{j=1}^N \softadj_{ij},
    &
    h_i^{(0)} &= \frac{\softdeg_i}{\max_j \softdeg_j}
\end{align}
\begin{align}
    \bar{d} &= \frac{1}{N} \sum_{j=1}^N h_j^{(0)},
    &
    h_i^{(1)} &= \frac{\left(h_i^{(0)} - \bar{d}\right)^2}{\bar{d}+\epsilon}
\end{align}
\begin{equation}
  h_i^{(2)}
  = \frac{
      \displaystyle\sum_{j=1}^{N}
        \bigl(\softadj^2\bigr)_{\!ij}\;\widetilde{A}_{ij}
    }{
      \softdeg_i\,(\softdeg_i - 1) + \epsilon
    },
  \label{eq:soft_clust}
\end{equation}
where $\softadj^2 = \softadj\softadj$ is the matrix product and $\epsilon$ is a small constant for numerical stability.
The numerator counts soft triangles: $\sum_{j,k}\widetilde{A}_{ik}\,\widetilde{A}_{kj}\,\widetilde{A}_{ij}$ sums the product of edge probabilities around every triangle involving node $i$, counting each triangle twice.
The denominator $\softdeg_i(\softdeg_i-1)$ is the direct algebraic extension of the binary term $d_i(d_i-1)$ and recovers the standard clustering coefficient when the adjacency is binary, which is the regime in which the encoder was trained.
For fractional adjacencies, this algebraic relaxation is a surrogate rather than a bounded clustering coefficient and need not remain in $[0,1]$, in particular when $\softdeg_i<1$.
It nevertheless recovers the training-time feature exactly when the adjacency is binary.
The per-edge normalization of Section~\ref{sec:injection} ensures that the $\ell_2$ norm of the injected update remains bounded by $\lambda\,\alpha(\rho_t)$, irrespective of the magnitude of the resulting feature gradient.


For the non-differentiable $k$-core feature, we compute it exactly on the hard graph obtained by thresholding the soft adjacency at $0.5$:
\begin{align}
    \hat{A}_{ij} &= \mathbf{1}[\softadj_{ij} > 0.5],
    &
    h_i^{(3)} &= \frac{\mathrm{core}(i)}{\max_{j} \mathrm{core}(j)}
\end{align}
In the backward pass, $h_i^{(3)}$ is treated as a constant with zero gradients, so only the three differentiable features carry the gradient signal.
The four features are stacked row-wise into the node-feature matrix
$\featmat \in \R^{N \times 4}$, whose $i$-th row is
$
    \featmat_{i,:} = \mathbf{h}_i^\top = \bigl[h_i^{(0)},\; h_i^{(1)},\; h_i^{(2)},\; h_i^{(3)}\bigr].
$
\paragraph{Message Passing on the Hard Graph}
while node features are computed from the soft adjacency, we retain a discrete topology for GNN message passing to remain compatible with the frozen Siamese encoder, which was trained on ordinary binary graphs.
We obtain the edge index by thresholding the soft adjacency: $\mathbf{E}_{\mathrm{hard}} = \{(i,j) : \softaij > 0.5\}$.
The role of $\softadj$ is twofold: it provides continuous values for feature computation (where gradients are needed) and a discrete topology for message passing (where they are not).
Because thresholding is a non-differentiable operation, the message-passing topology is fixed in the backward pass.
Importantly, gradients do flow through the full soft adjacency matrix $\softadj$ via the node features: the gradient $\partial S / \partial \mathbf{o}_{ij}$ captures how changing each edge probability reshapes the degree, clustering, and $\chi$-degree distributions of the graph, which in turn shift the Siamese embedding toward or away from the target prototype.
The guidance therefore steers the edge probability distribution toward class-discriminative structural patterns, even though the message-passing index used for the GNN forward pass is frozen during backpropagation.

\paragraph{Per-Edge Gradient Guidance}
with features $\featmat$ and edge index $\mathbf{E}_{\mathrm{hard}}$, we forward the graph through 
the frozen Siamese encoder to obtain the embedding $\emb = \backbone(\featmat, \mathbf{E}_{\mathrm{hard}}) \in \R^D$.
We compute the contrastive Siamese score $S(G)$ (Eq.~\ref{eq:score}) and backpropagate through the differentiable soft adjacency path to obtain per-edge gradients with respect to the model-specific denoiser output $\mathbf{o}_{ij}(t)$ for edge $(i,j)$ at reverse step $t$:
\begin{equation}
    \mathbf{g}_{ij} = \frac{\partial S}{\partial \mathbf{o}_{ij}(t)} \in \R^{d_E}
\end{equation}
To make the guidance scale $\lambda$ interpretable and consistent across different graph sizes and timesteps, we $\ell_2$-normalize each per-edge gradient vector:
\begin{equation}
    \hat{\mathbf{g}}_{ij} = \frac{\mathbf{g}_{ij}}{\norm{\mathbf{g}_{ij}}_2 + \epsilon}
\end{equation}

In both backbones, let $\mathbf{o}_{ij}(t)=\bigl(o^{(0)}_{ij}(t),o^{(1)}_{ij}(t)\bigr)$ denote the logits for edge absence and presence, respectively, with $\softaij=\mathrm{softmax}(\mathbf{o}_{ij}(t))_1$.
Neglecting the numerical-stability term $\epsilon$, the chain rule gives
\begin{equation}
\frac{\partial S}{\partial \mathbf{o}_{ij}(t)} = S'(\softaij)\, \softaij(1-\softaij)\, (-1,1)
\end{equation}
where $(-1,1)$ is the binary-softmax direction corresponding to decreasing the edge-absence logit and increasing the edge-presence logit.
Per-edge normalization removes the magnitude contributed by both $S'(\softaij)$ and the softmax Jacobian, retaining only the two-logit direction
\begin{equation}
\operatorname{sign}\!\bigl(S'(\softaij)\bigr) \frac{(-1,1)}{\sqrt{2}}.
\end{equation}
The magnitude of the class-discriminative signal is therefore discarded in favor of bounded edge-wise steering. For each edge, the resulting two-logit update has $\ell_2$ norm approximately $\lambda\,\alpha(\rho_t)$, and is further attenuated in DiGress when the two orientations of a pair disagree (Section~\ref{sec:injection}). This makes $\lambda$ the primary control of the update magnitude independently of the raw gradient magnitude and score scale, while its timestep dependence is controlled explicitly by $\alpha(\rho_t)$.

\paragraph{Cosine Guidance Schedule}

we modulate the guidance strength over the reverse process according to the reliability of the current soft adjacency.
At early reverse steps, the graph is still mostly noise, so the soft adjacency carries little structural information and the resulting gradients are unreliable.
As generation progresses and the graph structure solidifies, the soft adjacency becomes increasingly informative and guidance can steer the topology more effectively.
We define the generation progress as $\rho_t = 1 - t/T \in [0,1-1/T]$ for $t=T,\dotsc,1$ ($0$~=~the initial noisy state, with $\rho_t$ approaching $1$ near the final reverse step) and apply a cosine annealing factor to the guidance scale:

\begin{equation}
  \alpha(\rho_t) = \frac{1 - \cos(\pi\,\rho_t)}{2}
  \label{eq:annealing}
\end{equation}
This function equals $0$ at $\rho_t = 0$ (no guidance at pure noise), rises slowly through the early noisy steps and approaches its maximum strength near the final sample.
Additionally, guidance is skipped entirely for the first $5\%$ of reverse steps (i.e., $\rho_t < 0.05$), where the soft adjacency is dominated by noise and gradient signals are unreliable.
The cosine shape, combined with this early-step cutoff, suppresses guidance when the edge probabilities carry little structural information and gradually increases it as they become reliable.
We ablate this choice against constant and linear schedules in Appendix~\ref{ss:annealing}.

\subsection{Model-Specific Guidance Injection}\label{sec:injection}
The framework above is agnostic to the diffusion model: given a denoiser output, it constructs a soft adjacency, computes a contrastive score, and returns normalized per-edge gradients.
The injection point (the quantity that receives the gradient update) must, however, match the model's parametrization. We describe the two instantiations for both EDGE and DiGress below.
\looseness=-1

\subsubsection{EDGE: Reverse-Step Distribution Injection}

guidance is injected directly into the one-step reverse distribution $p_\theta(x_{t-1}^{ij} \mid \mathbf{x}_t)$, which EDGE parametrizes via log-probabilities $\logp_{ij} \in \R^2$ (Eq.~\ref{eq:logprobs}).
The soft adjacency is obtained by extracting the edge-presence probability:
\begin{equation}
  \softaij(t) = \mathrm{softmax}\!\bigl(\logp_{ij}(t)\bigr)_1 = \frac{\exp\!\bigl(\ell_{ij}^{(1)}(t)\bigr)}{\exp\!\bigl(\ell_{ij}^{(0)}(t)\bigr) + \exp\!\bigl(\ell_{ij}^{(1)}(t)\bigr)}
  \label{eq:edge_softadj}
\end{equation}
The denoiser output for the gradient computation is $\mathbf{o}_{ij}(t) = \logp_{ij}(t)$, so the guided log-probabilities are:
\begin{equation}
    \widetilde{\mathbf{o}}_{ij}(t) 
    = \mathbf{o}_{ij}(t) +
    \lambda \,\alpha(\rho_t)\, \hat{\mathbf{g}}_{ij},
    \label{eq:edge_update}
\end{equation}
followed by renormalization to obtain a valid distribution for sampling $x_{t-1}^{ij}$. Because the update acts on the reverse-step distribution directly, each sampling step produces an edge state that is immediately influenced by the guidance signal.
Two details are specific to EDGE's active-node mechanism.
First, the denoiser emits log-probabilities only for active pairs, so the soft adjacency is assembled by writing those predictions into the current edge state: active pairs contribute relaxed probabilities $\softaij(t)$, while all remaining pairs contribute their present binary value.
The Siamese score is therefore evaluated on the whole graph, but the update of Eq.~\eqref{eq:edge_update} is applied only to active-pair logits; non-active pairs are left untouched at that step.
Second, active-node selection at step $t$ is performed by EDGE's degree-based posterior before guidance is computed, so ProtoGuide never changes which nodes are active at that step.
The two mechanisms are nonetheless coupled across timesteps: the edges sampled under guidance alter the degree sequence that conditions active-node selection at subsequent steps, so the interaction is sequential through the sampled state rather than analytic within a single update.

\subsubsection{DiGress: Clean-Data Prediction Injection}

guidance is instead injected into the clean-data prediction $\hat{\mathbf{p}}_{ij}(t) \in [0,1]^{d_E}$ (Eq.~\ref{eq:digress_pred}), before the reverse-step distribution is derived via Bayes' rule (Eq.~\ref{eq:digress_posterior}).
The reverse-step posterior is obtained from the clean-data prediction through the forward transition matrices $\mathbf{Q}_t$ and $\bar{\mathbf{Q}}_{t-1}$ (Eq.~\ref{eq:digress_posterior}); perturbing it directly could produce a reverse-step distribution that no longer corresponds to any clean-data prediction under this construction.
Perturbing the clean-data prediction instead lets the posterior propagate the guidance through the same Bayesian mechanism used in unguided sampling.
The soft adjacency is derived from the predicted edge-presence probability:
\begin{equation}
  \softaij(t) = \hat{p}_{ij}^{(1)}(t),
  \label{eq:digress_softadj}
\end{equation}
and the predicted probabilities are converted to log-space for the gradient computation: $\mathbf{o}_{ij}(t) = \log \hat{\mathbf{p}}_{ij}(t)$, mirroring the EDGE formulation (Eq.~\ref{eq:edge_update}).
The guided log-prediction is:
\begin{equation}
  \widetilde{\mathbf{o}}_{ij}(t)
  = \log \hat{\mathbf{p}}_{ij}(t)
    + \lambda\, \alpha(\rho_t)\, \hat{\mathbf{g}}_{ij},
  \label{eq:digress_update}
\end{equation}
followed by softmax renormalization $\widetilde{\hat{\mathbf{p}}}_{ij}(t) = \operatorname{softmax}\!\bigl(\widetilde{\mathbf{o}}_{ij}(t)\bigr)$ to recover a valid probability vector.
The posterior distribution $p_\theta(E_{t-1}^{ij} \mid E_t^{ij})$ is then computed from the modified prediction $\widetilde{\hat{\mathbf{p}}}_{ij}(t)$ via Eq.~\eqref{eq:digress_posterior}, preserving the noise-schedule constraints.
Because DiGress operates in dense matrix format ($\mathbf{E} \in \R^{N \times N \times d_E}$), the raw gradient tensor may not respect the symmetry $\mathbf{g}_{ij} = \mathbf{g}_{ji}$ required for undirected graphs.
We therefore symmetrize the normalized gradient by averaging the two orientations before applying the update ($\hat{\mathbf{g}}_{ij} \leftarrow \tfrac{1}{2}(\hat{\mathbf{g}}_{ij} + \hat{\mathbf{g}}_{ji})$).
This averaging preserves the full guidance magnitude $\lambda\,\alpha(\rho_t)$ when the two orientations point in the same direction and attenuates the update when they disagree.

\subsection{Relation to Classifier Guidance}

ProtoGuide follows the central principle of classifier guidance (steering a frozen denoiser at sampling time through the gradient of a target-dependent objective) but differs from its standard probabilistic formulation in two respects.
In standard classifier guidance, the update is derived from $\nabla_{\mathbf{x}}\log p_\psi(y\mid\mathbf{x})$, whereas ProtoGuide uses a contrastive cosine-similarity score defined in the Siamese embedding space.
In addition, ProtoGuide $\ell_2$-normalizes each per-edge gradient, controlling the update magnitude through $\lambda\,\alpha(\rho_t)$ rather than retaining the raw gradient magnitude.
Both EDGE and DiGress apply this update in log-space (for DiGress, predicted probabilities are first converted to log-space; cf.\ Eq.~\ref{eq:digress_update}).
Despite these differences, the operational mechanism remains the same: a class-discriminative gradient modifies the denoiser output during sampling, and the directionality experiment in Appendix~\ref{ss:wrong_random_prot} supports that the resulting steering depends on the selected target prototype.

\subsection{Computational Cost of ProtoGuide}
\label{p:complexity}
We separate the cost of ProtoGuide's framework-level operations from that of the specific node features used by our Siamese encoder (Section~\ref{sec:guidance_framework}).

For a graph with $N$ nodes ($M=\binom{N}{2}$ candidate edges) and a Siamese GNN with $L$ layers and hidden dimension $d$, each reverse diffusion step adds three main operations:
(i)~soft adjacency construction, which costs $\mathcal{O}(N^2)$;
(ii)~a forward--backward pass through the Siamese GNN, whose GAT message-passing layers cost $\mathcal{O}(L|\mathcal{E}_{\mathrm{hard}}|d)$;
and (iii)~per-edge gradient normalization and injection, which costs $\mathcal{O}(N^2)$.
The GNN term becomes $\mathcal{O}(LN^2d)$ in the dense case and $\mathcal{O}(LNd)$ when the thresholded graph has bounded average degree.
Automatic differentiation yields all per-edge gradients in a single forward--backward pass, avoiding the $\mathcal{O}(N^2)$ separate model evaluations that a naive finite-difference implementation would require.
Excluding feature construction, the resulting per-step overhead is therefore $\mathcal{O}(N^2 + L|\mathcal{E}_{\mathrm{hard}}|d)$.

The node features used by our Siamese encoder (degree, $\chi$-degree, clustering coefficient, and $k$-core) are specific to the classifier adopted in our experiments rather than requirements of the ProtoGuide framework itself.
Alternative differentiable feature sets could in principle be used with a suitably trained classifier, although their effect on guidance quality is not evaluated here.
With the current features, degree and $\chi$-degree cost $\mathcal{O}(N^2)$.
The differentiable clustering coefficient (Eq.~\ref{eq:soft_clust}) involves a dense matrix product $\softadj^2$ and costs $\mathcal{O}(N^3)$, while $k$-core is computed on the hard-thresholded graph in $\mathcal{O}(N+|\mathcal{E}|)$ without gradient.
The clustering feature is therefore the sole asymptotic bottleneck in the current guidance feature set.

Over $T$ reverse steps, the resulting guidance cost is $\mathcal{O}(TN^3)$, added on top of the denoiser's generation cost.
This overhead is incurred only at sampling time; no retraining of either the denoiser or the Siamese GNN is required.
Appendix~\ref{app:runtime} reports the corresponding wall-clock measurements.

\subsection{Evaluation Protocol}
\label{sec:eval_protocol}
We assess the quality of generated graphs in terms of both controllability (does the generator produce graphs of the requested class?) and structural fidelity (are those graphs structurally realistic?).
Our per-class corpora are small: after $N_{\max}$ filtering, some DiGress classes retain as few as $52$ training graphs (Table~\ref{t:digress_setup}; Appendix~\ref{app:model_config}).
For this reason, the $k$-NN, coverage, Kolmogorov--Smirnov, and MMD evaluations compare generated graphs against the full training reference set rather than a held-out split, since holding out even a modest fraction would leave some classes with too few reference graphs to estimate distances or distributions meaningfully.
We complement these reference-set evaluations with the synthetic-to-real transfer test in Appendix~\ref{app:gen_only}, where a classifier trained exclusively on ProtoGuide-generated graphs is evaluated on real graphs held out from that classifier's training.
These real graphs are not disjoint from the generators' training data or from prototype computation, so the experiment measures transfer of class structure from generated to real graphs rather than generalization to unseen networks.

\subsubsection{\texorpdfstring{$k$-NN Classification}{k-NN Classification}}
each generated graph $\hat{G}$ is embedded by the frozen Siamese GNN and classified via $k$-NN against the real training set in the learned embedding space. We use two protocols: global $k$-NN ($k=\lceil\sqrt{\bar{N}}\rceil$, majority vote over all classes) and dynamic $k$-NN~\cite{DBLP:journals/corr/HassanatAAA14}, which assigns each class its own $k_c = \lceil\sqrt{N_c}\rceil$ and predicts $\hat{y}=\arg\min_c \mu_c(\hat{G})$, where:
\begin{equation}
    \mu_c(\hat{G}) = \frac{1}{k_c}\sum_{j=1}^{k_c} d_{\sigma_c{(j)}},
    \label{eq:dynamicknn}
\end{equation}
is the mean $\ell_2$ distance to the $k_c$ nearest same-class anchors ($d_i = \norm{\hat{\emb} - \backbone(G_i^{\mathrm{train}})}_2$). The dynamic variant is especially informative under our class imbalance ($529$ vs.\ $97$ graphs). For both protocols, we report the fraction of generated graphs classified as the target class $y^*$.

\subsubsection{Independent and Structural Evaluation}
\label{sec:structural_metrics}
since the same Siamese GNN is used for both guidance and evaluation, the $k$-NN accuracy confirms that the guidance operates as intended in the embedding space used to define the target, but does not provide an independent assessment of the generated graphs.

We therefore complement it with four evaluations:
(i)~two external-classifier experiments addressing circularity with the guidance encoder (an independent Siamese GNN trained only on real graphs, and a synthetic-to-real transfer test; Section~\ref{sec:indep}, Appendices~\ref{app:indep_classifier} and~\ref{app:gen_only});
(ii)~Structural Coverage at the 95th percentile (Coverage@95) and Pairwise Diversity Ratio (PW-Ratio), which assess per-graph structural plausibility and relative dispersion in descriptor space;
(iii)~a Calibrated Kolmogorov--Smirnov test~\cite{massey1951} (KS-Cal@95), which compares generated and real marginal distributions against a bootstrap-calibrated real-vs-real threshold;
and (iv)~Maximum Mean Discrepancy (MMD)~\cite{JMLR:v13:gretton12a} (Appendix~\ref{app:mmd}) on five kernel-based graph statistics (degree, clustering, orbits, spectral, NSPDK~\cite{10.5555/3104322.3104356}).
Full definitions of Coverage@95, KS-Cal@95, and PW-Ratio are given in Section~\ref{sec:structural}.
This multi-evaluation protocol is motivated by known limitations of MMD as a standalone metric for graph generative models~\cite{obray2022evaluation};
combining classifier-based, per-graph, distributional, and diversity assessments reduces reliance on any single evaluation criterion.

\section{Experimental Setup and Results}
\subsection{Dataset}
We evaluate ProtoGuide on five classes of real-world networks.
Four classes are drawn from the ATLAS benchmark~\cite{zhao2025adaptive,adler2026latent} and include: Connectome (brain networks), Infrastructure (international-trade and transport topologies), Internet (autonomous-system graphs), and Social (online and offline social networks).
The fifth class, Biological, combines $30$ ATLAS biological networks with $118$ additional ecological and molecular interaction networks from the corpus of Ghasemian et al.~\cite{doi:10.1073/pnas.1914950117}.
All graphs are undirected and unweighted.
The dataset in Table~\ref{t:dataset_info} is used to pre-train the Siamese GNN, with a maximum of $3\,000$ nodes per graph.
During generation with both graph generative models, we cap the graph size at a lower number of nodes, because both denoisers (EDGE and DiGress) operate on $\mathcal{O}(N^2)$ candidate edges, making memory and runtime scale quadratically with the node count.

\begin{table}[ht!]
\centering
\small
\caption{\textbf{Dataset statistics.}}
\label{t:dataset_info}
\rowcolors{3}{gray!20}{white}
\begin{tabular}{lrrrrr}
\toprule
\multirow{2}{*}{Class}
& \multirow{2}{*}{\#graphs}
& \multicolumn{2}{c}{\#nodes}
& \multicolumn{2}{c}{\#edges} \\
\cmidrule(lr){3-4} \cmidrule(lr){5-6}
& & Min & Max & Min & Max \\
\midrule
Biological     & 148 & 18  & 2\,989 & 60     & 157\,472 \\
Connectome     & 529 & 19  & 1\,771 & 74     & 179\,982 \\
Infrastructure & 299 & 16  & 2\,906 & 34     & 75\,894 \\
Internet       & 97  & 282 & 2\,888 & 2\,060 & 24\,562 \\
Social         & 111 & 17  & 2\,971 & 72     & 225\,970 \\
\bottomrule
\end{tabular}
\end{table}

\subsection{Experimental Setup}\label{sec:exp_setup}

\paragraph{Siamese Encoder}
it uses $L{=}4$ Graph Attention network (GAT~\cite{veličković2018graphattentionnetworks}) layers, $H{=}4$ attention heads, hidden dimension $d_h{=}8$,
output dimension $d_o{=}16$, yielding an embedding dimension $D{=}116$ and approximately $30\text{K}$ parameters.
The model was selected via grid search (Appendix~\ref{app:encoder_grid}) over 192 configurations (encoder architecture, number of layers,
hidden dimension, learning rate, margin) and trained with triplet margin loss ($m{=}1.0$, $\text{lr}{=}10^{-3}$) on the original dataset of Table~\ref{t:dataset_info}.
Its weights are frozen throughout generation and evaluation, and the prototypes $\{\proto_c\}$ are computed once before sampling begins.
\looseness-1

\paragraph{Graph Generative Model Training}
we train one model per class for both EDGE and DiGress, each for $2\,000$ epochs.
EDGE uses the architecture and hyperparameters of the original paper, with a reduced batch size; DiGress follows its original configuration with a reduced model dimension, both for computational efficiency.
Each model is used for both unguided (baseline, $\lambda{=}0$) and ProtoGuide generation, so the only difference between the two conditions is the Siamese GAT guidance signal at sampling time. 
Per-class configurations for both EDGE and DiGress, including training set sizes, maximum node counts, and selected guidance scales, are reported in Appendix~\ref{app:model_config} (Tables~\ref{t:edge_setup} and~\ref{t:digress_setup}).
A key difference between the two models is the maximum graph size ($N_{\max}$ up to $3\,000$ for EDGE vs.\ $1\,000$ for DiGress), which results in different reference set sizes per class.
\looseness-1

\paragraph{DiGress Conditional Baseline}
DiGress natively supports class-conditional generation by concatenating a one-hot class label to the global context vector $\mathbf{y}$ fed into the denoising transform.
To compare this built-in mechanism against ProtoGuide directly, we train a single conditional model jointly on all five classes for $2\,000$ epochs, matching the training budget of the per-class models, and sample $64$ graphs per class by setting the target label at generation time.
We cap this model at $N_{\max}=500$ nodes: because the conditional model must fit all five classes in each training batch and the edge tensor scales as $\mathcal{O}(N^2)$, $N_{\max}=1\,000$ was infeasible under our GPU memory budget. This choice matches the per-class cap for Biological, Connectome, and Infrastructure (Table~\ref{t:digress_setup}), but it is lower than the $N_{\max}=1\,000$ used by the per-class Internet and Social models; the resulting truncation of larger graphs in these two classes may contribute to the conditional model's poor performance on them. To isolate this confound, we compare the conditional model against ProtoGuide on Biological, where both operate at $N_{\max}{=}500$: the conditional model still drops to $23.9\%$ versus ProtoGuide's $48.6\%$ (Section~\ref{sec:results}), showing that truncation alone cannot explain the performance gap.
This represents a standard approach to class-conditional graph diffusion: the class signal is available during training, and the model must learn to partition its capacity across classes.

\paragraph{Guidance Scale Selection}
the guidance scale $\lambda$ controls how strongly the Siamese gradient steers the denoising trajectory.
Because the effective guidance range differs across class--model pairs in our experiments, we select $\lambda$ independently for each pair.
A full search over a wide range of values is computationally expensive, as each candidate requires an additional generation and evaluation step.
We therefore perform a coarse grid search over $\lambda\!\in\!\{0.1, 0.2, 0.25, 0.5, 1, 2, 3, 3.5, 5\}$ on a validation generation run and select the value that maximizes $k$-NN classification accuracy.
The selected values are reported in Tables~\ref{t:edge_setup} and~\ref{t:digress_setup}.

The ablation in Appendix~\ref{ss:guidance_scale} illustrates this class dependence on Connectome and Biological, which respond very differently as $\lambda$ increases.
This motivates class-specific scale selection rather than a single global value.

\subsection{Results}
\label{sec:results}
All results are computed over $10$ independent generation seeds, each producing $64$ graphs per class, unless stated otherwise.
Each model is compared against its own training set (Tables~\ref{t:edge_setup}-\ref{t:digress_setup}), so reference sizes differ across models; cross-model comparison should be interpreted with this caveat.

\subsubsection{Class Controllability via \texorpdfstring{$k$-NN}{k-NN}}
Table~\ref{t:accuracy_dynamic} summarizes the
per-class classification accuracy (the diagonal of the confusion matrix) for both denoising models under baseline ($\lambda{=}0$), ProtoGuide, and DiGress built-in conditional mechanism.
Full $5 {\times} 5$ confusion matrices and global $k$-NN results are provided in Appendix~\ref{app:confusion}.

\begin{table}[!t]
\centering
\small
\caption{\textbf{Per-class classification accuracy (\%) --- Dynamic $k$-NN}.
         Mean $\pm$ std over 10 generation seeds ($64$ graphs each).
         \textbf{Bold} = best per class across conditions within each backbone, ${}^\dagger$\, best macro across conditions within each backbone.}
\label{t:accuracy_dynamic}
\rowcolors{2}{gray!25}{white}
\begin{tabular}{ll cccccc}
\toprule
 & & \multicolumn{5}{c}{Target Class} & \\
\cmidrule(lr){3-7}
Model & Condition & Biological & Connectome & Infrastructure & Internet & Social & Macro \\
\midrule
EDGE
 & Baseline & $\mathbf{62.5 \pm 5.5}$ & $11.6 \pm 3.8$ & $\mathbf{39.2 \pm 10.7}$ & $88.3 \pm 5.1$ & $52.0 \pm 6.0$ & $50.7 \pm 3.0$ \\
 & ProtoGuide & $61.4 \pm 6.3$ & $\mathbf{85.0 \pm 3.7}$ & $34.4 \pm 6.7$ & $\mathbf{94.8 \pm 2.7}$ & $\mathbf{92.0 \pm 2.8}$ & ${}^\dagger{73.5 \pm 1.5}$ \\
\midrule
DiGress
 & Baseline & $41.4 \pm 9.7$ & $90.0 \pm 4.9$ & $\mathbf{90.0 \pm 5.8}$ & $95.8 \pm 3.0$ & $50.6 \pm 4.5$ & $73.6 \pm 3.0$ \\
 & ProtoGuide & $\mathbf{48.6 \pm 6.5}$ & $\mathbf{91.6 \pm 3.8}$ & $89.1 \pm 4.7$ & $\mathbf{97.7 \pm 1.6}$ & $\mathbf{91.9 \pm 4.0}$ & ${}^\dagger{83.8 \pm 1.6}$ \\
 & Conditional & $23.9 \pm 3.3$ & $61.1 \pm 5.8$ & $87.8 \pm 3.0$ & $7.0 \pm 2.8$ & $27.3 \pm 4.3$ & $41.4 \pm 2.1$ \\
\bottomrule
\end{tabular}
\end{table}

ProtoGuide improves macro classification accuracy on both backbones: EDGE rises from $50.7\%$ to $73.5\%$ and DiGress from $73.6\%$ to $83.8\%$.
Observing macro-level gains with two architecturally different denoisers shows that the guidance mechanism can improve class alignment under different discrete reverse-process parametrizations, although the effect is not uniform across classes.

The two graph generative models exhibit complementary per-class strengths.
DiGress achieves strong accuracy on the Internet ($95.8 \!\to\! 97.7\%$) and Infrastructure ($90.0 \!\to\! 89.1\%$), where EDGE reaches at best $94.8\%$ and $39.2\%$ under ProtoGuide, respectively.
By contrast, EDGE obtains the highest accuracy on Biological networks ($62.5\%$).
Social benefits strongly from guidance under both models, rising from $52.0\%$ to $92.0\%$ (EDGE) and from $50.6\%$ to $91.9\%$ (DiGress).

The largest absolute improvement appears for Connectome under EDGE, where accuracy jumps from $11.6\%$ to $85.0\%$; the baseline model misclassified nearly all Connectome graphs as Social ($84.4\%$, Table~\ref{t:app_edge_baseline_dynamic}), and ProtoGuide corrects this systematic confusion.
Infrastructure under EDGE is, together with Biological ($62.5\%$ vs.\ $61.4\%$), one of two classes where guidance does not improve accuracy ($39.2\%$ vs.\ $34.4\%$).
It remains a difficult case for EDGE under both baseline and guided sampling.
DiGress Conditional achieves only $41.4\%$ macro accuracy, substantially below both the unconditional DiGress baseline ($73.6\%$) and ProtoGuide ($83.8\%$).
It struggles most on Internet ($7.0\%$) and Social ($27.3\%$), where it confuses Internet primarily with Infrastructure and Social primarily with Connectome (Table~\ref{t:app_digress_conditional_dynamic}).
Under our training configuration, the jointly trained conditional model therefore develops a pronounced class bias, with Infrastructure the only class on which it retains high accuracy ($87.8\%$).
Because this baseline is a single jointly trained model whereas ProtoGuide is applied to class-specific generators, the comparison characterizes the two conditioning strategies under our experimental setup rather than isolating a single cause of the performance gap.
We describe the observed behavior as class bias rather than generative mode collapse, since the confusion matrix alone does not establish collapse of the generated distribution.

\subsubsection{Independent Classifier Evaluation}
\label{sec:indep}

to validate class controllability with an embedding model that is not used by ProtoGuide during guidance, we train an architecturally distinct Siamese GCN (3 layers, 32 hidden dimension, Jumping Knowledge concatenation, attentional pooling) exclusively on real graphs, with $N_{\max}$ matching Table~\ref{t:dataset_info} ($80\%$ train, $20\%$ validation).
We use this model to classify generated graphs via $k$-NN against the real training embeddings.
The classifier is independently initialized, architecturally distinct from the guidance GAT, and never sees generated graphs during training.
We evaluate all five conditions (EDGE baseline, EDGE + ProtoGuide, DiGress baseline, DiGress conditional, DiGress + ProtoGuide) across $10$ classifier seeds and report mean$\pm$std in Table~\ref{t:indep_classifier}.
\looseness=-1

\begin{table}[ht!]
\centering
\small
\caption{\textbf{Independent classifier validation.}
  Macro accuracy (\%) over $10$ classifier seeds ($3{,}200$ graphs per 
  condition). Parenthesized: improvement over baseline. 
  \textbf{Bold} = best per backbone.}
\label{t:indep_classifier}
\rowcolors{2}{gray!25}{white}
\begin{tabular}{l ccc ccc}
\toprule
 & \multicolumn{2}{c}{EDGE} & \multicolumn{3}{c}{DiGress} \\
\cmidrule(lr){2-3}\cmidrule(lr){4-6}
 & Baseline & +ProtoGuide & Baseline & +ProtoGuide & Conditional \\
\midrule
Global $k$-NN  & $63.1 {\pm} 3.6$ & $\mathbf{71.1 {\pm} 2.1}$\;{\scriptsize($+8.0$)}  & $69.4 {\pm} 1.5$ & $\mathbf{78.4 {\pm} 0.9}$\;{\scriptsize($+9.0$)}  & $41.7 {\pm} 0.6$ \\
Dynamic $k$-NN & $63.2 {\pm} 4.4$ & $\mathbf{73.4 {\pm} 2.1}$\;{\scriptsize($+10.2$)} & $71.5 {\pm} 1.2$ & $\mathbf{79.2 {\pm} 1.1}$\;{\scriptsize($+7.7$)} & $42.6 {\pm} 0.7$ \\
\bottomrule
\end{tabular}
\end{table}

Table~\ref{t:indep_classifier} corroborates the macro-level improvements observed with ProtoGuide using an architecturally distinct evaluation model.
The independent GCN assigns ProtoGuide-generated graphs to their target class more accurately in aggregate than the corresponding unguided samples: $+10.2$~pp for EDGE and $+7.7$~pp for DiGress under dynamic $k$-NN.
The per-class breakdown (Appendix~\ref{app:indep_classifier}) shows that these gains are not uniform: Social benefits strongly from guidance (EDGE: $54.5 \!\to\! 89.0\%$; DiGress: $52.3 \!\to\! 91.2\%$), while Connectome and Infrastructure already achieve $>84\%$ in all ProtoGuide conditions.

A notable exception is EDGE ProtoGuide on Internet class, where, as shown in Supplementary Material Table~\ref{t:indep_perclass}, the independent Siamese GCN reports a drop from $92.6\%$ to $76.0\%$ under ProtoGuide;
the independent classifier disagrees with Table~\ref{t:accuracy_dynamic} where the Siamese GAT evaluation for Internet increases from $88.3\%$ to $94.8\%$.
This divergence is accompanied by zero Coverage@95 for EDGE Internet both with and without guidance (Table~\ref{t:structural_fidelity}).
EDGE's generated Internet graphs frequently contain disconnected components, indicating a pre-existing structural limitation of the backbone on this class; ProtoGuide improves alignment in the guidance embedding but does not resolve that limitation, and the independent classifier in fact assigns the guided samples to Internet less often.
DiGress does not show the same disagreement on Internet ($98.5\% \to 99.1\%$ under the independent classifier).
DiGress Conditional falls below the unconditional baseline ($42.6\%$ vs.\
$71.5\%$) by nearly $29$~pp and trails ProtoGuide by $36.6$~pp.
Its Internet accuracy drops to $8.5\%$, showing that, under this external evaluation, the built-in conditional model fails to produce Internet graphs recognized as class-representative.
ProtoGuide does not exhibit this failure on DiGress, reaching $99.1\%$ on the same class.
Overall, the macro-level guidance gains are also detected by an independently initialized and architecturally distinct model trained on real graphs (Appendix~\ref{app:indep_classifier}).

A synthetic-to-real transfer test (Appendix~\ref{app:gen_only}) points the same way: a Siamese GIN trained only on graphs generated by ProtoGuide reaches $79.2\%$ macro accuracy on real networks held out from the GIN's own training, indicating that the generated graphs carry class-structural regularities recognizable on real data.

\subsubsection{Structural Fidelity}
\label{sec:structural}
we evaluate structural fidelity from three complementary perspectives: multivariate descriptor-space coverage and diversity, marginal distributional calibration through KS, and kernel-based comparison through MMD (Appendix~\ref{app:mmd}).

\paragraph{Coverage, calibrated KS, and diversity}

Table~\ref{t:structural_fidelity} reports three complementary structural metrics computed from nine graph-level descriptors:
number of nodes ($\log N$), density, assortativity, average $k$-core, transitivity, average clustering, number of connected components, LCC fraction, and inverse average path length.

Coverage@95 measures per-graph structural plausibility in the jointly IQR-standardized descriptor space.
For each generated graph, we compute its $k$-NN distance to the real reference set and compare it with the 95th percentile of the corresponding real-to-real distances.
Coverage@95 is the fraction of generated graphs that fall within this threshold.

KS-Cal@95 measures marginal distributional agreement for each descriptor.
The two-sample Kolmogorov--Smirnov (KS) distance~\cite{massey1951} is the maximum difference between the empirical cumulative distribution functions of the generated and real samples, with smaller values indicating closer marginal distributions.
Rather than aggregating raw KS distances directly, we calibrate each class--descriptor pair against its empirical real-vs-real variability.
Specifically, we bootstrap $1\,000$ real-vs-real splits matched in size to the generated set and use the 95th percentile of their KS distances as a calibration threshold.
A generated seed passes for a descriptor when its KS distance to the real set falls below this threshold.
KS-Cal@95 is the fraction of descriptor--seed comparisons that pass, so higher values indicate that more generated marginals fall within the empirically calibrated real-vs-real variability.

PW-Ratio measures relative dispersion in the same standardized descriptor space.
It is the ratio between the mean pairwise Euclidean distance among generated graphs and that among real graphs.
Values near $1$ indicate comparable dispersion, values below $1$ indicate under-dispersion, and values above $1$ indicate over-dispersion relative to the real graphs.

\begin{table}[ht!]
\small
\centering
\caption{\textbf{Structural fidelity and diversity} (mean $\pm$ std 
  over 10 seeds). \textbf{Bold} = best per class within each backbone.}
\label{t:structural_fidelity}
\rowcolors{2}{gray!25}{white}
\begin{tabular}{ll ccc}
\toprule
Condition & Class & Coverage@95 (\%) $\uparrow$ & KS-Cal@95 (\%) $\uparrow$ & PW-Ratio ${\approx}\,1$ \\
\midrule
EDGE Baseline & Biological & $\mathbf{98.9 {\pm} 0.7}$ & $\mathbf{46.7 {\pm} 4.4}$ & $0.72 {\pm} 0.04$ \\
 & Connectome & $94.8 {\pm} 2.4$ & $38.9 {\pm} 12.4$ & $\mathbf{0.99 {\pm} 0.19}$ \\
 & Infrastructure & $96.2 {\pm} 2.7$ & $\mathbf{20.0 {\pm} 4.4}$ & $\mathbf{0.85 {\pm} 0.37}$ \\
 & Internet & $0.0 {\pm} 0.0$ & $\mathbf{22.2 {\pm} 0.0}$ & $\mathbf{0.97 {\pm} 0.08}$ \\
 & Social & $98.9 {\pm} 1.6$ & $34.4 {\pm} 14.4$ & $0.75 {\pm} 0.05$ \\
\midrule
EDGE{+}ProtoGuide & Biological & $98.1 {\pm} 0.9$ & $\mathbf{46.7 {\pm} 4.4}$ & $\mathbf{0.73 {\pm} 0.06}$ \\
 & Connectome & $\mathbf{98.0 {\pm} 1.7}$ & $\mathbf{58.9 {\pm} 14.9}$ & $1.05 {\pm} 0.14$ \\
 & Infrastructure & $\mathbf{97.0 {\pm} 2.4}$ & $18.9 {\pm} 5.1$ & $0.65 {\pm} 0.21$ \\
 & Internet & $0.0 {\pm} 0.0$ & $16.7 {\pm} 7.5$ & $1.85 {\pm} 0.19$ \\
 & Social & $\mathbf{99.2 {\pm} 0.8}$ & $\mathbf{57.8 {\pm} 6.7}$ & $\mathbf{0.81 {\pm} 0.02}$ \\
\midrule
DiGress Baseline & Biological & $97.2 {\pm} 5.3$ & $23.3 {\pm} 9.2$ & $0.68 {\pm} 0.24$ \\
 & Connectome & $96.7 {\pm} 0.8$ & $\mathbf{47.8 {\pm} 13.2}$ & $0.93 {\pm} 0.09$ \\
 & Infrastructure & $\mathbf{93.4 {\pm} 7.1}$ & $17.8 {\pm} 7.4$ & $0.53 {\pm} 0.37$ \\
 & Internet & $70.3 {\pm} 5.8$ & $41.1 {\pm} 7.1$ & $1.10 {\pm} 0.09$ \\
 & Social & $98.3 {\pm} 1.8$ & $\mathbf{48.9 {\pm} 5.4}$ & $0.73 {\pm} 0.06$ \\
\midrule
DiGress{+}ProtoGuide & Biological & $\mathbf{99.8 {\pm} 0.5}$ & $\mathbf{24.4 {\pm} 4.4}$ & $0.56 {\pm} 0.06$ \\
 & Connectome & $\mathbf{97.3 {\pm} 1.4}$ & $26.7 {\pm} 15.9$ & $\mathbf{1.03 {\pm} 0.18}$ \\
 & Infrastructure & $91.7 {\pm} 7.2$ & $\mathbf{22.2 {\pm} 14.9}$ & $0.63 {\pm} 0.43$ \\
 & Internet & $\mathbf{71.9 {\pm} 5.0}$ & $\mathbf{44.4 {\pm} 0.0}$ & $\mathbf{1.08 {\pm} 0.07}$ \\
 & Social & $93.4 {\pm} 2.5$ & $28.9 {\pm} 13.3$ & $\mathbf{0.95 {\pm} 0.08}$ \\
\midrule
DiGress Conditional & Biological & $87.7 {\pm} 3.8$ & $13.3 {\pm} 8.3$ & $\mathbf{1.14 {\pm} 0.06}$ \\
 & Connectome & $77.0 {\pm} 5.5$ & $8.9 {\pm} 10.9$ & $2.82 {\pm} 0.36$ \\
 & Infrastructure & $91.2 {\pm} 7.3$ & $16.7 {\pm} 7.5$ & $\mathbf{1.07 {\pm} 0.64}$ \\
 & Internet & $0.0 {\pm} 0.0$ & $2.2 {\pm} 6.7$ & $9.63 {\pm} 0.48$ \\
 & Social & $\mathbf{99.2 {\pm} 1.0}$ & $34.4 {\pm} 10.5$ & $0.66 {\pm} 0.05$ \\
\bottomrule
\end{tabular}
\end{table}

Table~\ref{t:structural_fidelity} shows that the structural effect of guidance differs between the two backbones.
On EDGE, ProtoGuide is neutral to positive on several measures: coverage is maintained or slightly improved on four of five classes, and macro KS-Cal@95 rises from $32.4\%$ to $39.8\%$, with the largest gains on Connectome and Social.
On DiGress, the picture is more mixed: coverage remains above $91\%$ on four classes, but decreases on Social ($98.3\% \to 93.4\%$) and Infrastructure ($93.4\% \to 91.7\%$), while KS-Cal@95 decreases notably on Connectome ($47.8\% \to 26.7\%$) and Social ($48.9\% \to 28.9\%$).
Importantly, controllability and fidelity do not move in lockstep: Social combines a large accuracy gain with lower KS calibration, whereas Connectome shows only a small accuracy increase under DiGress despite a substantial KS-Cal@95 decrease.
The structural effect of guidance is therefore both backbone- and class-dependent and cannot be inferred directly from the magnitude of the controllability gain.
PW-Ratio under ProtoGuide lies between $0.56$ and $1.08$ for all classes except EDGE Internet ($1.85$), indicating varying degrees of under- or over-dispersion rather than, by itself, mode collapse.
DiGress Conditional is much more over-dispersed on Connectome ($2.82$) and especially Internet ($9.63$) relative to the real reference graphs.
\looseness=-1

Internet coverage is zero for EDGE both with and without guidance, so this failure cannot be attributed to ProtoGuide.
The reference Internet graphs are connected, whereas EDGE frequently generates samples with multiple disconnected components, causing them to fail the multivariate coverage criterion in both conditions.

Detailed per-metric KS results and per-$k$ coverage breakdowns are provided in Appendix~\ref{sec:evaluation_plus}.

\section{Discussion and Conclusion}

ProtoGuide shows how a classifier-guidance-like mechanism can be instantiated for discrete graph diffusion without retraining the generative model.
Relaxing the denoiser's per-edge output into a soft adjacency provides a differentiable path for graph-level guidance, while the frozen Siamese embedding and class prototypes define the target signal without requiring a classifier trained on noisy diffusion states.
Across two architecturally different backbones and five classes of real-world networks, ProtoGuide raises macro classification accuracy substantially (EDGE: $50.7\% \to 73.5\%$; DiGress: $73.6\% \to 83.8\%$).
Structural evaluation shows that these controllability gains coexist with high per-graph coverage in most settings, while distributional effects remain class- and backbone-dependent (Section~\ref{sec:results}).

The experiments also show that controllability and structural fidelity do not vary in lockstep.
Guidance can improve both, as for EDGE Connectome, or produce strong class-alignment gains alongside shifts in some structural distributions, as observed for DiGress Social.
Conversely, some distributional shifts occur even when the corresponding controllability gain is small.
The directionality experiment further shows that steering depends on the selected prototype: replacing the correct target with wrong or random directions substantially changes the resulting class alignment.
The few-shot analysis shows a complementary form of robustness, with prototype targets estimated from very small support sets producing performance close to those estimated from larger sets.

\paragraph{Limitations and future work}

ProtoGuide does not uniformly correct structural deficiencies of the underlying generative backbone.
For example, EDGE achieves zero Internet Coverage@95 even without guidance (Table~\ref{t:structural_fidelity}), and guidance does not remove this pre-existing limitation.
More importantly, our protocol trains one unconditional backbone per class and steers each model toward the class it already represents.
The present experiments therefore establish post-hoc class steering within the distribution learned by each backbone, rather than generation of a class absent from its training distribution.
Testing a single pooled unconditional backbone steered toward multiple class prototypes is a natural extension of this work.

The same distinction applies to the few-shot experiment: it demonstrates robustness of prototype construction when only a small support set is used after both the generative backbone and Siamese encoder have been trained, rather than few-shot learning of an unseen class.
The single-prototype formulation also summarizes each class by one centroid; classes with multi-modal structural organization may benefit from mixture or hierarchical prototypes.
Finally, the differentiable clustering feature used here is an algebraic relaxation of the binary clustering coefficient and can leave the $[0,1]$ range for fractional adjacencies with soft degree below one.
Evaluating alternative differentiable relaxations and feature sets is left to future work.
\looseness=-1

\paragraph{Conclusion}
Real-world complex networks provide an important setting for graph generation beyond molecular applications, and post-hoc control offers a way to decouple class steering from generative-model training.
ProtoGuide implements this idea for discrete graph diffusion through a differentiable soft adjacency relaxation and frozen prototype-based guidance.
On EDGE and DiGress, it substantially improves class controllability without retraining either backbone, with structural effects that vary across classes and models.
These results position prototype-driven guidance as a complementary alternative to training conditioning directly into discrete graph generative models.

\section{Data and Code Availability}
Code will be released upon acceptance. Data are available from the original sources.

\bibliographystyle{IEEEtran}
\bibliography{references}

@inproceedings{10.1007/978-3-031-70359-1_19,
  address   = {Cham},
  author    = {Ninniri, Matteo
               and Podda, Marco
               and Bacciu, Davide},
  booktitle = {Machine Learning and Knowledge Discovery in Databases. Research Track},
  editor    = {Bifet, Albert
               and Davis, Jesse
               and Krilavi{\v{c}}ius, Tomas
               and Kull, Meelis
               and Ntoutsi, Eirini
               and {\v{Z}}liobait{\.{e}}, Indr{\.{e}}},
  isbn      = {978-3-031-70359-1},
  pages     = {318--335},
  publisher = {Springer Nature Switzerland},
  title     = {Classifier-Free Graph Diffusion for Molecular Property Targeting},
  year      = {2024}
}

@inproceedings{10.1609/aaai.v37i4.25549,
  articleno = {480},
  author    = {Huang, Han and Sun, Leilei and Du, Bowen and Lv, Weifeng},
  booktitle = {Proceedings of the Thirty-Seventh AAAI Conference on Artificial Intelligence and Thirty-Fifth Conference on Innovative Applications of Artificial Intelligence and Thirteenth Symposium on Educational Advances in Artificial Intelligence},
  doi       = {10.1609/aaai.v37i4.25549},
  isbn      = {978-1-57735-880-0},
  numpages  = {10},
  publisher = {AAAI Press},
  series    = {AAAI'23/IAAI'23/EAAI'23},
  title     = {Conditional diffusion based on discrete graph structures for molecular graph generation},
  url       = {https://doi.org/10.1609/aaai.v37i4.25549},
  year      = {2023}
}

@inproceedings{10.5555/3104322.3104356,
  address   = {Madison, WI, USA},
  author    = {Costa, Fabrizio and Grave, Kurt De},
  booktitle = {Proceedings of the 27th International Conference on International Conference on Machine Learning},
  isbn      = {9781605589077},
  location  = {Haifa, Israel},
  numpages  = {8},
  pages     = {255–262},
  publisher = {Omnipress},
  series    = {ICML'10},
  title     = {Fast neighborhood subgraph pairwise distance Kernel},
  year      = {2010}
}

@inproceedings{1640964,
  author    = {Hadsell, R. and Chopra, S. and LeCun, Y.},
  booktitle = {2006 IEEE Computer Society Conference on Computer Vision and Pattern Recognition (CVPR'06)},
  doi       = {10.1109/CVPR.2006.100},
  number    = {},
  pages     = {1735-1742},
  title     = {Dimensionality Reduction by Learning an Invariant Mapping},
  volume    = {2},
  year      = {2006}
}

@inproceedings{3294771.3294869,
  address   = {Red Hook, NY, USA},
  author    = {Hamilton, William L. and Ying, Rex and Leskovec, Jure},
  booktitle = {Proceedings of the 31st International Conference on Neural Information Processing Systems},
  isbn      = {9781510860964},
  location  = {Long Beach, California, USA},
  numpages  = {11},
  pages     = {1025–1035},
  publisher = {Curran Associates Inc.},
  series    = {NIPS'17},
  title     = {Inductive representation learning on large graphs},
  year      = {2017}
}

@inproceedings{3294996.3295163,
  address   = {Red Hook, NY, USA},
  author    = {Snell, Jake and Swersky, Kevin and Zemel, Richard},
  booktitle = {Proceedings of the 31st International Conference on Neural Information Processing Systems},
  isbn      = {9781510860964},
  location  = {Long Beach, California, USA},
  numpages  = {11},
  pages     = {4080–4090},
  publisher = {Curran Associates Inc.},
  series    = {NIPS'17},
  title     = {Prototypical networks for few-shot learning},
  year      = {2017}
}

@inbook{3454287.3454670,
  address   = {Red Hook, NY, USA},
  articleno = {383},
  author    = {Liao, Renjie and Li, Yujia and Song, Yang and Wang, Shenlong and Hamilton, William L. and Duvenaud, David and Urtasun, Raquel and Zemel, Richard},
  booktitle = {Proceedings of the 33rd International Conference on Neural Information Processing Systems},
  numpages  = {11},
  publisher = {Curran Associates Inc.},
  title     = {Efficient graph generation with graph recurrent attention networks},
  year      = {2019}
}

@inbook{3454287.3455360,
  address   = {Red Hook, NY, USA},
  articleno = {1073},
  author    = {Yun, Seongjun and Jeong, Minbyul and Kim, Raehyun and Kang, Jaewoo and Kim, Hyunwoo J.},
  booktitle = {Proceedings of the 33rd International Conference on Neural Information Processing Systems},
  numpages  = {11},
  publisher = {Curran Associates Inc.},
  title     = {Graph transformer networks},
  year      = {2019}
}

@inproceedings{3495724.3496298,
  address   = {Red Hook, NY, USA},
  articleno = {574},
  author    = {Ho, Jonathan and Jain, Ajay and Abbeel, Pieter},
  booktitle = {Proceedings of the 34th International Conference on Neural Information Processing Systems},
  isbn      = {9781713829546},
  location  = {Vancouver, BC, Canada},
  numpages  = {12},
  publisher = {Curran Associates Inc.},
  series    = {NIPS '20},
  title     = {Denoising diffusion probabilistic models},
  year      = {2020}
}

@inproceedings{3540261.3540933,
  address   = {Red Hook, NY, USA},
  articleno = {672},
  author    = {Dhariwal, Prafulla and Nichol, Alex},
  booktitle = {Proceedings of the 35th International Conference on Neural Information Processing Systems},
  isbn      = {9781713845393},
  numpages  = {15},
  publisher = {Curran Associates Inc.},
  series    = {NIPS '21},
  title     = {Diffusion models beat GANs on image synthesis},
  year      = {2021}
}

@inproceedings{3540261.3541637,
  address   = {Red Hook, NY, USA},
  articleno = {1376},
  author    = {Austin, Jacob and Johnson, Daniel D. and Ho, Jonathan and Tarlow, Daniel and van den Berg, Rianne},
  booktitle = {Proceedings of the 35th International Conference on Neural Information Processing Systems},
  isbn      = {9781713845393},
  numpages  = {13},
  publisher = {Curran Associates Inc.},
  series    = {NIPS '21},
  title     = {Structured denoising diffusion models in discrete state-spaces},
  year      = {2021}
}

@inproceedings{3618408.3618589,
  articleno = {181},
  author    = {Chen, Xiaohui and He, Jiaxing and Han, Xu and Liu, Li-Ping},
  booktitle = {Proceedings of the 40th International Conference on Machine Learning},
  location  = {Honolulu, Hawaii, USA},
  numpages  = {26},
  publisher = {JMLR.org},
  series    = {ICML'23},
  title     = {Efficient and degree-guided graph generation via discrete diffusion modeling},
  year      = {2023}
}

@article{4700287,
  author   = {Scarselli, Franco and Gori, Marco and Tsoi, Ah Chung and Hagenbuchner, Markus and Monfardini, Gabriele},
  doi      = {10.1109/TNN.2008.2005605},
  journal  = {IEEE Transactions on Neural Networks},
  number   = {1},
  pages    = {61-80},
  title    = {The Graph Neural Network Model},
  volume   = {20},
  year     = {2009}
}

@article{48e4505b7f124ee89368447f32b5cf0e,
  author    = {Jane Bromley and J.W. Bentz and Leon Bottou and I. Guyon and Yann Lecun and C. Moore and Eduard Sackinger and R. Shah},
  issn      = {0218-0014},
  journal   = {International Journal of Pattern Recognition and Artificial Intelligence},
  language  = {English (US)},
  month     = aug,
  number    = {4},
  publisher = {World Scientific Publishing Co. Pte Ltd},
  title     = {Signature verification using a Siamese time delay neural network},
  volume    = {7},
  year      = {1993}
}

@inproceedings{adler2026latent,
  author    = {Thomas Adler and Marco Grassia and Ziheng Liao and Giuseppe Mangioni and Carlo Vittorio Cannistraci},
  booktitle = {The Fourteenth International Conference on Learning Representations},
  title     = {Latent Geometry-Driven Network Automata for Complex Network Dismantling},
  url       = {https://openreview.net/forum?id=yz29QCGVzC},
  year      = {2026}
}

@article{Baraba_si_1999,
  author    = {Barabási, Albert-László and Albert, Réka},
  doi       = {10.1126/science.286.5439.509},
  issn      = {1095-9203},
  journal   = {Science},
  month     = oct,
  number    = {5439},
  pages     = {509–512},
  publisher = {American Association for the Advancement of Science (AAAS)},
  title     = {Emergence of Scaling in Random Networks},
  url       = {http://dx.doi.org/10.1126/science.286.5439.509},
  volume    = {286},
  year      = {1999}
}

@article{Barabasi2004Network,
  address              = {Department of Physics, University of Notre Dame, Notre Dame, Indiana 46556, USA. alb@nd.edu},
  author               = {Barabasi, Albert-Laszlo and Oltvai, Zoltan N.},
  day                  = 01,
  doi                  = {10.1038/nrg1272},
  issn                 = {1471-0056},
  journal              = {Nat Rev Genet},
  month                = feb,
  number               = 2,
  pages                = {101--113},
  pmid                 = {14735121},
  publisher            = {Nature Publishing Group},
  title                = {Network biology: understanding the cell's functional organization},
  url                  = {http://dx.doi.org/10.1038/nrg1272},
  volume               = 5,
  year                 = 2004
}

@inproceedings{conf/icml/YouYRHL18,
  author    = {You, Jiaxuan and Ying, Rex and Ren, Xiang and Hamilton, William L. and Leskovec, Jure},
  booktitle = {ICML},
  editor    = {Dy, Jennifer G. and Krause, Andreas},
  ee        = {http://proceedings.mlr.press/v80/you18a.html},
  pages     = {5694-5703},
  publisher = {PMLR},
  series    = {Proceedings of Machine Learning Research},
  title     = {GraphRNN: Generating Realistic Graphs with Deep Auto-regressive Models.},
  url       = {http://dblp.uni-trier.de/db/conf/icml/icml2018.html#YouYRHL18},
  volume    = 80,
  year      = 2018
}

@inproceedings{DBLP:conf/iclr/0011SKKEP21,
  author    = {Yang Song and
               Jascha Sohl{-}Dickstein and
               Diederik P. Kingma and
               Abhishek Kumar and
               Stefano Ermon and
               Ben Poole},
  bibsource = {dblp computer science bibliography, https://dblp.org},
  booktitle = {9th International Conference on Learning Representations, {ICLR} 2021,
               Virtual Event, Austria, May 3-7, 2021},
  publisher = {OpenReview.net},
  title     = {Score-Based Generative Modeling through Stochastic Differential Equations},
  url       = {https://openreview.net/forum?id=PxTIG12RRHS},
  year      = {2021}
}

@inproceedings{DBLP:conf/iclr/VignacKSWCF23,
  author    = {Cl{\'{e}}ment Vignac and
               Igor Krawczuk and
               Antoine Siraudin and
               Bohan Wang and
               Volkan Cevher and
               Pascal Frossard},
  bibsource = {dblp computer science bibliography, https://dblp.org},
  booktitle = {The Eleventh International Conference on Learning Representations,
               {ICLR} 2023, Kigali, Rwanda, May 1-5, 2023},
  publisher = {OpenReview.net},
  title     = {DiGress: Discrete Denoising diffusion for graph generation},
  url       = {https://openreview.net/forum?id=UaAD-Nu86WX},
  year      = {2023}
}

@inproceedings{DBLP:conf/iclr/XuHLJ19,
  author    = {Keyulu Xu and
               Weihua Hu and
               Jure Leskovec and
               Stefanie Jegelka},
  bibsource = {dblp computer science bibliography, https://dblp.org},
  booktitle = {7th International Conference on Learning Representations, {ICLR} 2019,
               New Orleans, LA, USA, May 6-9, 2019},
  publisher = {OpenReview.net},
  title     = {How Powerful are Graph Neural Networks?},
  url       = {https://openreview.net/forum?id=ryGs6iA5Km},
  year      = {2019}
}

@article{DBLP:journals/corr/HassanatAAA14,
  author     = {Ahmad Basheer Hassanat and
                Mohammad Ali Abbadi and
                Ghada Awad Altarawneh and
                Ahmad Ali Alhasanat},
  bibsource  = {dblp computer science bibliography, https://dblp.org},
  eprint     = {1409.0919},
  eprinttype = {arXiv},
  journal    = {CoRR},
  title      = {Solving the Problem of the {K} Parameter in the {KNN} Classifier Using
                an Ensemble Learning Approach},
  url        = {http://arxiv.org/abs/1409.0919},
  volume     = {abs/1409.0919},
  year       = {2014}
}

@article{doi:10.1073/pnas.1914950117,
  author   = {Amir Ghasemian  and Homa Hosseinmardi  and Aram Galstyan  and Edoardo M. Airoldi  and Aaron Clauset },
  doi      = {10.1073/pnas.1914950117},
  eprint   = {https://www.pnas.org/doi/pdf/10.1073/pnas.1914950117},
  journal  = {Proceedings of the National Academy of Sciences},
  number   = {38},
  pages    = {23393-23400},
  title    = {Stacking models for nearly optimal link prediction in complex networks},
  url      = {https://www.pnas.org/doi/abs/10.1073/pnas.1914950117},
  volume   = {117},
  year     = {2020}
}

@inproceedings{dou2026plugandplay,
  author    = {Hongkun Dou and Zike Chen and Fengji Li and Hongjue Li and Yue Deng},
  booktitle = {Forty-third International Conference on Machine Learning},
  title     = {Plug-and-Play Guidance for Discrete Diffusion Models via Gradient-Informed Logit Correction},
  url       = {https://openreview.net/forum?id=IXsOS3hF8Y},
  year      = {2026}
}

@article{erdos59a,
  author    = {Erd\"{o}s, P. and R\'{e}nyi, A.},
  journal   = {Publicationes Mathematicae Debrecen},
  pages     = 290,
  title     = {On Random Graphs I},
  volume    = 6,
  year      = 1959
}

@article{grassia2024robustness,
  author        = {Artime, Oriol and Grassia, Marco and De Domenico, Manlio and Gleeson, James P. and Makse, Hern{\'a}n A. and Mangioni, Giuseppe and Perc, Matja{\v z} and Radicchi, Filippo},
  date          = {2024/02/01},
  doi           = {10.1038/s42254-023-00676-y},
  id            = {Artime2024},
  isbn          = {2522-5820},
  journal       = {Nature Reviews Physics},
  number        = {2},
  pages         = {114--131},
  title         = {Robustness and resilience of complex networks},
  url           = {https://doi.org/10.1038/s42254-023-00676-y},
  volume        = {6},
  year          = {2024}
}

@article{grassia2026machineenhanced,
  author        = {Grassia, Marco and d'Andrea, Valeria and Finc, Karolina and De Domenico, Manlio and Mangioni, Giuseppe},
  date          = {2026/04/09},
  doi           = {10.1038/s41598-026-47391-z},
  id            = {Grassia2026},
  isbn          = {2045-2322},
  journal       = {Scientific Reports},
  number        = {1},
  pages         = {16173},
  title         = {Machine-enhanced reconstruction of functional connectomes unravels discriminative brain sub-systems in health and disease},
  url           = {https://doi.org/10.1038/s41598-026-47391-z},
  volume        = {16},
  year          = {2026}
}

@misc{haefeli2023diffusionmodelsgraphsbenefit,
  archiveprefix = {arXiv},
  author        = {Kilian Konstantin Haefeli and Karolis Martinkus and Nathanaël Perraudin and Roger Wattenhofer},
  eprint        = {2210.01549},
  primaryclass  = {cs.LG},
  title         = {Diffusion Models for Graphs Benefit From Discrete State Spaces},
  url           = {https://arxiv.org/abs/2210.01549},
  year          = {2023}
}

@misc{ho2022classifierfreediffusionguidance,
  archiveprefix = {arXiv},
  author        = {Jonathan Ho and Tim Salimans},
  eprint        = {2207.12598},
  primaryclass  = {cs.LG},
  title         = {Classifier-Free Diffusion Guidance},
  url           = {https://arxiv.org/abs/2207.12598},
  year          = {2022}
}

@article{HOLLAND1983109,
  author   = {Paul W. Holland and Kathryn Blackmond Laskey and Samuel Leinhardt},
  doi      = {https://doi.org/10.1016/0378-8733(83)90021-7},
  issn     = {0378-8733},
  journal  = {Social Networks},
  number   = {2},
  pages    = {109-137},
  title    = {Stochastic blockmodels: First steps},
  url      = {https://www.sciencedirect.com/science/article/pii/0378873383900217},
  volume   = {5},
  year     = {1983}
}

@inproceedings{ICLR2025_597254dc,
  author    = {Nisonoff, Hunter and Xiong, Junhao and Allenspach, Stephan and Listgarten, Jennifer},
  booktitle = {International Conference on Learning Representations},
  editor    = {Y. Yue and A. Garg and N. Peng and F. Sha and R. Yu},
  pages     = {36052--36106},
  title     = {Unlocking Guidance for Discrete State-Space Diffusion and Flow Models},
  url       = {https://proceedings.iclr.cc/paper_files/paper/2025/file/597254dc45be8c166d3ccf0ba2d56325-Paper-Conference.pdf},
  volume    = {2025},
  year      = {2025}
}

@article{JMLR:v13:gretton12a,
  author  = {Arthur Gretton and Karsten M. Borgwardt and Malte J. Rasch and Bernhard Sch{{\"o}}lkopf and Alexander Smola},
  journal = {Journal of Machine Learning Research},
  number  = {25},
  pages   = {723-773},
  title   = {A Kernel Two-Sample Test},
  url     = {http://jmlr.org/papers/v13/gretton12a.html},
  volume  = {13},
  year    = {2012}
}

@inproceedings{Jo2022ScorebasedGM,
  author    = {Jaehyeong Jo and Seul Lee and Sung Ju Hwang},
  booktitle = {International Conference on Machine Learning},
  title     = {Score-based Generative Modeling of Graphs via the System of Stochastic Differential Equations},
  url       = {https://api.semanticscholar.org/CorpusID:246634850},
  year      = {2022}
}

@article{journals/corr/abs-2412-10193,
  author    = {Schiff, Yair and Sahoo, Subham Sekhar and Phung, Hao and Wang, Guanghan and Boshar, Sam and Dalla torre, Hugo and de Almeida, Bernardo P. and Rush, Alexander and Pierrot, Thomas and Kuleshov, Volodymyr},
  ee        = {https://doi.org/10.48550/arXiv.2412.10193},
  journal   = {CoRR},
  title     = {Simple Guidance Mechanisms for Discrete Diffusion Models.},
  url       = {http://dblp.uni-trier.de/db/journals/corr/corr2412.html#abs-2412-10193},
  volume    = {abs/2412.10193},
  year      = 2024
}

@misc{kipf2016variationalgraphautoencoders,
  archiveprefix = {arXiv},
  author        = {Thomas N. Kipf and Max Welling},
  eprint        = {1611.07308},
  primaryclass  = {stat.ML},
  title         = {Variational Graph Auto-Encoders},
  url           = {https://arxiv.org/abs/1611.07308},
  year          = {2016}
}

@misc{kipf2017semisupervisedclassificationgraphconvolutional,
  archiveprefix = {arXiv},
  author        = {Thomas N. Kipf and Max Welling},
  eprint        = {1609.02907},
  primaryclass  = {cs.LG},
  title         = {Semi-Supervised Classification with Graph Convolutional Networks},
  url           = {https://arxiv.org/abs/1609.02907},
  year          = {2017}
}

@article{Lancichinetti_2008,
  author    = {Lancichinetti, Andrea and Fortunato, Santo and Radicchi, Filippo},
  doi       = {10.1103/physreve.78.046110},
  issn      = {1550-2376},
  journal   = {Physical Review E},
  month     = oct,
  number    = {4},
  publisher = {American Physical Society (APS)},
  title     = {Benchmark graphs for testing community detection algorithms},
  url       = {http://dx.doi.org/10.1103/PhysRevE.78.046110},
  volume    = {78},
  year      = {2008}
}

@article{massey1951,
  author    = {Massey, Frank J},
  journal   = {Journal of the American Statistical Association},
  number    = {253},
  pages     = {68--78},
  publisher = {American Statistical Association},
  title     = {The {K}olmogorov-{S}mirnov test for goodness of fit},
  volume    = {46},
  year      = {1951}
}

@article{Muscoloni_2018,
  author    = {Muscoloni, Alessandro and Cannistraci, Carlo Vittorio},
  doi       = {10.1088/1367-2630/aac06f},
  journal   = {New Journal of Physics},
  month     = {may},
  number    = {5},
  pages     = {052002},
  publisher = {IOP Publishing},
  title     = {A nonuniform popularity-similarity optimization (nPSO) model to efficiently generate realistic complex networks with communities},
  url       = {https://dx.doi.org/10.1088/1367-2630/aac06f},
  volume    = {20},
  year      = {2018}
}

@inproceedings{NIPS2002_c3e4035a,
  author    = {Xing, Eric and Jordan, Michael and Russell, Stuart J and Ng, Andrew},
  booktitle = {Advances in Neural Information Processing Systems},
  editor    = {S. Becker and S. Thrun and K. Obermayer},
  pages     = {},
  publisher = {MIT Press},
  title     = {Distance Metric Learning with Application to Clustering with Side-Information},
  url       = {https://proceedings.neurips.cc/paper_files/paper/2002/file/c3e4035af2a1cde9f21e1ae1951ac80b-Paper.pdf},
  volume    = {15},
  year      = {2002}
}

@inproceedings{obray2022evaluation,
  author    = {Leslie O'Bray and Max Horn and Bastian Rieck and Karsten Borgwardt},
  booktitle = {International Conference on Learning Representations},
  title     = {Evaluation Metrics for Graph Generative Models: Problems, Pitfalls, and Practical Solutions},
  url       = {https://openreview.net/forum?id=tBtoZYKd9n},
  year      = 2022
}

@inproceedings{pmlr-v108-niu20a,
  author    = {Niu, Chenhao and Song, Yang and Song, Jiaming and Zhao, Shengjia and Grover, Aditya and Ermon, Stefano},
  booktitle = {Proceedings of the Twenty Third International Conference on Artificial Intelligence and Statistics},
  editor    = {Chiappa, Silvia and Calandra, Roberto},
  month     = {26--28 Aug},
  pages     = {4474--4484},
  publisher = {PMLR},
  series    = {Proceedings of Machine Learning Research},
  title     = {Permutation Invariant Graph Generation via Score-Based Generative Modeling},
  url       = {https://proceedings.mlr.press/v108/niu20a.html},
  volume    = {108},
  year      = {2020}
}

@inproceedings{pmlr-v37-sohl-dickstein15,
  address   = {Lille, France},
  author    = {Sohl-Dickstein, Jascha and Weiss, Eric and Maheswaranathan, Niru and Ganguli, Surya},
  booktitle = {Proceedings of the 32nd International Conference on Machine Learning},
  editor    = {Bach, Francis and Blei, David},
  month     = {07--09 Jul},
  pages     = {2256--2265},
  publisher = {PMLR},
  series    = {Proceedings of Machine Learning Research},
  title     = {Deep Unsupervised Learning using Nonequilibrium Thermodynamics},
  url       = {https://proceedings.mlr.press/v37/sohl-dickstein15.html},
  volume    = {37},
  year      = {2015}
}

@inproceedings{pmlr-v80-xu18c,
  author    = {Xu, Keyulu and Li, Chengtao and Tian, Yonglong and Sonobe, Tomohiro and Kawarabayashi, Ken-ichi and Jegelka, Stefanie},
  booktitle = {Proceedings of the 35th International Conference on Machine Learning},
  editor    = {Dy, Jennifer and Krause, Andreas},
  month     = {10--15 Jul},
  pages     = {5453--5462},
  publisher = {PMLR},
  series    = {Proceedings of Machine Learning Research},
  title     = {Representation Learning on Graphs with Jumping Knowledge Networks},
  url       = {https://proceedings.mlr.press/v80/xu18c.html},
  volume    = {80},
  year      = {2018}
}

@misc{romano2026mmdevaluatinggraphgenerative,
  archiveprefix = {arXiv},
  author        = {Salvatore Romano and Marco Grassia and Giuseppe Mangioni},
  eprint        = {2512.14241},
  primaryclass  = {cs.LG},
  title         = {Beyond MMD: Evaluating Graph Generative Models with Geometric Deep Learning},
  url           = {https://arxiv.org/abs/2512.14241},
  year          = {2026}
}

@inproceedings{Schroff_2015,
  author    = {Schroff, Florian and Kalenichenko, Dmitry and Philbin, James},
  booktitle = {2015 IEEE Conference on Computer Vision and Pattern Recognition (CVPR)},
  doi       = {10.1109/cvpr.2015.7298682},
  month     = {June},
  pages     = {815–823},
  publisher = {IEEE},
  title     = {FaceNet: A unified embedding for face recognition and clustering},
  url       = {http://dx.doi.org/10.1109/CVPR.2015.7298682},
  year      = {2015}
}

@inproceedings{sharma2024diffuse,
  author    = {Sharma, Kartik and Kumar, Srijan and Trivedi, Rakshit},
  booktitle = {Forty-first International Conference on Machine Learning},
  title     = {Diffuse, Sample, Project: Plug-And-Play Controllable Graph Generation},
  url       = {https://openreview.net/forum?id=ia0Z8d1DbY},
  year      = {2024}
}

@misc{simonovsky2018graphvaegenerationsmallgraphs,
  archiveprefix = {arXiv},
  author        = {Martin Simonovsky and Nikos Komodakis},
  eprint        = {1802.03480},
  primaryclass  = {cs.LG},
  title         = {GraphVAE: Towards Generation of Small Graphs Using Variational Autoencoders},
  url           = {https://arxiv.org/abs/1802.03480},
  year          = {2018}
}

@misc{tenorio2025graphguideddiffusionunified,
  archiveprefix = {arXiv},
  author        = {Victor M. Tenorio and Nicolas Zilberstein and Santiago Segarra and Antonio G. Marques},
  eprint        = {2505.19685},
  primaryclass  = {cs.LG},
  title         = {Graph Guided Diffusion: Unified Guidance for Conditional Graph Generation},
  url           = {https://arxiv.org/abs/2505.19685},
  year          = {2025}
}

@misc{veličković2018graphattentionnetworks,
  archiveprefix = {arXiv},
  author        = {Petar Veličković and Guillem Cucurull and Arantxa Casanova and Adriana Romero and Pietro Liò and Yoshua Bengio},
  eprint        = {1710.10903},
  primaryclass  = {stat.ML},
  title         = {Graph Attention Networks},
  url           = {https://arxiv.org/abs/1710.10903},
  year          = {2018}
}

@inproceedings{zhao2025adaptive,
  author    = {Jialin Zhao and Alessandro Muscoloni and Umberto Michieli and Yingtao Zhang and Carlo Vittorio Cannistraci},
  booktitle = {The Thirty-ninth Annual Conference on Neural Information Processing Systems},
  title     = {Adaptive Cannistraci-Hebb Network Automata Modelling of Complex Networks for Path-based Link Prediction},
  url       = {https://openreview.net/forum?id=fxxMReBRhi},
  year      = {2025}
}

\clearpage
\appendix

\section{ProtoGuide Algorithm and Runtime}\label{app:algorithm}

Algorithm~\ref{alg:protoguide} summarizes the guided reverse sampling procedure described in Section~\ref{sec:guidance_framework}.

\begin{algorithm}[ht!]
\small
\caption{ProtoGuide: Guided Reverse Sampling}
\label{alg:protoguide}
\begin{algorithmic}[1]
\REQUIRE Frozen denoiser $p_\theta$, frozen Siamese GNN $f_\phi$,
         target class $y^*$, guidance scale $\lambda$, diffusion steps $T$
\ENSURE  Generated graph $G$

\STATE \textbf{Offline:} Compute class prototypes
       $\proto_c = \overline{\emb}_c / \|\overline{\emb}_c\|_2$
       for all $c$ \hfill (Eq.~\ref{eq:prototype})

\STATE Sample initial state $\mathbf{x}_T \sim q(\mathbf{x}_T)$

\FOR{$t = T, T{-}1, \dotsc, 1$}
  \STATE Predict denoiser output $\mathbf{o}_{ij}(t)$ from $p_\theta$
  \STATE $\rho_t \gets 1 - t/T$ \hfill \COMMENT{generation progress}
  \IF{$\rho_t < 0.05$}
    \STATE Sample $\mathbf{x}_{t-1} \sim p_\theta(\mathbf{x}_{t-1} \mid \mathbf{x}_t)$;
           \textbf{continue} \hfill \COMMENT{skip early noisy steps}
  \ENDIF
  \STATE $\alpha \gets \tfrac{1}{2}\bigl(1 - \cos(\pi\,\rho_t)\bigr)$
         \hfill (Eq.~\ref{eq:annealing})
  \STATE Build soft adjacency
         $\softaij(t)$ from $\mathbf{o}_{ij}(t)$
         \hfill (Eq.~\ref{eq:edge_softadj} or \ref{eq:digress_softadj})
  \STATE Hard edge index
         $\mathbf{E}_{\mathrm{hard}} = \{(i,j) : \softaij > 0.5\}$
   \STATE Compute hybrid node features
         $\featmat \in \mathbb{R}^{N \times 4}$:
         degree, $\chi$-degree, clustering from $\softadj$
         (differentiable); \\ $k$-core from $\mathbf{E}_{\mathrm{hard}}$
         (detached)
         \hfill (Eqs.~\ref{eq:chi}--\ref{eq:soft_clust})
  \STATE Embed: $\emb = f_\phi(\featmat,\, \mathbf{E}_{\mathrm{hard}})$
  \STATE Contrastive score:
         $S = \cos(\emb, \proto_{y^*})
              - \max_{c \neq y^*} \cos(\emb, \proto_c)$
         \hfill (Eq.~\ref{eq:score})
  \STATE $\mathbf{g}_{ij} \gets \partial S / \partial \mathbf{o}_{ij}(t)$
         \hfill \COMMENT{autograd}
  \STATE $\hat{\mathbf{g}}_{ij} \gets
         \mathbf{g}_{ij} / (\|\mathbf{g}_{ij}\|_2+\epsilon)$
         \hfill \COMMENT{per-edge normalization}
  \STATE $\widetilde{\mathbf{o}}_{ij}(t) \gets
         \mathbf{o}_{ij}(t) + \lambda\,\alpha\,\hat{\mathbf{g}}_{ij}$
         \hfill (Eq.~\ref{eq:edge_update} or \ref{eq:digress_update})
  \STATE Renormalize $\widetilde{\mathbf{o}}_{ij}(t)$;
         sample $\mathbf{x}_{t-1}$
\ENDFOR
\RETURN $G = \mathbf{x}_0$
\end{algorithmic}
\end{algorithm}

\subsection{Wall-Clock Runtime}\label{app:runtime}

Table~\ref{t:runtime} reports the wall-clock time to generate a batch of $64$ graphs ($T{=}128$ diffusion steps) on a single NVIDIA H200 GPU, averaged over $5$ runs.
ProtoGuide's per-step overhead is dominated by the Siamese GNN forward--backward pass and the differentiable clustering feature (Section~\ref{p:complexity}), so its absolute cost scales with graph size rather than with the backbone.
For EDGE Baseline, whose unguided sampling takes approximately $1.0$--$5.4$~s per batch, guidance adds $22$--$29{\times}$ relative overhead.
For DiGress, whose Graph Transformer denoiser is the computational bottleneck, the same guidance mechanism adds $1.1$--$1.7{\times}$.

\begin{table}[ht!]
\centering\small
\caption{\textbf{Wall-clock runtime} (seconds, mean $\pm$ std over 5 runs)
  for generating 64 graphs ($T{=}128$) on a single NVIDIA H200 GPU.
  Ratio\,=\,ProtoGuide\,/\,Baseline.}
\label{t:runtime}
\rowcolors{2}{gray!25}{white}
\begin{tabular}{llrrr}
\toprule
Model & Class & Baseline (s) & ProtoGuide (s) & Ratio \\
\midrule
EDGE & Biological     & $1.5 {\pm} 0.2$    & $35.0 {\pm} 4.7$    & $23.5{\times}$ \\
     & Connectome     & $1.3 {\pm} 0.1$    & $37.8 {\pm} 4.2$    & $28.9{\times}$ \\
     & Infrastructure & $1.0 {\pm} 0.2$    & $23.6 {\pm} 3.6$    & $23.2{\times}$ \\
     & Internet       & $5.4 {\pm} 0.6$    & $121.0 {\pm} 15.3$  & $22.4{\times}$ \\
     & Social         & $1.4 {\pm} 0.3$    & $35.9 {\pm} 5.4$    & $25.9{\times}$ \\
\midrule
DiGress & Biological     & $142.5 {\pm} 18.2$   & $170.7 {\pm} 17.4$   & $1.2{\times}$ \\
        & Connectome     & $78.9 {\pm} 32.4$    & $135.0 {\pm} 42.9$   & $1.7{\times}$ \\
        & Infrastructure & $99.0 {\pm} 32.5$    & $129.5 {\pm} 31.4$   & $1.3{\times}$ \\
        & Internet       & $1{,}038.4 {\pm} 48.4$  & $1{,}108.8 {\pm} 49.1$ & $1.1{\times}$ \\
        & Social         & $415.7 {\pm} 113.9$  & $468.9 {\pm} 111.0$  & $1.1{\times}$ \\
\bottomrule
\end{tabular}
\end{table}

\section{Model Configurations}\label{app:model_config}

This appendix reports the exact per-class configuration used for EDGE (Table~\ref{t:edge_setup}) and DiGress (Table~\ref{t:digress_setup}): the number of training graphs, the maximum node count $N_{\max}$, and the guidance scale $\lambda$ selected as described in Section~\ref{sec:exp_setup} (paragraph ``Guidance Scale selection''). The training-graph counts are smaller than the class totals in Table~\ref{t:dataset_info} because each model discards, per class, the graphs that exceed its own $N_{\max}$; DiGress's smaller $N_{\max}$ on Biological, Connectome, and Infrastructure ($500$, versus $1{,}000$--$3{,}000$ for EDGE) is why its training sets for these classes are correspondingly smaller.
\begin{table}[ht!]
\centering\small
\caption{Per-class EDGE configuration.}
\label{t:edge_setup}
\rowcolors{2}{gray!25}{white}

\begin{tabular}{lccc}
\toprule
Class & Training Graphs & Max Nodes & Guidance Scale $\lambda$ \\
\midrule
Biological     & 135 & 2\,000 & 0.25 \\
Connectome     & 527 & 1\,000 & 5.0  \\
Infrastructure & 296 & 2\,000 & 1.0  \\
Internet       & 97  & 3\,000 & 3.0  \\
Social         & 85  & 1\,000 & 2.0  \\
\bottomrule
\end{tabular}
\end{table}

\begin{table}[ht!]
\centering\small
\caption{Per-class DiGress configuration.}
\label{t:digress_setup}
\rowcolors{2}{gray!25}{white}

\begin{tabular}{lccc}
\toprule
Class & Training Graphs & Max Nodes & Guidance Scale $\lambda$ \\
\midrule
Biological     & 97 & 500 & 0.50 \\
Connectome     & 418 & 500 & 3.0  \\
Infrastructure & 226 & 500 & 2.0  \\
Internet       & 52  & 1\,000 & 0.2  \\
Social         & 68  & 1\,000 & 3.5  \\
\bottomrule
\end{tabular}
\end{table}

\section{Siamese Encoder Selection}\label{app:encoder_grid}

The Siamese encoder was selected by an exhaustive grid search over $192$ configurations: architecture (GAT, GIN), depth $L\in\{2,3,4,5\}$, hidden dimension $d_h\in\{8,16,32,64\}$, learning rate $\in\{10^{-3},3\!\cdot\!10^{-4},10^{-4}\}$,
and triplet margin $\in\{0.5,1.0\}$. The output dimension ($d_o{=}16$) and the
number of attention heads ($H{=}4$, GAT only) were held fixed. Each
configuration was trained with triplet margin loss for at most $40$ epochs with
early stopping (patience $10$), and scored by validation balanced accuracy under
the same $k$-NN protocol used at evaluation time.

\begin{table}[h!]
\centering\small
\caption{\textbf{Encoder architecture comparison.} Validation balanced accuracy
  (\%) over $96$ configurations per architecture.}
\label{t:encoder_arch}
\rowcolors{2}{gray!25}{white}
\begin{tabular}{lcccc}
\toprule
Architecture & Configurations & Best & Mean & Median \\
\midrule
GAT & 96 & $\mathbf{88.81}$ & 78.89 & 78.54 \\
GIN & 96 & 84.00 & $\mathbf{81.44}$ & $\mathbf{81.39}$ \\
\bottomrule
\end{tabular}
\end{table}

\begin{table}[h!]
\centering\small
\caption{\textbf{Best validation balanced accuracy (\%) by architecture and
  depth.} GAT attains the higher value at every depth.}
\label{t:encoder_depth}
\rowcolors{2}{gray!25}{white}
\begin{tabular}{lcccc}
\toprule
Architecture & $L{=}2$ & $L{=}3$ & $L{=}4$ & $L{=}5$ \\
\midrule
GAT & $\mathbf{86.27}$ & $\mathbf{87.56}$ & $\mathbf{88.53}$ & $\mathbf{88.81}$ \\
GIN & 82.53 & 83.13 & 83.73 & 84.00 \\
\bottomrule
\end{tabular}
\end{table}

GAT reaches the higher ceiling at every depth (Table~\ref{t:encoder_depth}),
whereas GIN attains the higher mean and median across the grid
(Table~\ref{t:encoder_arch}): the GAT encoder is more sensitive to its
hyperparameters but achieves better optima when well configured. The two
best configurations overall are both GAT: $L{=}5$, $d_h{=}64$ ($88.81\%$,
$2.39$M parameters) and $L{=}4$, $d_h{=}8$ ($88.53\%$, $30$K parameters). We
adopt the latter, trading $0.28$~pp of validation accuracy for a $78\times$
reduction in parameter count.

\section{Full Confusion Matrices}\label{app:confusion}
Table~\ref{t:accuracy_global} reports per-class classification accuracy for both denoising models under baseline ($\lambda=0$), ProtoGuide, and DiGress built-in conditional mechanism.
Tables~\ref{t:app_edge_baseline_global}--\ref{t:app_digress_guided_dynamic} report
the full $5 {\times} 5$ confusion matrices for all experimental conditions
(Siamese GAT evaluation of generated graphs).
Rows correspond to the intended target class; columns to the predicted class.
Values are percentages of the row total; diagonal entries (correct
classifications) are \textbf{bolded}.

\begin{table}[!t]
\centering\small
\caption{\textbf{Per-class classification accuracy (\%) --- Global $k$-NN}.
         Mean $\pm$ std over 10 generation seeds ($64$ graphs each).
         \textbf{Bold} = best per class across conditions within each backbone. ${}^\dagger$\, best macro across conditions within each backbone.}
\label{t:accuracy_global}
\rowcolors{2}{gray!25}{white}
\begin{tabular}{ll cccccc}
\toprule
 & & \multicolumn{5}{c}{Target Class} & \\
\cmidrule(lr){3-7}
Model & Condition & Biological & Connectome & Infrastructure & Internet & Social & Macro \\
\midrule
EDGE
  & Baseline & $\mathbf{47.7 \pm 6.9}$ & $14.7 \pm 3.6$ & $\mathbf{46.9 \pm 9.8}$ & $88.1 \pm 5.2$ & $47.3 \pm 6.8$ & $48.9 \pm 3.1$ \\
 & ProtoGuide & $45.9 \pm 6.7$ & $\mathbf{86.7 \pm 3.4}$ & $44.1 \pm 6.4$ & $\mathbf{94.4 \pm 3.0}$ & $\mathbf{90.8 \pm 3.0}$ & ${}^\dagger{72.4 \pm 2.0}$ \\
\midrule
DiGress
 & Baseline & $30.0 \pm 9.2$ & $91.1 \pm 3.5$ & $\mathbf{92.2 \pm 5.6}$ & $95.6 \pm 3.2$ & $43.1 \pm 4.3$ & $70.4 \pm 2.3$ \\
 & ProtoGuide & $\mathbf{35.5 \pm 6.3}$ & $\mathbf{93.0 \pm 3.2}$ & $91.1 \pm 5.0$ & $\mathbf{97.7 \pm 1.6}$ & $\mathbf{87.3 \pm 4.7}$ & ${}^\dagger{80.9 \pm 1.7}$ \\
 & Conditional & $12.7 \pm 2.7$ & $68.3 \pm 4.9$ & $89.5 \pm 2.5$ & $7.0 \pm 2.8$ & $24.5 \pm 3.7$ & $40.4 \pm 1.6$ \\
\bottomrule
\end{tabular}
\end{table}

\begin{table}[!t]
\centering\small
\caption{\textbf{Global $k$-NN confusion --- EDGE Baseline ($\lambda{=}0$)}.
         Mean $\pm$ std (\%) over 10 seeds.}
\label{t:app_edge_baseline_global}
\rowcolors{2}{gray!25}{white}

\begin{tabular}{l ccccc}
\toprule
 & Biological & Connectome & Infrastructure & Internet & Social \\
\midrule
Biological      & \textbf{47.7$\pm$6.9} & 7.7$\pm$1.8 & 33.9$\pm$6.7 & 1.7$\pm$1.1 & 9.1$\pm$3.5 \\
Connectome      & 0.9$\pm$1.0 & \textbf{14.7$\pm$3.6} & 2.0$\pm$1.6 & 0.2$\pm$0.5 & 82.2$\pm$3.8 \\
Infrastructure  & 35.3$\pm$7.5 & 6.6$\pm$2.1 & \textbf{46.9$\pm$9.8} & 3.4$\pm$2.2 & 7.8$\pm$3.4 \\
Internet        & 8.0$\pm$3.9 & 0.0$\pm$0.0 & 3.9$\pm$2.3 & \textbf{88.1$\pm$5.2} & 0.0$\pm$0.0 \\
Social          & 15.8$\pm$3.0 & 20.5$\pm$4.8 & 14.5$\pm$3.9 & 1.9$\pm$1.5 & \textbf{47.3$\pm$6.8} \\
\bottomrule
\end{tabular}
\end{table}

\begin{table}[!t]
\centering\small
\caption{\textbf{Global $k$-NN confusion --- EDGE+ProtoGuide}.
         Mean $\pm$ std (\%) over 10 seeds.}
\label{t:app_edge_guided_global}
\rowcolors{2}{gray!25}{white}

\begin{tabular}{l ccccc}
\toprule
 & Biological & Connectome & Infrastructure & Internet & Social \\
\midrule
Biological      & \textbf{45.9$\pm$6.7} & 12.3$\pm$3.5 & 27.8$\pm$4.4 & 1.9$\pm$1.8 & 12.0$\pm$5.0 \\
Connectome      & 0.2$\pm$0.5 & \textbf{86.7$\pm$3.4} & 0.5$\pm$1.0 & 0.0$\pm$0.0 & 12.7$\pm$3.0 \\
Infrastructure  & 37.7$\pm$5.2 & 6.2$\pm$2.8 & \textbf{44.1$\pm$6.4} & 4.2$\pm$2.6 & 7.8$\pm$2.3 \\
Internet        & 0.9$\pm$1.0 & 1.2$\pm$0.9 & 0.2$\pm$0.5 & \textbf{94.4$\pm$3.0} & 3.3$\pm$2.9 \\
Social          & 4.5$\pm$2.7 & 2.0$\pm$1.7 & 1.9$\pm$1.7 & 0.8$\pm$1.4 & \textbf{90.8$\pm$3.0} \\
\bottomrule
\end{tabular}
\end{table}

\begin{table}[!t]
\centering\small
\caption{\textbf{Dynamic $k$-NN confusion --- EDGE Baseline ($\lambda{=}0$)}.
         Mean $\pm$ std (\%) over 10 seeds.}
\label{t:app_edge_baseline_dynamic}
\rowcolors{2}{gray!25}{white}

\begin{tabular}{l ccccc}
\toprule
 & Biological & Connectome & Infrastructure & Internet & Social \\
\midrule
Biological      & \textbf{62.5$\pm$5.5} & 0.3$\pm$0.6 & 24.2$\pm$6.3 & 1.7$\pm$1.1 & 11.2$\pm$3.6 \\
Connectome      & 2.5$\pm$1.2 & \textbf{11.6$\pm$3.8} & 1.4$\pm$1.1 & 0.2$\pm$0.5 & 84.4$\pm$3.7 \\
Infrastructure  & 46.4$\pm$8.8 & 0.2$\pm$0.5 & \textbf{39.2$\pm$10.7} & 3.4$\pm$2.2 & 10.8$\pm$3.9 \\
Internet        & 9.2$\pm$5.3 & 0.0$\pm$0.0 & 2.5$\pm$1.9 & \textbf{88.3$\pm$5.1} & 0.0$\pm$0.0 \\
Social          & 33.0$\pm$6.9 & 3.9$\pm$1.7 & 9.2$\pm$2.5 & 1.9$\pm$1.5 & \textbf{52.0$\pm$6.0} \\
\bottomrule
\end{tabular}
\end{table}

\begin{table}[!t]
\centering\small
\caption{\textbf{Dynamic $k$-NN confusion --- EDGE+ProtoGuide}.
         Mean $\pm$ std (\%) over 10 seeds.}
\label{t:app_edge_guided_dynamic}
\rowcolors{2}{gray!25}{white}

\begin{tabular}{l ccccc}
\toprule
 & Biological & Connectome & Infrastructure & Internet & Social \\
\midrule
Biological      & \textbf{61.4$\pm$6.3} & 0.5$\pm$0.7 & 19.8$\pm$3.0 & 1.9$\pm$1.8 & 16.4$\pm$6.4 \\
Connectome      & 0.8$\pm$0.8 & \textbf{85.0$\pm$3.7} & 0.3$\pm$0.6 & 0.0$\pm$0.0 & 13.9$\pm$3.4 \\
Infrastructure  & 50.2$\pm$5.6 & 0.2$\pm$0.5 & \textbf{34.4$\pm$6.7} & 4.5$\pm$2.6 & 10.8$\pm$2.5 \\
Internet        & 1.1$\pm$1.0 & 0.0$\pm$0.0 & 0.0$\pm$0.0 & \textbf{94.8$\pm$2.7} & 4.1$\pm$3.0 \\
Social          & 6.2$\pm$2.5 & 0.0$\pm$0.0 & 0.9$\pm$1.4 & 0.8$\pm$1.4 & \textbf{92.0$\pm$2.8} \\
\bottomrule
\end{tabular}
\end{table}

\begin{table}[!t]
\centering\small
\caption{\textbf{Global $k$-NN confusion --- DiGress Conditional (built-in)}.
         Mean $\pm$ std (\%) over 10 seeds.}
\label{t:app_digress_conditional_global}
\rowcolors{2}{gray!25}{white}

\begin{tabular}{l ccccc}
\toprule
 & Biological & Connectome & Infrastructure & Internet & Social \\
\midrule
Biological      & \textbf{12.7$\pm$2.7} & 21.2$\pm$2.9 & 51.6$\pm$5.4 & 2.8$\pm$1.8 & 11.7$\pm$2.2 \\
Connectome      & 4.1$\pm$1.9 & \textbf{68.3$\pm$4.9} & 7.2$\pm$2.2 & 2.0$\pm$1.2 & 18.4$\pm$3.2 \\
Infrastructure  & 6.1$\pm$3.7 & 0.8$\pm$0.8 & \textbf{89.5$\pm$2.5} & 1.4$\pm$1.6 & 2.2$\pm$1.0 \\
Internet        & 17.0$\pm$4.8 & 10.5$\pm$2.9 & 56.2$\pm$3.8 & \textbf{7.0$\pm$2.8} & 9.2$\pm$2.4 \\
Social          & 5.2$\pm$3.3 & 65.5$\pm$5.2 & 4.8$\pm$2.1 & 0.0$\pm$0.0 & \textbf{24.5$\pm$3.7} \\
\bottomrule
\end{tabular}
\end{table}

\begin{table}[!t]
\centering\small
\caption{\textbf{Global $k$-NN confusion --- DiGress Baseline ($\lambda{=}0$)}.
         Mean $\pm$ std (\%) over 10 seeds.}
\label{t:app_digress_baseline_global}
\rowcolors{2}{gray!25}{white}

\begin{tabular}{l ccccc}
\toprule
 & Biological & Connectome & Infrastructure & Internet & Social \\
\midrule
Biological      & \textbf{30.0$\pm$9.2} & 8.9$\pm$7.6 & 53.6$\pm$4.7 & 1.1$\pm$2.4 & 6.4$\pm$3.3 \\
Connectome      & 1.6$\pm$1.0 & \textbf{91.1$\pm$3.5} & 0.8$\pm$1.3 & 0.2$\pm$0.5 & 6.4$\pm$4.0 \\
Infrastructure  & 6.1$\pm$4.8 & 0.8$\pm$1.3 & \textbf{92.2$\pm$5.6} & 0.5$\pm$1.4 & 0.5$\pm$1.0 \\
Internet        & 2.2$\pm$2.0 & 0.0$\pm$0.0 & 2.2$\pm$1.7 & \textbf{95.6$\pm$3.2} & 0.0$\pm$0.0 \\
Social          & 12.8$\pm$4.5 & 21.7$\pm$6.7 & 22.3$\pm$6.4 & 0.0$\pm$0.0 & \textbf{43.1$\pm$4.3} \\
\bottomrule
\end{tabular}
\end{table}

\begin{table}[!t]
\centering\small
\caption{\textbf{Global $k$-NN confusion --- DiGress+ProtoGuide}.
         Mean $\pm$ std (\%) over 10 seeds.}
\label{t:app_digress_guided_global}
\rowcolors{2}{gray!25}{white}

\begin{tabular}{l ccccc}
\toprule
 & Biological & Connectome & Infrastructure & Internet & Social \\
\midrule
Biological      & \textbf{35.5$\pm$6.3} & 3.6$\pm$2.6 & 56.1$\pm$5.3 & 0.0$\pm$0.0 & 4.8$\pm$2.8 \\
Connectome      & 0.6$\pm$0.8 & \textbf{93.0$\pm$3.2} & 0.9$\pm$0.8 & 0.2$\pm$0.5 & 5.3$\pm$3.1 \\
Infrastructure  & 7.0$\pm$4.6 & 0.3$\pm$0.6 & \textbf{91.1$\pm$5.0} & 0.3$\pm$0.9 & 1.2$\pm$1.2 \\
Internet        & 0.9$\pm$1.2 & 0.0$\pm$0.0 & 1.4$\pm$1.1 & \textbf{97.7$\pm$1.6} & 0.0$\pm$0.0 \\
Social          & 5.6$\pm$2.8 & 5.8$\pm$2.7 & 1.2$\pm$1.2 & 0.0$\pm$0.0 & \textbf{87.3$\pm$4.7} \\
\bottomrule
\end{tabular}
\end{table}

\begin{table}[!t]
\centering\small
\caption{\textbf{Dynamic $k$-NN confusion --- DiGress Conditional (built-in)}.
         Mean $\pm$ std (\%) over 10 seeds.}
\label{t:app_digress_conditional_dynamic}
\rowcolors{2}{gray!25}{white}

\begin{tabular}{l ccccc}
\toprule
 & Biological & Connectome & Infrastructure & Internet & Social \\
\midrule
Biological      & \textbf{23.9$\pm$3.3} & 14.7$\pm$4.9 & 44.1$\pm$6.6 & 2.8$\pm$1.8 & 14.5$\pm$2.9 \\
Connectome      & 9.7$\pm$3.1 & \textbf{61.1$\pm$5.8} & 5.3$\pm$2.4 & 2.0$\pm$1.2 & 21.9$\pm$3.2 \\
Infrastructure  & 8.3$\pm$4.1 & 0.0$\pm$0.0 & \textbf{87.8$\pm$3.0} & 1.6$\pm$1.6 & 2.3$\pm$1.3 \\
Internet        & 31.9$\pm$3.3 & 2.7$\pm$1.6 & 46.7$\pm$3.9 & \textbf{7.0$\pm$2.8} & 11.7$\pm$2.9 \\
Social          & 9.5$\pm$3.2 & 60.5$\pm$5.6 & 2.7$\pm$2.1 & 0.0$\pm$0.0 & \textbf{27.3$\pm$4.3} \\
\bottomrule
\end{tabular}
\end{table}

\begin{table}[!t]
\centering\small
\caption{\textbf{Dynamic $k$-NN confusion --- DiGress Baseline ($\lambda{=}0$)}.
         Mean $\pm$ std (\%) over 10 seeds.}
\label{t:app_digress_baseline_dynamic}
\rowcolors{2}{gray!25}{white}

\begin{tabular}{l ccccc}
\toprule
 & Biological & Connectome & Infrastructure & Internet & Social \\
\midrule
Biological      & \textbf{41.4$\pm$9.7} & 3.9$\pm$7.9 & 45.0$\pm$5.4 & 1.1$\pm$2.4 & 8.6$\pm$3.3 \\
Connectome      & 2.2$\pm$1.0 & \textbf{90.0$\pm$4.9} & 0.6$\pm$1.0 & 0.2$\pm$0.5 & 7.0$\pm$5.0 \\
Infrastructure  & 8.9$\pm$5.2 & 0.0$\pm$0.0 & \textbf{90.0$\pm$5.8} & 0.5$\pm$1.4 & 0.6$\pm$1.0 \\
Internet        & 2.5$\pm$2.1 & 0.0$\pm$0.0 & 1.7$\pm$1.5 & \textbf{95.8$\pm$3.0} & 0.0$\pm$0.0 \\
Social          & 29.8$\pm$3.4 & 5.0$\pm$3.2 & 14.5$\pm$3.4 & 0.0$\pm$0.0 & \textbf{50.6$\pm$4.5} \\
\bottomrule
\end{tabular}
\end{table}

\begin{table}[!t]
\centering\small
\caption{\textbf{Dynamic $k$-NN confusion --- DiGress+ProtoGuide}.
         Mean $\pm$ std (\%) over 10 seeds.}
\label{t:app_digress_guided_dynamic}
\rowcolors{2}{gray!25}{white}

\begin{tabular}{l ccccc}
\toprule
 & Biological & Connectome & Infrastructure & Internet & Social \\
\midrule
Biological      & \textbf{48.6$\pm$6.5} & 0.3$\pm$0.6 & 45.5$\pm$4.8 & 0.0$\pm$0.0 & 5.6$\pm$3.0 \\
Connectome      & 1.6$\pm$1.4 & \textbf{91.6$\pm$3.8} & 0.8$\pm$0.8 & 0.2$\pm$0.5 & 5.9$\pm$3.6 \\
Infrastructure  & 8.9$\pm$4.0 & 0.0$\pm$0.0 & \textbf{89.1$\pm$4.7} & 0.5$\pm$1.0 & 1.6$\pm$1.4 \\
Internet        & 1.2$\pm$1.5 & 0.0$\pm$0.0 & 1.1$\pm$1.0 & \textbf{97.7$\pm$1.6} & 0.0$\pm$0.0 \\
Social          & 7.0$\pm$3.6 & 0.0$\pm$0.0 & 1.1$\pm$1.0 & 0.0$\pm$0.0 & \textbf{91.9$\pm$4.0} \\
\bottomrule
\end{tabular}
\end{table}

\section{Independent Classifier --- Per-Class Breakdown}\label{app:indep_classifier}
Table~\ref{t:indep_perclass} reports the per-class dynamic $k$-NN accuracy
of the independent Siamese GCN (trained on real graphs only) when classifying
generated graphs from each condition (mean $\pm$ std over $10$ classifier seeds).
This classifier is architecturally distinct from the guidance encoder and independently initialized, but it is trained on the same real corpus, so the two are not independent in training data. The standard deviations reported here and in Table~\ref{t:indep_classifier} are computed over classifier-initialization seeds with all generation seeds pooled, and therefore measure classifier stability rather than variability across generation runs.
\begin{table}[!t]
\centering\small
\caption{\textbf{Independent classifier --- per-class dynamic $k$-NN accuracy (\%)}.
         A Siamese GCN trained only on real graphs classifies generated graphs.
         Mean $\pm$ std over $10$ seeds. \textbf{Bold} = best per class across condition within each backbone. ${}^\dagger$\, best macro across condition within each backbone.}
\label{t:indep_perclass}
\rowcolors{2}{gray!25}{white}
\begin{tabular}{l ccccc c}
\toprule
Condition & Biological & Connectome & Infrastructure & Internet & Social & Macro \\
\midrule
EDGE Baseline       & $\mathbf{29.0 {\pm} 5.3}$ & $53.7 {\pm} 20.9$ & $\mathbf{86.2 {\pm} 4.0}$ & $\mathbf{92.6 {\pm} 5.3}$  & $54.5 {\pm} 5.6$  & $63.2 {\pm} 4.4$ \\
EDGE+ProtoGuide     & $24.4 {\pm} 6.3$ & $\mathbf{93.2 {\pm} 1.2}$  & $84.4 {\pm} 5.5$ & $76.0 {\pm} 10.8$ & $\mathbf{89.0 {\pm} 4.2}$  & ${}^\dagger{73.4 {\pm} 2.1}$ \\
\midrule
DiGress Baseline    & $16.8 {\pm} 6.0$ & $95.2 {\pm} 0.5$  & $\mathbf{94.8 {\pm} 1.8}$ & $98.5 {\pm} 1.3$  & $52.3 {\pm} 3.6$  & $71.5 {\pm} 1.2$ \\
DiGress+ProtoGuide  & $14.6 {\pm} 6.4$ & $\mathbf{96.5 {\pm} 0.5}$  & $94.7 {\pm} 1.8$ & $\mathbf{99.1 {\pm} 0.8}$  & $\mathbf{91.2 {\pm} 0.8}$  & ${}^\dagger{79.2 {\pm} 1.1}$ \\
DiGress Conditional & $\mathbf{24.0 {\pm} 3.3}$ & $67.6 {\pm} 1.3$  & $92.9 {\pm} 2.4$ & $8.5 {\pm} 1.5$   & $20.1 {\pm} 1.8$  & $42.6 {\pm} 0.7$ \\
\bottomrule
\end{tabular}
\end{table}

ProtoGuide's largest per-class gains appear on Social (EDGE: $54.5 \!\to\! 89.0\%$;
DiGress: $52.3 \!\to\! 91.2\%$) and Connectome under EDGE ($53.7 \!\to\! 93.2\%$),
mirroring the patterns observed with the Siamese GAT evaluation
(Table~\ref{t:accuracy_dynamic}).
Biological remains the hardest class under this independent evaluator, with per-class accuracy between $14\%$ and $29\%$ across conditions.
The DiGress Conditional model also performs poorly on Internet ($8.5\%$) and Social ($20.1\%$) under the independent classifier, consistent with the class-bias pattern observed in the main evaluation.

\section{Synthetic-to-Real Generalization}\label{app:gen_only}

To further examine the behavior of the generated graphs, and to avoid any circularity with the Siamese GAT used inside ProtoGuide, we train an independent Siamese GIN, selected via grid search over $24$ configurations (architecture, depth, hidden dimension), \textbf{only} on ProtoGuide-generated graphs ($5{,}120$ training, $1{,}280$ validation). The best configuration is a GIN with $4$ layers, hidden dimension $32$, and attentional pooling ($90.4\%$ validation macro accuracy). At test time, the frozen GIN embeds $122$ real graphs held out from its own training (Biological~$14$, Connectome~$54$, Infrastructure~$31$, Internet~$11$, Social~$12$; a per-class $80/10/10$ split of the real corpus) and classifies them via $k$-NN against the generated training set. This split is drawn independently of the generators' own train/test splits and of prototype computation, which uses every graph of a class below the model's node cap. The majority of these $122$ graphs therefore did contribute to denoiser training and to prototype construction, and the experiment measures transfer of class structure from generated to real graphs rather than generalization to unseen networks.

\begin{table}[!t]
\centering\small
\caption{\textbf{Synthetic-to-real generalization.}
  Macro and per-class accuracy (\%, mean $\pm$ std over $10$ seeds) on $122$ held-out real graphs for a Siamese GIN trained only on ProtoGuide-generated graphs.
  \textbf{Bold} = best per class; ${}^\dagger$ = best macro.}
\label{t:gen_only}
\rowcolors{2}{gray!25}{white}
\begin{tabular}{l c ccccc}
\toprule
$k$-NN protocol & Macro & Biological & Connectome & Infrastructure & Internet & Social \\
\midrule
Global $k$-NN   & ${}^\dagger{79.2 {\pm} 4.9}$ & $\mathbf{67.1 {\pm} 5.7}$ & $93.7 {\pm} 1.2$ & $\mathbf{72.9 {\pm} 12.2}$ & $98.2 {\pm} 3.6$ & $64.2 {\pm} 13.5$ \\
Dynamic $k$-NN  & $78.4 {\pm} 5.5$ & $57.1 {\pm} 12.8$ & $\mathbf{94.1 {\pm} 1.4}$ & $71.0 {\pm} 13.6$ & $\mathbf{100.0 {\pm} 0.0}$ & $\mathbf{70.0 {\pm} 10.0}$ \\
\bottomrule
\end{tabular}
\end{table}

Table~\ref{t:gen_only} reports the results.
A Siamese GIN trained only on ProtoGuide-generated graphs achieves $79.2\%$ macro accuracy (global $k$-NN) on real networks.
Connectome and Internet are classified nearly perfectly ($93.7\%$ and $98.2\%$), while Biological ($67.1\%$), Social ($64.2\%$), and Infrastructure ($72.9\%$) are lower and show larger uncertainty on some classes.
This indicates that ProtoGuide-generated graphs carry class-structural regularities that transfer to real networks.

Tables~\ref{t:app_gen_only_global} and~\ref{t:app_gen_only_dyn} report the full confusion matrices. Diagonal entries (correct classification) are \textbf{bolded}.

\begin{table}[!t]
\centering\small
\caption{\textbf{Global $k$-NN confusion matrix --- Synthetic-to-Real (GIN, generated-only training)}.
         Mean $\pm$ std (\%) over $10$ seeds.
         Rows = true class, columns = predicted class.}
\label{t:app_gen_only_global}
\rowcolors{2}{gray!25}{white}

\begin{tabular}{l ccccc}
\toprule
 & Biological & Connectome & Infrastructure & Internet & Social \\
\midrule
Biological      & $\mathbf{67.1\pm5.7}$ & $0.7\pm2.1$ & $19.3\pm4.6$ & $0.0\pm0.0$ & $12.9\pm5.3$ \\
Connectome      & $2.4\pm1.4$ & $\mathbf{93.7\pm1.2}$ & $0.9\pm1.2$ & $0.2\pm0.6$ & $2.8\pm1.2$ \\
Infrastructure  & $21.3\pm11.8$ & $2.6\pm1.3$ & $\mathbf{72.9\pm12.2}$ & $1.0\pm1.5$ & $2.3\pm1.5$ \\
Internet        & $0.9\pm2.7$ & $0.0\pm0.0$ & $0.0\pm0.0$ & $\mathbf{98.2\pm3.6}$ & $0.9\pm2.7$ \\
Social          & $13.3\pm5.5$ & $15.0\pm6.2$ & $5.0\pm4.1$ & $2.5\pm3.8$ & $\mathbf{64.2\pm13.5}$ \\
\bottomrule
\end{tabular}
\end{table}

\begin{table}[!t]
\centering\small
\caption{\textbf{Dynamic $k$-NN confusion matrix --- Synthetic-to-Real (GIN, generated-only training)}.
         Per-class $k_c = \lceil\sqrt{N_c}\rceil$.
         Mean $\pm$ std (\%) over $10$ seeds.
         Rows = true class, columns = predicted class.}
\label{t:app_gen_only_dyn}
\rowcolors{2}{gray!25}{white}

\begin{tabular}{l ccccc}
\toprule
 & Biological & Connectome & Infrastructure & Internet & Social \\
\midrule
Biological      & $\mathbf{57.1\pm12.8}$ & $0.7\pm2.1$ & $25.0\pm10.2$ & $0.7\pm2.1$ & $16.4\pm3.3$ \\
Connectome      & $2.4\pm0.8$ & $\mathbf{94.1\pm1.4}$ & $0.6\pm0.8$ & $0.4\pm0.7$ & $2.6\pm0.9$ \\
Infrastructure  & $22.6\pm13.6$ & $3.2\pm0.0$ & $\mathbf{71.0\pm13.6}$ & $1.3\pm1.6$ & $1.9\pm1.6$ \\
Internet        & $0.0\pm0.0$ & $0.0\pm0.0$ & $0.0\pm0.0$ & $\mathbf{100.0\pm0.0}$ & $0.0\pm0.0$ \\
Social          & $14.2\pm7.5$ & $10.8\pm3.8$ & $2.5\pm3.8$ & $2.5\pm3.8$ & $\mathbf{70.0\pm10.0}$ \\
\bottomrule
\end{tabular}
\end{table}

\section{Evaluation}\label{sec:evaluation_plus}

\subsection{Kolmogorov--Smirnov Test}

For each seed and each of the nine structural descriptors listed in Section~\ref{sec:results}, we run a two-sample Kolmogorov--Smirnov test comparing the metric distribution of the $64$ generated graphs against the training graphs of the respective model.
Table~\ref{t:ks_test} omits the connected-components row: a graph is connected if and only if its LCC fraction equals one, and both descriptors are dominated by this atom in our corpora, so in every comparison the KS supremum is attained exactly at the connected/disconnected boundary and the two descriptors return identical KS distances (verified on all $250$ per-seed comparisons). We therefore report only the LCC fraction, while KS-Cal@95 aggregates over all nine descriptors.
Table~\ref{t:ks_test} reports the mean KS statistic $D {\pm} \sigma$ over $10$ seeds;
\textbf{bold} entries indicate that $D$ is not statistically significant ($p > 0.05$) in at least $50\%$ of seeds. The KS-Cal@95 metric in Table~\ref{t:structural_fidelity} is derived from the same per-seed KS distance but uses a calibrated threshold: for each class and descriptor, we bootstrap $1\,000$ real-vs-real splits and take the 95th percentile of the resulting KS distances as the acceptance threshold, so that pass rates are comparable across descriptors with different intrinsic scales.

\begin{table}[h!]
\centering
\caption{Two-sample Kolmogorov--Smirnov test ($D \pm \sigma$ over 10 seeds). Each condition is tested against its own model's training set.
\textbf{Bold} indicates $D$ is not significant at $\alpha=0.05$ in $\geq$50\% of seeds (pass-rate $\geq$50\%).}
\label{t:ks_test}
\resizebox{\textwidth}{!}{%
\rowcolors{2}{gray!25}{white}
\begin{tabular}{llccccc}
\toprule
Class & Metric & EDGE & EDGE+ProtoGuide & DiGress & DiGress+ProtoGuide & DiGress-Conditional \\
\midrule
Biological & $\log N$ & $\bm{0.110 {\pm} 0.029}$ & $\bm{0.120 {\pm} 0.040}$ & $\bm{0.190 {\pm} 0.079}$ & $\bm{0.120 {\pm} 0.015}$ & $0.319 {\pm} 0.030$ \\
 & Density & $\bm{0.125 {\pm} 0.032}$ & $\bm{0.120 {\pm} 0.035}$ & $0.509 {\pm} 0.064$ & $0.513 {\pm} 0.028$ & $0.418 {\pm} 0.052$ \\
 & Assort. & $0.268 {\pm} 0.046$ & $0.270 {\pm} 0.052$ & $\bm{0.144 {\pm} 0.067}$ & $\bm{0.123 {\pm} 0.021}$ & $\bm{0.188 {\pm} 0.060}$ \\
 & Avg $k$-core & $\bm{0.167 {\pm} 0.031}$ & $\bm{0.154 {\pm} 0.023}$ & $0.666 {\pm} 0.146$ & $0.742 {\pm} 0.031$ & $0.354 {\pm} 0.034$ \\
 & Transit. & $0.400 {\pm} 0.044$ & $0.460 {\pm} 0.040$ & $0.333 {\pm} 0.061$ & $0.288 {\pm} 0.032$ & $0.368 {\pm} 0.039$ \\
 & Avg Clust. & $0.388 {\pm} 0.047$ & $0.459 {\pm} 0.047$ & $0.358 {\pm} 0.051$ & $0.346 {\pm} 0.032$ & $0.379 {\pm} 0.043$ \\
 & LCC frac. & $0.422 {\pm} 0.077$ & $0.366 {\pm} 0.044$ & $0.467 {\pm} 0.231$ & $0.581 {\pm} 0.052$ & $\bm{0.027 {\pm} 0.014}$ \\
 & Glob. Eff. & $\bm{0.167 {\pm} 0.020}$ & $\bm{0.176 {\pm} 0.022}$ & $0.576 {\pm} 0.054$ & $0.592 {\pm} 0.048$ & $0.536 {\pm} 0.042$ \\
\midrule
Connectome & $\log N$ & $\bm{0.031 {\pm} 0.009}$ & $\bm{0.030 {\pm} 0.007}$ & $\bm{0.033 {\pm} 0.006}$ & $\bm{0.078 {\pm} 0.039}$ & $0.297 {\pm} 0.057$ \\
 & Density & $\bm{0.137 {\pm} 0.044}$ & $\bm{0.167 {\pm} 0.035}$ & $0.194 {\pm} 0.044$ & $0.376 {\pm} 0.059$ & $0.345 {\pm} 0.041$ \\
 & Assort. & $0.405 {\pm} 0.052$ & $0.262 {\pm} 0.023$ & $\bm{0.159 {\pm} 0.055}$ & $0.492 {\pm} 0.048$ & $0.496 {\pm} 0.049$ \\
 & Avg $k$-core & $\bm{0.135 {\pm} 0.040}$ & $\bm{0.126 {\pm} 0.029}$ & $\bm{0.167 {\pm} 0.053}$ & $0.707 {\pm} 0.054$ & $0.369 {\pm} 0.045$ \\
 & Transit. & $0.813 {\pm} 0.032$ & $\bm{0.180 {\pm} 0.054}$ & $0.216 {\pm} 0.050$ & $0.902 {\pm} 0.020$ & $0.302 {\pm} 0.061$ \\
 & Avg Clust. & $0.942 {\pm} 0.005$ & $0.558 {\pm} 0.051$ & $0.526 {\pm} 0.032$ & $0.776 {\pm} 0.034$ & $0.529 {\pm} 0.054$ \\
 & LCC frac. & $\bm{0.008 {\pm} 0.008}$ & $\bm{0.003 {\pm} 0.006}$ & $\bm{0.003 {\pm} 0.006}$ & $\bm{0.011 {\pm} 0.016}$ & $\bm{0.011 {\pm} 0.010}$ \\
 & Glob. Eff. & $0.497 {\pm} 0.041$ & $\bm{0.168 {\pm} 0.051}$ & $0.261 {\pm} 0.048$ & $\bm{0.133 {\pm} 0.038}$ & $0.381 {\pm} 0.045$ \\
\midrule
Infrastructure & $\log N$ & $\bm{0.109 {\pm} 0.033}$ & $\bm{0.124 {\pm} 0.024}$ & $\bm{0.239 {\pm} 0.204}$ & $\bm{0.247 {\pm} 0.184}$ & $0.358 {\pm} 0.202$ \\
 & Density & $\bm{0.167 {\pm} 0.036}$ & $\bm{0.183 {\pm} 0.072}$ & $0.235 {\pm} 0.097$ & $0.219 {\pm} 0.116$ & $\bm{0.203 {\pm} 0.126}$ \\
 & Assort. & $0.588 {\pm} 0.037$ & $0.615 {\pm} 0.054$ & $0.343 {\pm} 0.078$ & $0.353 {\pm} 0.140$ & $0.273 {\pm} 0.053$ \\
 & Avg $k$-core & $0.254 {\pm} 0.043$ & $0.272 {\pm} 0.056$ & $0.394 {\pm} 0.124$ & $0.379 {\pm} 0.174$ & $0.469 {\pm} 0.121$ \\
 & Transit. & $0.521 {\pm} 0.034$ & $0.601 {\pm} 0.044$ & $0.369 {\pm} 0.097$ & $0.346 {\pm} 0.101$ & $0.315 {\pm} 0.118$ \\
 & Avg Clust. & $0.497 {\pm} 0.052$ & $0.531 {\pm} 0.031$ & $0.368 {\pm} 0.062$ & $0.339 {\pm} 0.096$ & $0.353 {\pm} 0.066$ \\
 & LCC frac. & $0.402 {\pm} 0.076$ & $0.472 {\pm} 0.066$ & $\bm{0.134 {\pm} 0.047}$ & $\bm{0.134 {\pm} 0.067}$ & $\bm{0.113 {\pm} 0.076}$ \\
 & Glob. Eff. & $0.304 {\pm} 0.072$ & $0.290 {\pm} 0.068$ & $0.287 {\pm} 0.140$ & $0.308 {\pm} 0.141$ & $0.399 {\pm} 0.121$ \\
\midrule
Internet & $\log N$ & $\bm{0.106 {\pm} 0.027}$ & $\bm{0.117 {\pm} 0.032}$ & $\bm{0.127 {\pm} 0.044}$ & $\bm{0.106 {\pm} 0.029}$ & $0.947 {\pm} 0.019$ \\
 & Density & $\bm{0.141 {\pm} 0.045}$ & $0.295 {\pm} 0.040$ & $\bm{0.150 {\pm} 0.055}$ & $\bm{0.163 {\pm} 0.036}$ & $0.953 {\pm} 0.020$ \\
 & Assort. & $0.669 {\pm} 0.054$ & $0.638 {\pm} 0.045$ & $0.560 {\pm} 0.038$ & $0.562 {\pm} 0.043$ & $0.687 {\pm} 0.053$ \\
 & Avg $k$-core & $0.683 {\pm} 0.044$ & $0.959 {\pm} 0.012$ & $\bm{0.216 {\pm} 0.067}$ & $\bm{0.235 {\pm} 0.042}$ & $0.469 {\pm} 0.026$ \\
 & Transit. & $0.538 {\pm} 0.056$ & $0.552 {\pm} 0.042$ & $0.631 {\pm} 0.057$ & $0.654 {\pm} 0.069$ & $0.617 {\pm} 0.050$ \\
 & Avg Clust. & $0.974 {\pm} 0.013$ & $0.302 {\pm} 0.049$ & $0.800 {\pm} 0.065$ & $0.835 {\pm} 0.042$ & $0.643 {\pm} 0.041$ \\
 & LCC frac. & $0.761 {\pm} 0.054$ & $0.522 {\pm} 0.044$ & $0.620 {\pm} 0.076$ & $0.605 {\pm} 0.048$ & $\bm{0.037 {\pm} 0.022}$ \\
 & Glob. Eff. & $0.426 {\pm} 0.051$ & $0.855 {\pm} 0.021$ & $0.284 {\pm} 0.052$ & $0.286 {\pm} 0.030$ & $0.875 {\pm} 0.036$ \\
\midrule
Social & $\log N$ & $\bm{0.110 {\pm} 0.026}$ & $\bm{0.093 {\pm} 0.016}$ & $\bm{0.090 {\pm} 0.018}$ & $\bm{0.101 {\pm} 0.023}$ & $0.316 {\pm} 0.060$ \\
 & Density & $0.273 {\pm} 0.053$ & $\bm{0.108 {\pm} 0.028}$ & $\bm{0.162 {\pm} 0.023}$ & $0.398 {\pm} 0.040$ & $0.501 {\pm} 0.061$ \\
 & Assort. & $0.297 {\pm} 0.040$ & $\bm{0.209 {\pm} 0.052}$ & $0.334 {\pm} 0.047$ & $0.301 {\pm} 0.046$ & $0.249 {\pm} 0.038$ \\
 & Avg $k$-core & $0.230 {\pm} 0.050$ & $\bm{0.165 {\pm} 0.036}$ & $0.247 {\pm} 0.034$ & $0.542 {\pm} 0.059$ & $0.413 {\pm} 0.050$ \\
 & Transit. & $0.513 {\pm} 0.056$ & $0.387 {\pm} 0.024$ & $0.386 {\pm} 0.064$ & $0.294 {\pm} 0.032$ & $0.347 {\pm} 0.066$ \\
 & Avg Clust. & $0.616 {\pm} 0.052$ & $0.475 {\pm} 0.026$ & $0.474 {\pm} 0.043$ & $0.362 {\pm} 0.048$ & $0.284 {\pm} 0.040$ \\
 & LCC frac. & $\bm{0.186 {\pm} 0.038}$ & $\bm{0.072 {\pm} 0.034}$ & $\bm{0.097 {\pm} 0.036}$ & $\bm{0.048 {\pm} 0.032}$ & $\bm{0.027 {\pm} 0.021}$ \\
 & Glob. Eff. & $0.238 {\pm} 0.056$ & $\bm{0.169 {\pm} 0.037}$ & $\bm{0.184 {\pm} 0.015}$ & $0.515 {\pm} 0.046$ & $0.540 {\pm} 0.063$ \\
\bottomrule
\end{tabular}%
}
\end{table}

The KS analysis reveals several patterns.
For $\log N$, the KS test is non-significant in at least half of the seeds for EDGE, DiGress, and ProtoGuide across all five classes, whereas DiGress Conditional fails this criterion on every class and most strongly on Internet ($D=0.947$).
EDGE also matches density comparatively well, passing on four of five classes, while DiGress baseline shows significant density differences on Biological, Connectome, and Infrastructure.
Assortativity, transitivity, and average clustering are among the more difficult descriptors across conditions.
ProtoGuide can either improve or worsen individual marginals depending on the class and descriptor.
For example, EDGE Connectome transitivity decreases from $D\!=\!0.813$ (significant in all seeds) to $D\!=\!0.180$ (non-significant in more than half of the seeds) under ProtoGuide.
DiGress Conditional shows a different pattern: its LCC-fraction test is non-significant in at least half of the seeds, while $\log N$ and density on Internet have near-maximal KS distances ($D>0.94$), indicating a severe mismatch in graph size and density.

\subsection{\texorpdfstring{Structural $k$-NN Coverage}{Structural k-NN Coverage}}
To complement the per-metric KS analysis with a multivariate assessment,
we compute a structural $k$-NN Coverage@95 metric.
For each seed, we standardize all nine metrics using the IQR (Interquartile Range) of the real distribution
(dropping any metric with zero IQR, i.e.\ constant across all real graphs), compute
leave-one-out $k$-NN distances among real graphs to establish a 95th-percentile
threshold $\tau_{95}$, and report the fraction of generated graphs whose $k$-NN
distance to the real set falls within $\tau_{95}$.
We evaluate in two ways: a fixed $k_{\mathrm{global}}{=}5$ and a per-class
$k_{\text{dyn}}{=}\lfloor\sqrt{n_{\text{real}}}\rfloor$.
Table~\ref{t:coverage} reports mean $\pm$ std over $10$ seeds.

\begin{table}[h!]
\centering
\caption{Structural $k$-NN Coverage@95 (mean $\pm$ std over 10 seeds).
Each method is evaluated against its own model's training set;
$k_{\mathrm{global}}=5$; $k_{\text{dyn}}=\lfloor\sqrt{n_{\text{real}}}\rfloor$ per class
(actual $k_{\text{dyn}}$ values range from $7$ to $22$ depending on model and class).
Higher is better. \textbf{Bold} = best per class between Baseline, ProtoGuide, and Conditional across both $k$.}
\label{t:coverage}

\resizebox{\textwidth}{!}{%
\rowcolors{2}{gray!25}{white}
\begin{tabular}{llccccc}
\toprule
Condition & $k$ & Biological & Connectome & Infrastructure & Internet & Social \\
\midrule
EDGE-Baseline & $k_{\text{glob}}$ & $98.3 {\pm} 1.1$ & $83.9 {\pm} 5.2$ & $96.2 {\pm} 2.7$ & $0.0 {\pm} 0.0$ & $99.2 {\pm} 1.0$ \\
     & $k_{\text{dyn}}$ & $\mathbf{98.9 {\pm} 0.7}$ & $94.8 {\pm} 2.4$ & $96.2 {\pm} 2.7$ & $0.0 {\pm} 0.0$ & $98.9 {\pm} 1.6$ \\
\midrule
EDGE+ProtoGuide & $k_{\text{glob}}$ & $98.1 {\pm} 1.4$ & $95.3 {\pm} 1.8$ & $\mathbf{97.0 {\pm} 2.1}$ & $0.0 {\pm} 0.0$ & $\mathbf{99.5 {\pm} 0.7}$ \\
                & $k_{\text{dyn}}$ & $98.1 {\pm} 0.9$ & $\mathbf{98.0 {\pm} 1.7}$ & $\mathbf{97.0 {\pm} 2.4}$ & $0.0 {\pm} 0.0$ & $99.2 {\pm} 0.8$ \\
\midrule
DiGress-Baseline & $k_{\text{glob}}$ & $97.8 {\pm} 4.0$ & $96.6 {\pm} 1.2$ & $\mathbf{94.7 {\pm} 5.6}$ & $71.7 {\pm} 6.3$ & $98.0 {\pm} 1.7$ \\
        & $k_{\text{dyn}}$ & $97.2 {\pm} 5.3$ & $96.7 {\pm} 0.8$ & $93.4 {\pm} 7.1$ & $70.3 {\pm} 5.8$ & $98.3 {\pm} 1.8$ \\
\midrule
DiGress+ProtoGuide & $k_{\text{glob}}$ & $\mathbf{99.8 {\pm} 0.5}$ & $92.3 {\pm} 3.6$ & $92.8 {\pm} 5.7$ & $\mathbf{73.3 {\pm} 4.5}$ & $93.3 {\pm} 2.5$ \\
                   & $k_{\text{dyn}}$ & $\mathbf{99.8 {\pm} 0.5}$ & $\mathbf{97.3 {\pm} 1.4}$ & $91.7 {\pm} 7.2$ & $71.9 {\pm} 5.0$ & $93.4 {\pm} 2.5$ \\
\midrule
DiGress-Conditional & $k_{\text{glob}}$ & $88.8 {\pm} 3.0$ & $70.2 {\pm} 5.0$ & $93.0 {\pm} 5.9$ & $0.0 {\pm} 0.0$ & $\mathbf{99.2 {\pm} 0.8}$ \\
                    & $k_{\text{dyn}}$ & $87.7 {\pm} 3.8$ & $77.0 {\pm} 5.5$ & $91.2 {\pm} 7.3$ & $0.0 {\pm} 0.0$ & $99.2 {\pm} 1.0$ \\
\bottomrule
\end{tabular}%
}
\end{table}

Coverage@95 remains high under ProtoGuide for most class--backbone pairs in Table~\ref{t:coverage}.
For EDGE, guidance maintains or improves coverage on most classes, most notably Connectome, which rises from $83.9\%$ to $95.3\%$ under $k_{\text{glob}}$.
Biological changes only marginally, while Internet remains at the coverage floor under both conditions.
For DiGress, guidance improves coverage on Biological and Internet, while Infrastructure and Social decrease by at most $5$~pp under both $k$ choices.
Connectome decreases under $k_{\text{glob}}$ ($96.6\% \!\to\! 92.3\%$) but is essentially unchanged under $k_{\text{dyn}}$ ($96.7\% \!\to\! 97.3\%$), making this result sensitive to the choice of $k$.
EDGE achieves $0.0\%$ Internet coverage under both baseline and guidance, indicating a clear gap between generated and real Internet graphs in the descriptor space used by Coverage@95.
DiGress Conditional also achieves $0.0\%$ on Internet, alongside the severe $\log N$ mismatch reported by the KS analysis.
Results on the remaining classes are broadly similar across the two $k$ choices.

\subsection{MMD Assessment}\label{app:mmd}
Table~\ref{t:mmd_perclass} reports five MMD metrics: degree, clustering, orbits, spectral, and NSPDK~\cite{10.5555/3104322.3104356}. In this way, we capture distributional fidelity at different structural scales.

\begin{table}[h!]
\centering
\caption{\textbf{MMD per class and condition} (mean $\pm$ std over 10 generation seeds).
  Each cell reports the Maximum Mean Discrepancy between 64 generated graphs and the
  class-specific reference set. Lower is better. \textbf{Bold} = best per class across conditions within each backbone.}
\label{t:mmd_perclass}
\rowcolors{2}{gray!25}{white}
\resizebox{\textwidth}{!}{%
\begin{tabular}{ll ccccc}
\toprule
Condition & Class & Degree $\downarrow$ & Clustering $\downarrow$ & Orbits $\downarrow$ & Spectral $\downarrow$ & NSPDK $\downarrow$ \\
\midrule
EDGE Baseline & Biological & $0.0452 {\pm} 0.0100$ & $\mathbf{0.162 {\pm} 0.028}$ & $0.0313 {\pm} 0.0056$ & $\mathbf{0.0301 {\pm} 0.0020}$ & $\mathbf{0.0255 {\pm} 0.0017}$ \\
 & Connectome & $\mathbf{0.0232 {\pm} 0.0065}$ & $0.878 {\pm} 0.045$ & $0.0177 {\pm} 0.0001$ & $1.094 {\pm} 0.042$ & $0.0189 {\pm} 0.0005$ \\
 & Infrastructure & $\mathbf{0.0575 {\pm} 0.0091}$ & $\mathbf{0.311 {\pm} 0.071}$ & $0.0329 {\pm} 0.0062$ & $0.0263 {\pm} 0.0012$ & $0.0290 {\pm} 0.0025$ \\
 & Internet & $\mathbf{0.214 {\pm} 0.015}$ & $0.778 {\pm} 0.048$ & $0.0269 {\pm} 0.0001$ & $\mathbf{0.314 {\pm} 0.014}$ & $\mathbf{0.0287 {\pm} 0.0008}$ \\
 & Social & $\mathbf{0.0332 {\pm} 0.0077}$ & $0.264 {\pm} 0.050$ & $\mathbf{0.0381 {\pm} 0.0043}$ & $0.0366 {\pm} 0.0018$ & $0.0298 {\pm} 0.0025$ \\
\midrule
EDGE +ProtoGuide & Biological & $\mathbf{0.0396 {\pm} 0.0112}$ & $0.227 {\pm} 0.042$ & $\mathbf{0.0285 {\pm} 0.0065}$ & $0.0305 {\pm} 0.0024$ & $\mathbf{0.0255 {\pm} 0.0014}$ \\
 & Connectome & $0.0366 {\pm} 0.0065$ & $\mathbf{0.0896 {\pm} 0.0198}$ & $\mathbf{0.0175 {\pm} 0.0000}$ & $\mathbf{0.235 {\pm} 0.032}$ & $\mathbf{0.0170 {\pm} 0.0005}$ \\
 & Infrastructure & $0.0699 {\pm} 0.0195$ & $0.351 {\pm} 0.065$ & $\mathbf{0.0323 {\pm} 0.0061}$ & $\mathbf{0.0253 {\pm} 0.0006}$ & $\mathbf{0.0288 {\pm} 0.0032}$ \\
 & Internet & $0.466 {\pm} 0.025$ & $\mathbf{0.477 {\pm} 0.027}$ & $\mathbf{0.0268 {\pm} 0.0000}$ & $0.320 {\pm} 0.017$ & $0.0301 {\pm} 0.0017$ \\
 & Social & $0.0426 {\pm} 0.0071$ & $\mathbf{0.178 {\pm} 0.018}$ & $0.0399 {\pm} 0.0051$ & $\mathbf{0.0345 {\pm} 0.0031}$ & $\mathbf{0.0269 {\pm} 0.0013}$ \\
\midrule
DiGress Baseline & Biological & $0.374 {\pm} 0.152$ & $0.108 {\pm} 0.048$ & $0.226 {\pm} 0.104$ & $0.0366 {\pm} 0.0032$ & $0.0764 {\pm} 0.0151$ \\
 & Connectome & $\mathbf{0.0380 {\pm} 0.0075}$ & $\mathbf{0.0780 {\pm} 0.0133}$ & $\mathbf{0.0180 {\pm} 0.0001}$ & $\mathbf{0.549 {\pm} 0.034}$ & $\mathbf{0.0253 {\pm} 0.0005}$ \\
 & Infrastructure & $0.131 {\pm} 0.080$ & $0.102 {\pm} 0.049$ & $\mathbf{0.0412 {\pm} 0.0285}$ & $0.0341 {\pm} 0.0134$ & $0.0385 {\pm} 0.0055$ \\
 & Internet & $0.105 {\pm} 0.017$ & $\mathbf{0.339 {\pm} 0.030}$ & $\mathbf{0.0366 {\pm} 0.0002}$ & $0.162 {\pm} 0.020$ & $0.0420 {\pm} 0.0019$ \\
 & Social & $\mathbf{0.0408 {\pm} 0.0071}$ & $\mathbf{0.177 {\pm} 0.035}$ & $\mathbf{0.0298 {\pm} 0.0032}$ & $\mathbf{0.0417 {\pm} 0.0012}$ & $\mathbf{0.0371 {\pm} 0.0019}$ \\
\midrule
DiGress +ProtoGuide & Biological & $0.444 {\pm} 0.030$ & $\mathbf{0.0857 {\pm} 0.0095}$ & $0.294 {\pm} 0.055$ & $\mathbf{0.0332 {\pm} 0.0018}$ & $0.0820 {\pm} 0.0060$ \\
 & Connectome & $0.269 {\pm} 0.018$ & $0.279 {\pm} 0.033$ & $0.0181 {\pm} 0.0000$ & $0.888 {\pm} 0.039$ & $0.0766 {\pm} 0.0030$ \\
 & Infrastructure & $\mathbf{0.128 {\pm} 0.080}$ & $\mathbf{0.0884 {\pm} 0.0553}$ & $0.0454 {\pm} 0.0257$ & $\mathbf{0.0331 {\pm} 0.0123}$ & $\mathbf{0.0379 {\pm} 0.0050}$ \\
 & Internet & $\mathbf{0.102 {\pm} 0.019}$ & $0.367 {\pm} 0.025$ & $\mathbf{0.0366 {\pm} 0.0001}$ & $\mathbf{0.161 {\pm} 0.018}$ & $\mathbf{0.0416 {\pm} 0.0022}$ \\
 & Social & $0.0919 {\pm} 0.0121$ & $0.265 {\pm} 0.027$ & $0.0546 {\pm} 0.0039$ & $0.0962 {\pm} 0.0139$ & $0.0737 {\pm} 0.0089$ \\
\midrule
DiGress Conditional & Biological & $\mathbf{0.0773 {\pm} 0.0129}$ & $0.137 {\pm} 0.038$ & $\mathbf{0.0363 {\pm} 0.0032}$ & $0.0365 {\pm} 0.0029$ & $\mathbf{0.0475 {\pm} 0.0026}$ \\
 & Connectome & $0.125 {\pm} 0.009$ & $0.142 {\pm} 0.039$ & $0.0228 {\pm} 0.0032$ & $0.588 {\pm} 0.022$ & $0.0426 {\pm} 0.0053$ \\
 & Infrastructure & $0.182 {\pm} 0.077$ & $0.104 {\pm} 0.043$ & $0.0631 {\pm} 0.0351$ & $0.0420 {\pm} 0.0143$ & $0.0408 {\pm} 0.0047$ \\
 & Internet & $0.585 {\pm} 0.024$ & $0.800 {\pm} 0.066$ & $0.0486 {\pm} 0.0055$ & $0.269 {\pm} 0.004$ & $0.110 {\pm} 0.003$ \\
 & Social & $0.0681 {\pm} 0.0083$ & $0.263 {\pm} 0.041$ & $0.0547 {\pm} 0.0036$ & $0.140 {\pm} 0.034$ & $0.0672 {\pm} 0.0091$ \\
\bottomrule
\end{tabular}%
}
\end{table}

ProtoGuide has a mixed and strongly class-dependent effect on MMD (Table~\ref{t:mmd_perclass}). In some cases guidance improves distributional agreement: for EDGE Connectome, clustering MMD decreases from $0.878$ to $0.0896$ and spectral MMD from $1.094$ to $0.235$. Other metrics move in the opposite direction; for example, EDGE Internet degree MMD increases from $0.214$ to $0.466$. The clearest distributional shift occurs for DiGress Connectome, where degree MMD rises from $0.0380$ to $0.269$, clustering from $0.0780$ to $0.279$, and spectral and NSPDK MMD increase as well, despite only a modest change in classification accuracy ($90.0\% \to 91.6\%$). This reinforces that class controllability and distributional fidelity capture distinct properties of the generated graphs and should be evaluated separately.

DiGress Conditional also shows large MMD values on Internet (degree $0.585$, clustering $0.800$) and elevated values on several other statistics, consistent with the poor structural fidelity observed for this condition in the complementary KS and coverage evaluations.

\section{Ablation}\label{sec:ablation}
\subsection{Effect of Guidance Scale}
\label{ss:guidance_scale}
To understand how the guidance scale $\lambda$ interacts with class-specific generative difficulty, we sweep $\lambda \in \{0,0.1, 0.25, 0.5,1,2,3,5,7,10\}$ on two contrasting classes (Connectome and Biological) using EDGE with dynamic $k$-NN evaluation (10 seeds, 64 graphs each).

\begin{figure}[!t]
    \centering
    \includegraphics[width=0.70\textwidth]{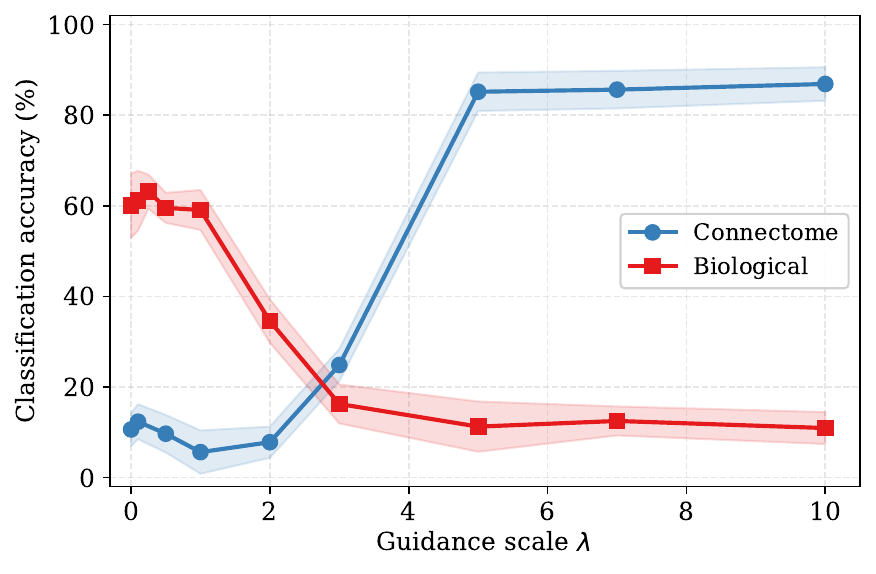}
    \caption{\textbf{Guidance scale ablation} (EDGE, dynamic $k$-NN). Connectome accuracy is close to zero without guidance and exhibits a sharp transition at $\lambda\!\approx\!5$, saturating at $\sim\!87\%$. Biological accuracy starts at $\sim\!60\%$ without guidance and degrades monotonically for $\lambda\!>\!1$, collapsing below $15\%$ at high scales. Shaded bands show $\pm1$ std over 10 seeds.}
    \label{fig:ablation_lambda}
\end{figure}

Figure~\ref{fig:ablation_lambda} reveals markedly different sensitivity to the guidance scale across the two classes. For Connectome, where the EDGE baseline achieves only $11.6\%$ accuracy, the response is threshold-like: accuracy remains low up to $\lambda\!=\!2$, rises from $24.8\%$ at $\lambda\!=\!3$ to $85.2\%$ at $\lambda\!=\!5$, and then saturates around $87\%$. Thus, for this class, weak guidance has little measurable effect on class alignment, whereas stronger updates substantially alter the generated samples.

Biological shows the opposite behavior. Its baseline accuracy is already $62.5\%$, and the sweep reaches a similar value at $\lambda\!=\!0.25$ ($63.1\%$; the small offset from the $61.4\%$ main-experiment value reflects the independent generation seeds used in this sweep) before degrading as $\lambda$ increases, reaching $11.3\%$ at $\lambda\!=\!5$. Together, the two curves show that guidance strength must be calibrated to the class: increasing $\lambda$ can be essential when the baseline is poorly aligned, but excessive guidance can substantially reduce accuracy when stronger steering is not beneficial.

\subsection{Annealing Schedule}
\label{ss:annealing}
ProtoGuide modulates the guidance scale $\lambda$ with a cosine annealing factor (Eq.~\ref{eq:annealing}) that suppresses the gradient during early noisy timesteps and increases it as the graph structure builds.
To validate this design choice, we compare three schedules: constant ($\alpha{=}1$ at every step), linear ($\alpha{=}\rho_t$), and cosine ($\alpha{=}\tfrac{1}{2}(1-\cos\pi\rho_t)$), across all five classes for EDGE under ProtoGuide, $10$~seeds, and $64$~graphs per seed.
The cosine row reproduces the results reported for the main experiment in Tables~\ref{t:accuracy_dynamic} and~\ref{t:structural_fidelity}.

\begin{table}[h!]
\centering\small
\caption{\textbf{Annealing schedule ablation} (EDGE, mean $\pm$ std over $10$ seeds).
  Dyn.\ Acc = dynamic $k$-NN accuracy; Cov@95 = structural coverage;
  KS-Cal@95 = calibrated KS pass rate; PW-Ratio = pairwise diversity ratio (gen/real, ${\approx}\,1$ is ideal).
  \textbf{Bold} = best schedule per class per metric.}
\label{tab:annealing}
\rowcolors{2}{gray!25}{white}
\begin{tabular}{ll cccc}
\toprule
Class & Schedule & Dyn.\ Acc (\%) $\uparrow$ & Cov@95 (\%) $\uparrow$ & KS-Cal@95 (\%) $\uparrow$ & PW-Ratio ${\approx}\,1$ \\
\midrule
Biological & Constant & $\mathbf{63.3 {\pm} 6.3}$ & $97.7 {\pm} 1.9$ & $44.4 {\pm} 5.0$ & $0.71 {\pm} 0.06$ \\
           & Linear   & $63.1 {\pm} 6.4$ & $98.1 {\pm} 1.8$ & $45.6 {\pm} 3.3$ & $0.72 {\pm} 0.06$ \\
           & Cosine   & $61.4 {\pm} 6.3$ & $\mathbf{98.1 {\pm} 0.9}$ & $\mathbf{46.7 {\pm} 4.4}$ & $\mathbf{0.73 {\pm} 0.06}$ \\
\midrule
Connectome & Constant & $\mathbf{87.3 {\pm} 3.5}$ & $\mathbf{98.0 {\pm} 1.2}$ & $22.2 {\pm} 7.0$ & $0.96 {\pm} 0.13$ \\
           & Linear   & $81.9 {\pm} 3.6$ & $97.7 {\pm} 1.4$ & $17.8 {\pm} 8.9$ & $\mathbf{1.01 {\pm} 0.10}$ \\
           & Cosine   & $85.0 {\pm} 3.7$ & $98.0 {\pm} 1.7$ & $\mathbf{58.9 {\pm} 14.9}$ & $1.05 {\pm} 0.14$ \\
\midrule
Infrastructure & Constant & $40.6 {\pm} 3.8$ & $94.2 {\pm} 3.0$ & $25.6 {\pm} 5.1$ & $\mathbf{0.89 {\pm} 0.40}$ \\
               & Linear   & $\mathbf{43.8 {\pm} 6.4}$ & $96.1 {\pm} 1.4$ & $\mathbf{26.7 {\pm} 5.4}$ & $0.79 {\pm} 0.31$ \\
               & Cosine   & $34.4 {\pm} 6.7$ & $\mathbf{97.0 {\pm} 2.4}$ & $18.9 {\pm} 5.1$ & $0.65 {\pm} 0.21$ \\
\midrule
Internet & Constant & $77.3 {\pm} 6.1$ & $0.0 {\pm} 0.0$ & $11.1 {\pm} 0.0$ & $2.80 {\pm} 0.54$ \\
         & Linear   & $\mathbf{95.0 {\pm} 3.5}$ & $0.0 {\pm} 0.0$ & $\mathbf{20.0 {\pm} 4.4}$ & $\mathbf{1.81 {\pm} 0.29}$ \\
         & Cosine   & $94.8 {\pm} 2.7$ & $0.0 {\pm} 0.0$ & $16.7 {\pm} 7.5$ & $1.85 {\pm} 0.19$ \\
\midrule
Social & Constant & $\mathbf{93.8 {\pm} 2.5}$ & $99.1 {\pm} 1.6$ & $54.4 {\pm} 3.3$ & $0.77 {\pm} 0.05$ \\
       & Linear   & $88.3 {\pm} 2.4$ & $98.8 {\pm} 1.2$ & $54.4 {\pm} 3.3$ & $0.79 {\pm} 0.05$ \\
       & Cosine   & $92.0 {\pm} 2.8$ & $\mathbf{99.2 {\pm} 0.8}$ & $\mathbf{57.8 {\pm} 6.7}$ & $\mathbf{0.81 {\pm} 0.02}$ \\
\bottomrule
\end{tabular}
\end{table}

Table~\ref{tab:annealing} shows that no single schedule dominates across all classes and metrics. The largest difference appears on Internet, where the constant schedule reaches only $77.3\%$ accuracy and a PW-Ratio of $2.80$, compared with $95.0\%$/$1.81$ for linear and $94.8\%$/$1.85$ for cosine. Because the constant schedule applies the full guidance scale at every active timestep, this result indicates that delaying strong guidance is beneficial for this class.

The remaining classes show a more mixed pattern. On Connectome, constant guidance attains the highest accuracy ($87.3\%$), whereas cosine obtains a substantially higher KS-Cal@95 ($58.9\%$ versus $22.2\%$ for constant and $17.8\%$ for linear). On Social, constant again gives slightly higher accuracy ($93.8\%$ versus $92.0\%$), while cosine yields the highest Coverage@95, KS-Cal@95, and PW-Ratio. Infrastructure favors the linear schedule in both accuracy and KS-Cal@95, whereas Biological shows only small accuracy differences across schedules.

We therefore retain the cosine schedule as a default overall compromise rather than because it dominates every class: it provides competitive controllability, avoids the strongest degradation observed with constant guidance on Internet, and achieves favorable structural-fidelity results on several classes.

\subsection{Contrastive vs Attraction-only}
\label{ss:attraction_vs_contrastive}
ProtoGuide uses a contrastive objective that both attracts generated graphs toward the target prototype and repels them from the nearest competing class prototype (see Eq.~\ref{eq:score}). To test whether the repulsion term is necessary, we ablate it by removing negative prototypes and compare against the full method and the unguided baseline across all five EDGE classes (10 seeds, 64 graphs each, dynamic $k$-NN). Specifically, the contrastive score is now defined as:
\begin{equation}
    S(G)=\underbrace{\scos{\backbone(G)}{\proto_{y^*}}}_{\text{attract to target}}
\end{equation}

\begin{table}[h!]
\centering
\caption{\textbf{Ablation: contrastive vs.\ attraction-only} (EDGE, dynamic $k$-NN, mean $\pm$ std over 10 seeds). Dyn.\ Acc: dynamic $k$-NN accuracy; Cov@95: structural coverage; KS-Cal@95: calibrated KS goodness-of-fit; PW-Ratio: pairwise diversity ratio (gen/real). \textbf{Bold} = best per class per metric across all conditions}
\label{tab:ablation_contrastive}
\small
\rowcolors{2}{gray!25}{white}
\resizebox{\textwidth}{!}{%
\begin{tabular}{l l cccc}
\toprule
Class & Condition & Dyn.\ Acc (\%) $\uparrow$ & Cov@95 (\%) $\uparrow$ & KS-Cal@95 (\%) $\uparrow$ & PW-Ratio \\
\midrule
Biological     & Baseline         & $62.5 \pm 5.5$  & $\mathbf{98.9 \pm 0.7}$  & $\mathbf{46.7 \pm 4.4}$  & $0.72 \pm 0.04$ \\
               & Attraction-only  & $\mathbf{63.4 \pm 5.9}$  & $98.6 \pm 1.3$  & $45.6 \pm 3.3$  & $0.71 \pm 0.05$ \\
               & ProtoGuide       & $61.4 \pm 6.3$  & $98.1 \pm 0.9$  & $\mathbf{46.7 \pm 4.4}$  & $\mathbf{0.73 \pm 0.06}$ \\
\midrule
Connectome     & Baseline         & $11.6 \pm 3.8$  & $94.8 \pm 2.4$  & $38.9 \pm 12.4$  & $\mathbf{0.99 \pm 0.19}$ \\
               & Attraction-only  & $\mathbf{92.5 \pm 3.1}$  & $95.8 \pm 2.0$  & $15.6 \pm 7.4$   & $0.81 \pm 0.12$ \\
               & ProtoGuide       & $85.0 \pm 3.7$  & $\mathbf{98.0 \pm 1.7}$  & $\mathbf{58.9 \pm 14.9}$ & $1.05 \pm 0.14$ \\
\midrule
Infrastructure & Baseline         & $\mathbf{39.2 \pm 10.7}$ & $96.2 \pm 2.7$  & $20.0 \pm 4.4$   & $\mathbf{0.85 \pm 0.37}$ \\
               & Attraction-only  & $28.1 \pm 5.8$  & $94.7 \pm 3.1$  & $\mathbf{23.3 \pm 6.0}$  & $0.77 \pm 0.29$ \\
               & ProtoGuide       & $34.4 \pm 6.7$  & $\mathbf{97.0 \pm 2.4}$  & $18.9 \pm 5.1$   & $0.65 \pm 0.21$ \\
\midrule
Internet       & Baseline         & $88.3 \pm 5.1$  & $0.0 \pm 0.0$   & $\mathbf{22.2 \pm 0.0}$  & $\mathbf{0.97 \pm 0.08}$ \\
               & Attraction-only  & $88.6 \pm 3.9$  & $0.0 \pm 0.0$   & $12.2 \pm 3.3$   & $2.29 \pm 0.47$ \\
               & ProtoGuide       & $\mathbf{94.8 \pm 2.7}$  & $0.0 \pm 0.0$   & $16.7 \pm 7.5$   & $1.85 \pm 0.19$ \\
\midrule
Social         & Baseline         & $52.0 \pm 6.0$  & $98.9 \pm 1.6$  & $34.4 \pm 14.4$  & $0.75 \pm 0.05$ \\
               & Attraction-only  & $91.6 \pm 2.0$  & $98.8 \pm 0.9$  & $55.6 \pm 0.0$   & $0.77 \pm 0.04$ \\
               & ProtoGuide       & $\mathbf{92.0 \pm 2.8}$  & $\mathbf{99.2 \pm 0.8}$  & $\mathbf{57.8 \pm 6.7}$  & $\mathbf{0.81 \pm 0.02}$ \\
\bottomrule
\end{tabular}
}
\end{table}

Table~\ref{tab:ablation_contrastive} shows that the contribution of the repulsive term is class-dependent. Attraction-only guidance performs particularly well on Connectome ($92.5\%$ versus $85.0\%$ for the full objective), whereas on Internet it provides essentially no accuracy gain over baseline ($88.6\%$ versus $88.3\%$) and the full contrastive objective reaches $94.8\%$. Social is nearly unchanged between the two guided variants, while Infrastructure remains difficult under both. Across the five classes, attraction-only reaches $72.8\%$ macro accuracy and the full contrastive objective $73.5\%$. We therefore retain the contrastive formulation as the default because it provides slightly stronger aggregate performance and benefits some classes, not because the repulsive term is universally advantageous.

\subsection{Wrong \& Random Prototypes: a Guidance Directionality Test}
\label{ss:wrong_random_prot}

Here we test whether ProtoGuide's effect depends on the identity and direction of the target prototype, rather than on the mere presence of an additional gradient signal.
We replace the correct target prototype with two alternatives and measure how the generated class distribution changes:
\begin{itemize}
    \item \textbf{Wrong-class prototype.}
    For each target class~$c$, we select the farthest competitor in cosine similarity:
    \begin{equation}
        c_{\mathrm{wrong}}=\argmin_{c'\neq c}\;\cos\!(\proto_c,\;\proto_{c'})
    \end{equation}
    This choice maximizes separation from the target in prototype space, providing a stringent test of whether guidance follows the selected class direction.
    A closer incorrect prototype would make the comparison less distinct because the two prototype directions are more similar.
    \item \textbf{Random prototype.}
    We sample a random unit vector $\mathbf{r}\in\mathbb{R}^{D}$ ($\|\mathbf{r}\|=1$) uniformly over all directions and use it as the target prototype, with all real-class prototypes as negatives.
    This creates a class-agnostic guidance direction with the same dimensionality and norm as a valid prototype.
    If arbitrary gradient steering were sufficient, random directions would be expected to produce comparable improvements; degradation instead supports a dependence on the class-specific prototype geometry.
\end{itemize}

The diffusion model and all hyperparameters remain identical to the corresponding ProtoGuide condition except for the target prototype. One exception is DiGress Internet: its selected guidance scale $\lambda{=}0.2$ produces little visible response under wrong/random targets, so those two runs use $\lambda{=}3.0$. We therefore treat the DiGress Internet rows as an elevated-strength stress test rather than a matched directional comparison and do not use them to establish the directionality claim. The EDGE Connectome, EDGE Internet, and DiGress Connectome comparisons retain the same $\lambda$ across correct, wrong, and random targets and provide the matched evidence. We evaluate Connectome and Internet with both EDGE and DiGress (5~seeds, 64~graphs each), reporting dynamic $k$-NN classification accuracy, structural Coverage@95, KS-Cal@95, and PW-Ratio.

\begin{table}[h!]
\centering
\caption{\textbf{Guidance directionality test} (mean $\pm$ std; wrong/random: 5~seeds, Baseline/ProtoGuide: 10~seeds from main experiment). Target Acc: fraction classified as the intended class; Wrong Acc: fraction classified as $c_{\mathrm{wrong}}$; Cov@95: structural plausibility; KS-Cal@95: calibrated KS goodness-of-fit; PW-Ratio: diversity ratio (gen/real). \textbf{Bold}= best value per metric across condition.  }
\label{tab:proto_sanity}
\small
\rowcolors{2}{gray!25}{white}
\resizebox{\textwidth}{!}{%
\begin{tabular}{l l ccccc}
\toprule
Condition & Class & Target Acc (\%) $\uparrow$ & Wrong Acc (\%) & Cov@95 (\%) $\uparrow$ & KS-Cal@95 (\%) $\uparrow$ & PW-Ratio \\
\midrule

EDGE Baseline ($\lambda{=}0$)      & Connectome & $11.6 \pm 3.8$ & ---             & $94.8 \pm 2.4$ & $38.9 \pm 12.4$ & $\mathbf{0.99 \pm 0.19}$ \\
EDGE ProtoGuide ($\lambda{=}5$)       & Connectome & $\mathbf{85.0 \pm 3.7}$ & ---    & $\mathbf{98.0 \pm 1.7}$ & $\mathbf{58.9 \pm 14.9}$ & $1.05 \pm 0.14$ \\
Wrong-class ($\lambda{=}5$)   & Connectome & $11.2 \pm 1.5$ & $0.0 \pm 0.0$  & $91.9 \pm 3.6$ & $13.3 \pm 8.3$ & $1.11 \pm 0.15$ \\
Random ($\lambda{=}5$)        & Connectome & $8.2 \pm 4.0$  & $0.0 \pm 0.0$  & $96.7 \pm 2.5$ & $8.9 \pm 4.4$ & $1.02 \pm 0.23$ \\
\rowcolor{white}
EDGE Baseline ($\lambda{=}0$)      & Internet   & $88.3 \pm 5.1$ & ---             & $0.0 \pm 0.0$ & $\mathbf{22.2 \pm 0.0}$ & $\mathbf{0.97 \pm 0.08}$ \\
EDGE ProtoGuide ($\lambda{=}3$)       & Internet   & $\mathbf{94.8 \pm 2.7}$ & ---    & $0.0 \pm 0.0$ & $16.7 \pm 7.5$ & $1.85 \pm 0.19$ \\
Wrong-class ($\lambda{=}3$)   & Internet   & $51.9 \pm 4.6$ & $2.8 \pm 2.3$  & $0.0 \pm 0.0$ & $11.1 \pm 0.0$ & $2.41 \pm 1.11$ \\
Random ($\lambda{=}3$)        & Internet   & $60.3 \pm 8.4$ & $4.9 \pm 3.9$  & $0.0 \pm 0.0$ & $11.1 \pm 0.0$ & $2.49 \pm 0.70$ \\
\midrule

DiGress Baseline ($\lambda{=}0$)      & Connectome & $90.0 \pm 4.9$ & ---             & $96.7 \pm 0.8$ & $\mathbf{47.8 \pm 13.2}$ & $0.93 \pm 0.09$ \\
DiGress ProtoGuide ($\lambda{=}3$)       & Connectome & $\mathbf{91.6 \pm 3.8}$ & ---    & $97.3 \pm 1.4$ & $26.7 \pm 15.9$ & $\mathbf{1.03 \pm 0.18}$ \\
Wrong-class ($\lambda{=}3$)   & Connectome & $10.6 \pm 2.7$ & $0.0 \pm 0.0$  & $\mathbf{98.4 \pm 1.4}$ & $15.6 \pm 8.9$ & $1.79 \pm 0.23$ \\
Random ($\lambda{=}3$)        & Connectome & $8.5 \pm 4.7$  & $1.4 \pm 2.1$  & $93.9 \pm 4.2$ & $25.2 \pm 6.4$ & $1.89 \pm 0.24$ \\
\rowcolor{white}
DiGress Baseline ($\lambda{=}0$)      & Internet   & $95.8 \pm 3.0$ & ---             & $70.3 \pm 5.8$ & $41.1 \pm 7.1$ & $1.10 \pm 0.09$ \\
DiGress ProtoGuide ($\lambda{=}0.2$)     & Internet   & $\mathbf{97.7 \pm 1.6}$ & ---    & $\mathbf{71.9 \pm 5.0}$ & $\mathbf{44.4 \pm 0.0}$ & $\mathbf{1.08 \pm 0.07}$ \\
Wrong-class ($\lambda{=}3$)   & Internet   & $14.7 \pm 2.5$ & $51.9 \pm 6.3$ & $0.0 \pm 0.0$ & $11.1 \pm 0.0$ & $5.04 \pm 0.43$ \\
Random ($\lambda{=}3$)        & Internet   & $8.3 \pm 12.2$ & $32.3 \pm 5.3$ & $0.1 \pm 0.4$ & $11.1 \pm 0.0$ & $5.52 \pm 2.47$ \\
\bottomrule
\end{tabular}
}
\end{table}

Table~\ref{tab:proto_sanity} supports that the guidance effect depends on the direction defined by the target prototype rather than on the mere presence of an additional gradient. In the matched Connectome experiments, replacing the correct prototype with either a wrong-class or random target reduces target accuracy sharply for both EDGE and DiGress. The same pattern appears for EDGE Internet: target accuracy falls from $94.8\%$ with the correct prototype to $51.9\%$ with the wrong prototype and $60.3\%$ with a random direction. The unmatched DiGress Internet runs show that high-strength wrong or random guidance can strongly redirect the generated class distribution, but, because $\lambda$ differs from the main ProtoGuide condition, these rows are interpreted only as a stress test.

Changing the prototype can also alter the structural distribution of the generated graphs. For example, on DiGress Connectome the PW-Ratio increases from $1.03$ with the correct prototype to $1.79$ and $1.89$ under wrong and random guidance, respectively. Taken together, the matched experiments show that ProtoGuide is target-directed: its effect depends on which prototype defines the guidance objective, rather than acting as a generic gradient regularizer.

\subsection{Few-Shot Prototype Analysis}
\label{ss:few_shot_prot}
The main experiments build each class prototype as the centroid of all training embeddings (Eq.~\ref{eq:prototype}).
We use this analysis to assess how many labeled support graphs are needed to estimate a useful prototype.
We evaluate EDGE across all five classes by constructing the prototype from $m\in\{1,5,10,25,50\}$ randomly sampled training graphs and measuring target accuracy, structural coverage, KS calibration, and pairwise diversity. For $m\leq10$ we repeat the random draw multiple times: $5$ draws for $m{=}1$, and $3$ draws for $m{=}5,10$, to capture prototype variance; each draw is evaluated over $5$ generation seeds.

\begin{table}[h!]
\centering
\caption{\textbf{Few-shot prototype ablation} (EDGE+ProtoGuide, mean $\pm$ std over draws and seeds). Target Acc: dynamic $k$-NN accuracy; Cov@95: structural coverage; KS-Cal@95: calibrated KS goodness-of-fit; PW-Ratio: pairwise diversity ratio (gen/real), ${\approx}\,1$ is ideal. \textbf{Bold}= best per class per metric.}
\label{tab:fewshot}
\small
\rowcolors{2}{gray!25}{white}
\begin{tabular}{l r cccc}
\toprule
Class & $m$ & Target Acc (\%) & Cov@95 (\%) & KS-Cal@95 (\%) $\uparrow$ & PW-Ratio \\
\midrule
Biological       &  1 & $\mathbf{63.2 \pm 4.5}$  & $98.7 \pm 1.0$ & $\mathbf{48.9 \pm 5.4}$  & $\mathbf{0.71 \pm 0.04}$ \\
                 &  5 & $63.1 \pm 4.8$  & $98.5 \pm 1.1$ & $48.1 \pm 6.6$  & $0.70 \pm 0.03$ \\
                 & 10 & $61.4 \pm 5.2$  & $98.8 \pm 0.6$ & $46.7 \pm 4.4$  & $0.70 \pm 0.03$ \\
                 & 25 & $60.9 \pm 5.0$  & $98.8 \pm 0.6$ & $46.7 \pm 4.4$  & $\mathbf{0.71 \pm 0.04}$ \\
                 & 50 & $60.3 \pm 5.5$  & $\mathbf{99.1 \pm 0.8}$ & $\mathbf{48.9 \pm 5.4}$  & $0.70 \pm 0.04$ \\
\midrule
Connectome       &  1 & $86.6 \pm 3.2$  & $\mathbf{96.5 \pm 1.9}$ & $20.0 \pm 7.0$ & $0.92 \pm 0.12$ \\
                 &  5 & $82.0 \pm 4.7$  & $95.9 \pm 2.0$ & $14.8 \pm 6.6$  & $\mathbf{0.98 \pm 0.12}$ \\
                 & 10 & $85.0 \pm 4.1$  & $96.4 \pm 1.9$ & $19.3 \pm 8.6$  & $0.94 \pm 0.12$ \\
                 & 25 & $82.8 \pm 4.3$  & $94.4 \pm 2.1$ & $15.6 \pm 8.9$  & $\mathbf{0.98 \pm 0.17}$ \\
                 & 50 & $\mathbf{86.9 \pm 3.8}$  & $95.9 \pm 2.3$ & $\mathbf{22.2 \pm 7.0}$  & $0.94 \pm 0.15$ \\
\midrule
Infrastructure   &  1 & $47.1 \pm 5.8$  & $95.5 \pm 2.4$ & $23.1 \pm 3.0$  & $0.76 \pm 0.19$ \\
                 &  5 & $\mathbf{49.5 \pm 7.7}$  & $\mathbf{95.7 \pm 2.4}$ & $\mathbf{24.4 \pm 6.0}$  & $0.76 \pm 0.20$ \\
                 & 10 & $47.0 \pm 7.0$  & $95.4 \pm 2.0$ & $23.7 \pm 3.8$  & $0.77 \pm 0.21$ \\
                 & 25 & $43.1 \pm 7.6$  & $94.7 \pm 1.2$ & $22.2 \pm 0.0$  & $\mathbf{0.78 \pm 0.19}$ \\
                 & 50 & $44.7 \pm 5.3$  & $95.6 \pm 2.3$ & $\mathbf{24.4 \pm 4.4}$  & $0.74 \pm 0.18$ \\
\midrule
Internet         &  1 & $95.4 \pm 3.4$  & $0.0 \pm 0.0$ & $13.3 \pm 4.4$  & $\mathbf{1.76 \pm 0.21}$ \\
                 &  5 & $93.4 \pm 3.5$  & $0.0 \pm 0.0$ & $14.1 \pm 6.4$  & $1.98 \pm 0.32$ \\
                 & 10 & $95.7 \pm 2.2$  & $0.0 \pm 0.0$ & $\mathbf{16.3 \pm 6.9}$  & $1.88 \pm 0.28$ \\
                 & 25 & $\mathbf{97.5 \pm 2.5}$  & $0.0 \pm 0.0$ & $15.6 \pm 5.4$  & $1.95 \pm 0.23$ \\
                 & 50 & $93.8 \pm 4.9$  & $0.0 \pm 0.0$ & $11.1 \pm 0.0$  & $2.01 \pm 0.34$ \\
\midrule
Social           &  1 & $87.8 \pm 3.3$  & $\mathbf{99.4 \pm 1.0}$ & $55.1 \pm 2.2$  & $\mathbf{0.78 \pm 0.03}$ \\
                 &  5 & $\mathbf{91.2 \pm 2.1}$  & $\mathbf{99.4 \pm 1.0}$ & $\mathbf{55.6 \pm 0.0}$  & $0.77 \pm 0.02$ \\
                 & 10 & $88.6 \pm 4.2$  & $99.1 \pm 1.1$ & $\mathbf{55.6 \pm 0.0}$  & $\mathbf{0.78 \pm 0.03}$ \\
                 & 25 & $90.6 \pm 1.7$  & $99.1 \pm 0.8$ & $53.3 \pm 4.4$  & $\mathbf{0.78 \pm 0.03}$ \\
                 & 50 & $88.1 \pm 2.3$  & $99.1 \pm 1.2$ & $\mathbf{55.6 \pm 0.0}$  & $\mathbf{0.78 \pm 0.03}$ \\
\bottomrule
\end{tabular}
\end{table}
Table~\ref{tab:fewshot} shows that prototype-based guidance is robust to the support-set size over the tested range. Target accuracy, structural coverage, KS calibration, and pairwise diversity remain broadly stable across $m\in\{1,5,10,25,50\}$. In particular, the largest difference in target accuracy between $m{=}1$ and $m{=}50$ is only $2.9$~pp, observed on Biological ($63.2\%$ versus $60.3\%$).

We therefore retain full-training-set prototypes in the main experiments to remove prototype-sampling variance, while these results show that a usable guidance target can be estimated from very small support sets for the classes considered here. Importantly, this experiment varies only prototype construction after both the generative backbone and Siamese encoder have been trained; it therefore demonstrates few-shot prototype specification rather than few-shot learning of a previously unseen class.

\section{Best-of-N Selection}\label{app:bestofn}

We evaluate Best-of-$N$ selection as a sampling-time alternative and complement to ProtoGuide.
We generate a pool of $N$ graphs, score each graph with the frozen Siamese GAT using the same contrastive prototype score employed by ProtoGuide, and retain the top $K{=}64$ samples.
We consider two variants: unconditional Best-of-$N$, where the candidate pool is generated without guidance, and ProtoGuide Best-of-$N$, where ProtoGuide is first used to generate the pool and the same score is then used for selection.
The former tests how far post-hoc selection alone can improve class alignment without altering the reverse process, while the latter tests whether guidance and selection provide complementary gains.
We evaluate $N\in\{128,256\}$, corresponding to $2{\times}$ and $4{\times}$ oversampling, on both EDGE and DiGress across all five classes and $10$ seeds.

\begin{table}[h!]
\centering\small
\caption{\textbf{Best-of-$N$ and ProtoGuide Best-of-$N$ vs.\ ProtoGuide --- EDGE} (mean $\pm$ std over 10 seeds).
  Best-of-$N$ generates $N$ unconditional graphs and keeps the top-64 by contrastive Siamese score; ProtoGuide Best-of-$N$ generates the pool with ProtoGuide before selection.
  \textbf{Bold} = best per class per metric across each condition; $^{\dagger}$ = PW-Ratio $< 0.5$ (strong under-dispersion).}
\label{tab:bestofn_edge}
\rowcolors{2}{gray!25}{white}
\begin{tabular}{ll cccc}
\toprule
Condition & Class & Dyn.\ Acc (\%) $\uparrow$ & Cov@95 (\%) $\uparrow$ & KS-Cal@95 (\%) $\uparrow$ & PW-Ratio ${\approx}\,1$ \\
\midrule
Baseline & Biological & $62.5 {\pm} 5.5$ & $98.9 {\pm} 0.7$ & $\mathbf{46.7 {\pm} 4.4}$ & $0.72 {\pm} 0.04$ \\
 & Connectome & $11.6 {\pm} 3.8$ & $94.8 {\pm} 2.4$ & $38.9 {\pm} 12.4$ & $\mathbf{0.99 {\pm} 0.19}$ \\
 & Infrastructure & $39.2 {\pm} 10.7$ & $96.2 {\pm} 2.7$ & $\mathbf{20.0 {\pm} 4.4}$ & $\mathbf{0.85 {\pm} 0.37}$ \\
 & Internet & $88.3 {\pm} 5.1$ & $0.0 {\pm} 0.0$ & $\mathbf{22.2 {\pm} 0.0}$ & $\mathbf{0.97 {\pm} 0.08}$ \\
 & Social & $52.0 {\pm} 6.0$ & $98.9 {\pm} 1.6$ & $34.4 {\pm} 14.4$ & $0.75 {\pm} 0.05$ \\
\midrule
Best-of-128 & Biological & $67.5 {\pm} 4.1$ & $99.2 {\pm} 0.8$ & $\mathbf{46.7 {\pm} 6.7}$ & $0.58 {\pm} 0.03$ \\
 & Connectome & $23.3 {\pm} 4.1$ & $\mathbf{100.0 {\pm} 0.0}$ & $33.3 {\pm} 0.0$ & $0.42 {\pm} 0.03^{\dagger}$ \\
 & Infrastructure & $60.6 {\pm} 9.9$ & $99.2 {\pm} 0.8$ & $7.8 {\pm} 5.1$ & $0.38 {\pm} 0.04^{\dagger}$ \\
 & Internet & $\mathbf{100.0 {\pm} 0.0}$ & $0.0 {\pm} 0.0$ & $\mathbf{22.2 {\pm} 7.0}$ & $0.74 {\pm} 0.04$ \\
 & Social & $99.5 {\pm} 1.0$ & $98.9 {\pm} 1.0$ & $\mathbf{61.1 {\pm} 5.6}$ & $0.79 {\pm} 0.05$ \\
\midrule
Best-of-256 & Biological & $55.2 {\pm} 2.8$ & $98.9 {\pm} 1.0$ & $17.8 {\pm} 11.3$ & $0.57 {\pm} 0.04$ \\
 & Connectome & $49.5 {\pm} 5.9$ & $\mathbf{100.0 {\pm} 0.0}$ & $33.3 {\pm} 0.0$ & $0.38 {\pm} 0.02^{\dagger}$ \\
 & Infrastructure & $\mathbf{88.8 {\pm} 3.4}$ & $\mathbf{99.7 {\pm} 0.6}$ & $4.4 {\pm} 5.4$ & $0.39 {\pm} 0.03^{\dagger}$ \\
 & Internet & $\mathbf{100.0 {\pm} 0.0}$ & $0.0 {\pm} 0.0$ & $12.2 {\pm} 9.2$ & $0.66 {\pm} 0.03$ \\
 & Social & $\mathbf{100.0 {\pm} 0.0}$ & $99.7 {\pm} 0.6$ & $57.8 {\pm} 4.4$ & $0.75 {\pm} 0.06$ \\
\midrule
ProtoGuide & Biological & $61.4 {\pm} 6.3$ & $98.1 {\pm} 0.9$ & $\mathbf{46.7 {\pm} 4.4}$ & $\mathbf{0.73 {\pm} 0.06}$ \\
 & Connectome & $85.0 {\pm} 3.7$ & $98.0 {\pm} 1.7$ & $58.9 {\pm} 14.9$ & $1.05 {\pm} 0.14$ \\
 & Infrastructure & $34.4 {\pm} 6.7$ & $97.0 {\pm} 2.4$ & $18.9 {\pm} 5.1$ & $0.65 {\pm} 0.21$ \\
 & Internet & $94.8 {\pm} 2.7$ & $0.0 {\pm} 0.0$ & $16.7 {\pm} 7.5$ & $1.85 {\pm} 0.19$ \\
 & Social & $92.0 {\pm} 2.8$ & $99.2 {\pm} 0.8$ & $57.8 {\pm} 6.7$ & $\mathbf{0.81 {\pm} 0.02}$ \\
 \midrule
ProtoGuide Best-of-128 & Biological & $\mathbf{69.5 {\pm} 6.0}$ & $\mathbf{100.0 {\pm} 0.0}$ & $45.6 {\pm} 10.5$ & $0.53 {\pm} 0.02$ \\
 & Connectome & $\mathbf{100.0 {\pm} 0.0}$ & $\mathbf{100.0 {\pm} 0.0}$ & $\mathbf{63.3 {\pm} 5.1}$ & $0.62 {\pm} 0.03$ \\
 & Infrastructure & $51.6 {\pm} 8.2$ & $98.9 {\pm} 1.4$ & $3.3 {\pm} 5.1$ & $0.38 {\pm} 0.04^{\dagger}$ \\
 & Internet & $\mathbf{100.0 {\pm} 0.0}$ & $0.0 {\pm} 0.0$ & $20.0 {\pm} 4.4$ & $1.59 {\pm} 0.13$ \\
 & Social & $\mathbf{100.0 {\pm} 0.0}$ & $99.7 {\pm} 0.6$ & $55.6 {\pm} 0.0$ & $0.68 {\pm} 0.07$ \\
\midrule
ProtoGuide Best-of-256 & Biological & $56.7 {\pm} 6.2$ & $99.7 {\pm} 0.6$ & $24.4 {\pm} 13.9$ & $0.53 {\pm} 0.05$ \\
 & Connectome & $\mathbf{100.0 {\pm} 0.0}$ & $99.8 {\pm} 0.5$ & $52.2 {\pm} 13.2$ & $0.61 {\pm} 0.05$ \\
 & Infrastructure & $86.7 {\pm} 5.5$ & $97.8 {\pm} 1.7$ & $2.2 {\pm} 4.4$ & $0.40 {\pm} 0.03^{\dagger}$ \\
 & Internet & $\mathbf{100.0 {\pm} 0.0}$ & $0.0 {\pm} 0.0$ & $11.1 {\pm} 5.0$ & $1.43 {\pm} 0.18$ \\
 & Social & $\mathbf{100.0 {\pm} 0.0}$ & $\mathbf{100.0 {\pm} 0.0}$ & $41.1 {\pm} 10.0$ & $0.54 {\pm} 0.06$ \\
\bottomrule
\end{tabular}
\end{table}

Table~\ref{tab:bestofn_edge} shows a clear accuracy--diversity trade-off.
At $2{\times}$ oversampling, unconditional Best-of-128 reaches $70.2\%$ macro accuracy, below ProtoGuide's $73.5\%$; at $4{\times}$ oversampling, Best-of-256 reaches $78.7\%$ and therefore exceeds ProtoGuide, with particularly large gains on Infrastructure ($88.8\%$ versus $34.4\%$) and Social ($100.0\%$ versus $92.0\%$).
The higher accuracy is accompanied by strong distributional concentration on some classes: PW-Ratio falls to $0.38$--$0.42$ on Connectome and Infrastructure, indicating substantially lower dispersion than in the corresponding real graph sets.
Best-of-$N$ accuracy is also non-monotonic in pool size for Biological, decreasing from $67.5\%$ at $N{=}128$ to $55.2\%$ at $N{=}256$.

Combining ProtoGuide with selection yields the highest macro accuracy on EDGE: $84.2\%$ at $N{=}128$ and $88.7\%$ at $N{=}256$, compared with $73.5\%$ for ProtoGuide alone and $78.7\%$ for unconditional Best-of-256.
Under EDGE, the two mechanisms are therefore complementary in terms of class accuracy: guidance shifts the candidate distribution before selection, while reranking further concentrates the retained samples.
This combination also has the highest sampling cost, since the ProtoGuide gradient overhead is incurred for every candidate in the oversampled pool before $N-K$ graphs are discarded.

The distributional effect remains class-dependent under selection. Infrastructure exhibits strong under-dispersion, whereas Internet remains over-dispersed, showing that increasing selection pressure does not uniformly move generated distributions toward the real reference set.

\begin{table}[!t]
\centering\small
\caption{\textbf{Best-of-$N$ and ProtoGuide Best-of-$N$ vs.\ ProtoGuide --- DiGress} (mean $\pm$ std over 10 seeds).
  Same protocol as Table~\ref{tab:bestofn_edge}.
  \textbf{Bold} = best per class per metric across each condition; $^{\dagger}$ = PW-Ratio $< 0.5$ (strong under-dispersion).}
\label{tab:bestofn_digress}
\rowcolors{2}{gray!25}{white}
\begin{tabular}{ll cccc}
\toprule
Condition & Class & Dyn.\ Acc (\%) $\uparrow$ & Cov@95 (\%) $\uparrow$ & KS-Cal@95 (\%) $\uparrow$ & PW-Ratio ${\approx}\,1$ \\
\midrule
Baseline & Biological & $41.4 {\pm} 9.7$ & $97.2 {\pm} 5.3$ & $23.3 {\pm} 9.2$ & $\mathbf{0.68 {\pm} 0.24}$ \\
 & Connectome & $90.0 {\pm} 4.9$ & $96.7 {\pm} 0.8$ & $\mathbf{47.8 {\pm} 13.2}$ & $0.93 {\pm} 0.09$ \\
 & Infrastructure & $90.0 {\pm} 5.8$ & $93.4 {\pm} 7.1$ & $17.8 {\pm} 7.4$ & $0.53 {\pm} 0.37$ \\
 & Internet & $95.8 {\pm} 3.0$ & $70.3 {\pm} 5.8$ & $41.1 {\pm} 7.1$ & $1.10 {\pm} 0.09$ \\
 & Social & $50.6 {\pm} 4.5$ & $\mathbf{98.3 {\pm} 1.8}$ & $48.9 {\pm} 5.4$ & $0.73 {\pm} 0.06$ \\
\midrule
Best-of-128 & Biological & $62.8 {\pm} 4.6$ & $\mathbf{100.0 {\pm} 0.0}$ & $30.0 {\pm} 7.1$ & $0.51 {\pm} 0.03$ \\
 & Connectome & $\mathbf{100.0 {\pm} 0.0}$ & $\mathbf{100.0 {\pm} 0.0}$ & $45.6 {\pm} 11.6$ & $0.39 {\pm} 0.02^{\dagger}$ \\
 & Infrastructure & $99.2 {\pm} 1.4$ & $\mathbf{100.0 {\pm} 0.0}$ & $14.4 {\pm} 5.1$ & $0.33 {\pm} 0.02^{\dagger}$ \\
 & Internet & $\mathbf{100.0 {\pm} 0.0}$ & $46.9 {\pm} 3.8$ & $0.0 {\pm} 0.0$ & $0.90 {\pm} 0.04$ \\
 & Social & $\mathbf{100.0 {\pm} 0.0}$ & $94.8 {\pm} 3.4$ & $88.9 {\pm} 0.0$ & $0.95 {\pm} 0.10$ \\
\midrule
Best-of-256 & Biological & $\mathbf{67.3 {\pm} 7.3}$ & $\mathbf{100.0 {\pm} 0.0}$ & $\mathbf{42.2 {\pm} 6.7}$ & $0.47 {\pm} 0.03^{\dagger}$ \\
 & Connectome & $\mathbf{100.0 {\pm} 0.0}$ & $\mathbf{100.0 {\pm} 0.0}$ & $40.0 {\pm} 7.4$ & $0.37 {\pm} 0.02^{\dagger}$ \\
 & Infrastructure & $\mathbf{100.0 {\pm} 0.0}$ & $\mathbf{100.0 {\pm} 0.0}$ & $21.1 {\pm} 7.8$ & $0.33 {\pm} 0.03^{\dagger}$ \\
 & Internet & $\mathbf{100.0 {\pm} 0.0}$ & $53.6 {\pm} 6.6$ & $0.0 {\pm} 0.0$ & $0.77 {\pm} 0.06$ \\
 & Social & $\mathbf{100.0 {\pm} 0.0}$ & $94.7 {\pm} 2.8$ & $\mathbf{91.1 {\pm} 8.3}$ & $\mathbf{1.02 {\pm} 0.10}$ \\
\midrule
ProtoGuide & Biological & $48.6 {\pm} 6.5$ & $99.8 {\pm} 0.5$ & $24.4 {\pm} 4.4$ & $0.56 {\pm} 0.06$ \\
 & Connectome & $91.6 {\pm} 3.8$ & $97.3 {\pm} 1.4$ & $26.7 {\pm} 15.9$ & $\mathbf{1.03 {\pm} 0.18}$ \\
 & Infrastructure & $89.1 {\pm} 4.7$ & $91.7 {\pm} 7.2$ & $\mathbf{22.2 {\pm} 14.9}$ & $0.63 {\pm} 0.43$ \\
 & Internet & $97.7 {\pm} 1.6$ & $\mathbf{71.9 {\pm} 5.0}$ & $\mathbf{44.4 {\pm} 0.0}$ & $\mathbf{1.08 {\pm} 0.07}$ \\
 & Social & $91.9 {\pm} 4.0$ & $93.4 {\pm} 2.5$ & $28.9 {\pm} 13.3$ & $0.95 {\pm} 0.08$ \\
 \midrule
ProtoGuide Best-of-128 & Biological & $66.9 {\pm} 8.3$ & $\mathbf{100.0 {\pm} 0.0}$ & $30.0 {\pm} 7.1$ & $0.52 {\pm} 0.02$ \\
 & Connectome & $\mathbf{100.0 {\pm} 0.0}$ & $99.7 {\pm} 0.6$ & $11.1 {\pm} 8.6$ & $0.68 {\pm} 0.06$ \\
 & Infrastructure & $97.2 {\pm} 1.8$ & $87.3 {\pm} 4.3$ & $20.0 {\pm} 8.3$ & $0.49 {\pm} 0.09^{\dagger}$ \\
 & Internet & $\mathbf{100.0 {\pm} 0.0}$ & $36.2 {\pm} 4.2$ & $0.0 {\pm} 0.0$ & $\mathbf{0.92 {\pm} 0.05}$ \\
 & Social & $\mathbf{100.0 {\pm} 0.0}$ & $66.1 {\pm} 5.1$ & $45.6 {\pm} 3.3$ & $0.91 {\pm} 0.07$ \\
\midrule
ProtoGuide Best-of-256 & Biological & $\mathbf{67.3 {\pm} 6.1}$ & $\mathbf{100.0 {\pm} 0.0}$ & $40.0 {\pm} 8.9$ & $0.43 {\pm} 0.03^{\dagger}$ \\
 & Connectome & $\mathbf{100.0 {\pm} 0.0}$ & $99.8 {\pm} 0.5$ & $11.1 {\pm} 0.0$ & $0.73 {\pm} 0.05$ \\
 & Infrastructure & $97.7 {\pm} 1.9$ & $75.2 {\pm} 5.7$ & $6.7 {\pm} 8.9$ & $\mathbf{0.64 {\pm} 0.07}$ \\
 & Internet & $\mathbf{100.0 {\pm} 0.0}$ & $41.9 {\pm} 5.0$ & $0.0 {\pm} 0.0$ & $0.82 {\pm} 0.05$ \\
 & Social & $\mathbf{100.0 {\pm} 0.0}$ & $57.7 {\pm} 4.4$ & $42.2 {\pm} 6.7$ & $0.94 {\pm} 0.06$ \\
\bottomrule
\end{tabular}
\end{table}

For DiGress (Table~\ref{tab:bestofn_digress}), unconditional Best-of-$N$ is particularly effective at increasing class accuracy.
Best-of-256 reaches $93.5\%$ macro accuracy compared with $83.8\%$ for ProtoGuide, with gains especially on Biological ($67.3\%$ versus $48.6\%$) and Infrastructure ($100.0\%$ versus $89.1\%$).
As for EDGE, however, several classes become substantially under-dispersed: PW-Ratio falls to $0.33$--$0.39$ on Connectome and Infrastructure under unconditional Best-of-128/256.

Combining guidance with selection yields $92.8\%$ macro accuracy at $N{=}128$ and $93.0\%$ at $N{=}256$, essentially matching unconditional Best-of-256 ($93.5\%$) rather than improving on it.
The guided pools also show lower coverage on some classes; for example, Social decreases from $94.8\%$ under Best-of-128 to $66.1\%$ under ProtoGuide Best-of-128, and Infrastructure from $100.0\%$ to $87.3\%$.
Thus, unlike EDGE, DiGress does not show an additional macro-accuracy benefit from applying ProtoGuide before Best-of-$N$ selection under this protocol.

Across both backbones, Best-of-$N$ selection can match or exceed ProtoGuide in classification accuracy when a sufficiently large candidate pool is available, but requires $2{\times}$--$4{\times}$ oversampling and can substantially reduce dispersion in descriptor space.
ProtoGuide instead modifies the reverse trajectory directly and does not require discarding generated samples, although its own fidelity effects remain class-dependent.
Combining guidance and selection is beneficial for EDGE accuracy, whereas for DiGress unconditional selection already reaches the same macro-accuracy range as the combined procedure.

\vfill

\end{document}